\documentclass{article}

\usepackage[table,dvipsnames,svgnames]{xcolor}

\usepackage{microtype}
\usepackage{graphicx}
\usepackage{subfigure}
\usepackage{booktabs} 
\usepackage{dsfont}
\usepackage{hyperref}
\usepackage{ulem}
\usepackage{soul}

\usepackage[accepted]{icml2024}

\usepackage{amsmath}
\usepackage{amssymb}
\usepackage{mathtools}
\usepackage{amsthm}

\usepackage{tabularx}
\usepackage{multirow}
\usepackage{booktabs}
\usepackage{tabulary}  
\usepackage[abs]{overpic}
\usepackage[skins]{tcolorbox}
\usepackage{listings}
\usepackage{enumitem}
\usepackage{soul}

\definecolor{codegreen}{rgb}{0,0.6,0}
\definecolor{codegray}{rgb}{0.5,0.5,0.5}
\definecolor{codepurple}{rgb}{0.58,0,0.82}
\definecolor{backcolour}{rgb}{0.95,0.95,0.92}

\lstdefinestyle{mystyle}{
    backgroundcolor=\color{backcolour},
    commentstyle=\color{codegreen},
    keywordstyle=\color{magenta},
    stringstyle=\color{codepurple},
    basicstyle=\ttfamily\footnotesize,
    breakatwhitespace=false,         
    breaklines=true,                 
    keepspaces=true,                
    showspaces=false,                
    showstringspaces=false,
    showtabs=false,                  
    tabsize=2
}

\definecolor{lightgreen}{HTML}{EDF8FB}
\definecolor{mediumgreen}{HTML}{66C2A4}
\definecolor{darkgreen}{HTML}{006D2C}

\usepackage{caption}
\usepackage{verbatim} 

\usepackage[capitalize,noabbrev]{cleveref}

\theoremstyle{plain}
\newtheorem{theorem}{Theorem}[section]

\theoremstyle{definition}

\theoremstyle{remark}

\usepackage[textsize=tiny]{todonotes}

\newcommand{\lingjiao}[1]{\noindent{\textcolor{green}{\textbf{Lingjiao: } \textsf{#1} }}}

\newcommand{\blockcomment}[1]{}
\renewcommand{\l}{\left}
\renewcommand{\r}{\right}

\newcommand{\I}{\mathds{1}}

\newcommand{\cL}{\mathcal{L}}

\renewcommand{\a}{\alpha}

\icmltitlerunning{
Monitoring AI-Modified Content at Scale: A Case Study on the Impact of ChatGPT on AI Conference Peer Reviews
}

\begin{document}


\twocolumn[
\icmltitle{
Monitoring AI-Modified Content at Scale: \\
A Case Study on the Impact of ChatGPT on AI Conference Peer Reviews
}



\icmlsetsymbol{equal}{*}
\begin{icmlauthorlist}
\icmlauthor{Weixin Liang}{cs,equal}
\icmlauthor{Zachary Izzo}{nec,equal}
\icmlauthor{Yaohui Zhang}{ee,equal}
\icmlauthor{Haley Lepp}{gse}
\icmlauthor{Hancheng Cao}{cs,msande}
\icmlauthor{Xuandong Zhao}{ucsbcs}
\icmlauthor{Lingjiao Chen}{cs}
\icmlauthor{Haotian Ye}{cs}
\icmlauthor{Sheng Liu}{bds}
\icmlauthor{Zhi Huang}{bds}
\icmlauthor{Daniel A. McFarland}{gse,soc,gsb}
\icmlauthor{James Y. Zou}{cs,ee,bds}
\end{icmlauthorlist}

\icmlaffiliation{cs}{Department of Computer Science, Stanford University}
\icmlaffiliation{nec}{Machine Learning Department, NEC Labs America}
\icmlaffiliation{bds}{Department of Biomedical Data Science, Stanford University}
\icmlaffiliation{ee}{Department of Electrical Engineering, Stanford University}
\icmlaffiliation{gse}{Graduate School of Education, Stanford University}
\icmlaffiliation{soc}{Department of Sociology, Stanford University}
\icmlaffiliation{gsb}{Graduate School of Business, Stanford University}
\icmlaffiliation{msande}{Department of Management Science and Engineering, Stanford University} 
\icmlaffiliation{ucsbcs}{Department of Computer Science, UC Santa Barbara}

\icmlcorrespondingauthor{Weixin Liang}{wxliang@stanford.edu}

\icmlkeywords{language models, LLMs}

\vskip 0.3in
]



\printAffiliationsAndNotice{\icmlEqualContribution} 

\begin{abstract}






We present an approach for estimating the fraction of text in a large corpus which is likely to be substantially modified or produced by a large language model (LLM). Our maximum likelihood model leverages expert-written and AI-generated reference texts to accurately and efficiently examine real-world LLM-use at the corpus level. 
We apply this approach to a case study of scientific peer review in AI conferences that took place after the release of ChatGPT: \textit{ICLR} 2024, \textit{NeurIPS} 2023, \textit{CoRL} 2023 and \textit{EMNLP} 2023. Our results suggest that between 6.5\% and 16.9\% of text submitted as peer reviews to these conferences could have been substantially modified by LLMs, i.e. beyond spell-checking or minor writing updates. The circumstances in which generated text occurs offer insight into user behavior: the estimated fraction of LLM-generated text is higher in reviews which report lower confidence, were submitted close to the deadline, and from reviewers who are less likely to respond to author rebuttals. We also observe corpus-level trends in generated text which may be too subtle to detect at the individual level, and discuss the implications of such trends on peer review. We call for future interdisciplinary work to examine how LLM use is changing our information and knowledge practices. 

 %
 %

\end{abstract}

\begin{figure}[ht!]
    \centering
    \includegraphics[width=0.475\textwidth]{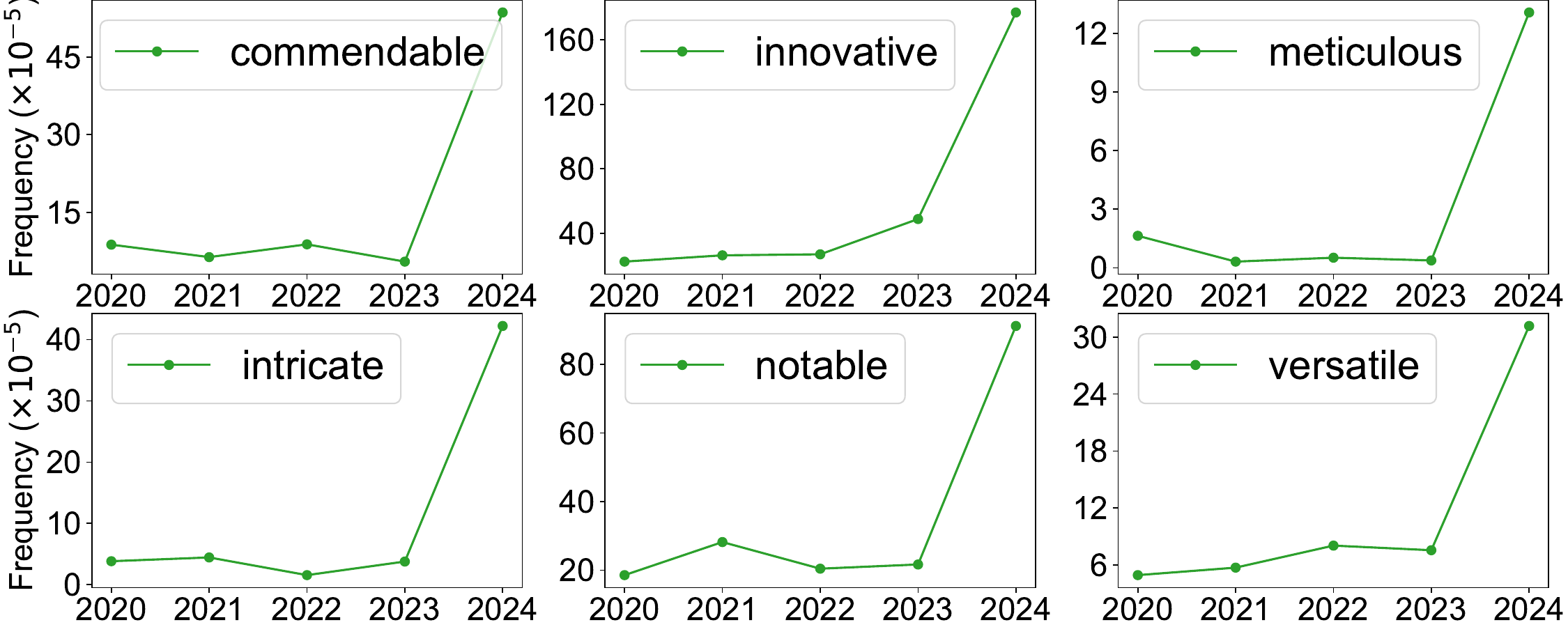}
    \caption{
        \textbf{Shift in Adjective Frequency in \textit{ICLR} 2024 Peer Reviews.} 
        We find a significant shift in the frequency of certain tokens in \textit{ICLR} 2024, with
        adjectives such as “commendable”, “meticulous”, and “intricate” showing 9.8, 34.7, and 11.2-fold increases in probability of occurring in a sentence. We find a similar trend in \textit{NeurIPS} but not in \textit{Nature Portfolio} journals. Supp. Table~\ref{table:word_adj_list} and Supp. Figure~\ref{fig:word-cloud-adj} in the Appendix provide a visualization of the top 100 adjectives produced disproportionately by AI. 
        }
    \label{fig:word-frequency-commendable}
\end{figure}

\section{Introduction}

While the last year has brought extensive discourse and speculation about the widespread use of large language models (LLM) in sectors as diverse as education \cite{bearman2023discourses}, the sciences \cite{van2023ai, messeri}, and global media \cite{fake-news-66}, as of yet it has been impossible to precisely measure the scale of such use or evaluate the ways that the introduction of generated text may be affecting information ecosystems. 
To complicate the matter, it is increasingly difficult to distinguish examples of LLM-generated texts from human-written content ~\citep{Abstracts-written-by-ChatGPT-fool-scientists,huaman-detect-gpt3}. Human capability to discern AI-generated text from human-written content barely exceeds that of a random classifier~\cite{human-hard-to-detect-generated-text,Nature-news-Abstracts-written-by-ChatGPT-fool-scientists, huaman-detect-gpt3}, heightening the risk that unsubstantiated generated text can masquerade as authoritative, evidence-based writing. In scientific research, for example, studies have found that ChatGPT-generated medical abstracts may frequently bypass AI-detectors and experts~\cite{Nature-news-Abstracts-written-by-ChatGPT-fool-scientists,Abstracts-written-by-ChatGPT-fool-scientists}. In media, one study identified over 700 unreliable AI-generated news sites across 15 languages which could mislead consumers~\cite{NewsGuard2023,Cantor2023}.

Despite the fact that generated text may be indistinguishable on a case-by-case basis from content written by humans, studies of LLM-use at scale find corpus-level trends which contrast with at-scale human behavior. For example, the increased consistency of LLM output can amplify biases at the corpus-level in a way that is too subtle to grasp by examining individual cases of use. \citeauthor{bommasani2022picking} find that the ``monocultural" use of a single algorithm for hiring decisions can lead to ``outcome homogenization" of who gets hired---an effect which could not be detected by evaluating hiring decisions one-by-one. \citeauthor{cao2023assessing} find that prompts to ChatGPT in certain languages can reduce the variance in model responses, ``flattening out cultural differences and biasing them towards American culture"; a subtle yet persistent effect that would be impossible to detect at an individual level. These studies rely on experiments and simulations to demonstrate the importance of analyzing and evaluating LLM output at an aggregate level. As LLM-generated content spreads to increasingly high-stakes information ecosystems, there is an urgent need for efficient methods which allow for comparable evaluations on \textit{real-world datasets} which contain uncertain amounts of AI-generated text.

We propose a new framework to efficiently monitor AI-modified content in an information ecosystem: \textit{distributional GPT quantification} (Figure~\ref{fig: schematic}). In contrast with \textit{instance}-level detection, this framework focuses on \textit{population}-level estimates (Section $\S$~\ref{sec: notation}). 
We demonstrate how to estimate the proportion of content in a given corpus that has been generated or significantly modified by AI, without the need to perform inference on any \textit{individual} instance (Section $\S$~\ref{subsec:overview}). 
Framing the challenge as a parametric inference problem, we combine reference text which is known to be human-written or AI-generated with a maximum likelihood estimation (MLE) of text from uncertain origins (Section $\S$~\ref{sec: mle}).
Our approach is more than 10 million times (i.e., 7 orders of magnitude) more computationally efficient than state-of-the-art AI text detection methods (Table~\ref{table:baseline-computation-cost}), while still outperforming them by reducing the in-distribution estimation error by a factor of 3.4, and the out-of-distribution estimation error by a factor of 4.6 (Section $\S$~\ref{subsec:validation},\ref{subsec:baseline}).

Inspired by empirical evidence that the usage frequency of these specific adjectives like ``commendable'' suddenly spikes in the most recent \textit{ICLR} reviews (Figure~\ref{fig:word-frequency-commendable}), we run systematic validation experiments to show that these adjectives occur disproportionately more frequently in AI-generated texts than in human-written reviews (Supp. Table~\ref{table:word_adj_list},\ref{table:word_adv_list}, Supp. Figure~\ref{fig:word-cloud-adj},\ref{fig:word-cloud-adv}). These adjectives allow us to parameterize our compound probability distribution framework (Section $\S$~\ref{sec: dist}), thereby producing more empirically stable and pronounced results (Section $\S$~\ref{subsec:validation}, Figure~\ref{fig: val}). However, we also demonstrate that similar results can be achieved with adverbs, verbs, and non-technical nouns (Appendix~\ref{Appendix:subsec:adverbs}, \ref{Appendix:subsec:verbs}, \ref{Appendix:subsec:nouns}).

We demonstrate this approach through an in-depth case study of texts submitted as reviews to several top AI conferences, including \textit{ICLR}, \textit{NeurIPS}, \textit{EMNLP}, and \textit{CoRL} (Section $\S$~\ref{subsec:Data}, Table~\ref{tab:data_split}) as well as through reviews submitted to the \textit{Nature} family journals (Section $\S$~\ref{sec: main-results}). We find evidence that a small but significant fraction of reviews written for AI conferences after the release of ChatGPT could be substantially modified by AI beyond simple grammar and spell checking (Section $\S$~\ref{subsec:Proofreading},\ref{subsec:expand}, Figure~\ref{fig: temporal},\ref{fig: proofread},\ref{fig:expand-verfication}). In contrast, we do not detect this change in reviews in \textit{Nature} family journals (Figure~\ref{fig: temporal}), and we did not observe a similar trend of Figure~\ref{fig:word-frequency-commendable} (Section $\S$~\ref{sec: main-results}). Finally, we show several ways to measure the implications of generated text in this information ecosystem (Section $\S$~\ref{subsec:fine-grained-analysis}). First, we explore the circumstances in AI-generated text appears more frequently, and second, we demonstrate how AI-generated text appears to differ from expert-written reviews \textit{at the corpus level} (See summary in Box 1). 

Throughout this paper, we refer to texts written by human experts as ``peer reviews" and texts produced by LLMs as ``generated texts``. We do not intend to make an ontological claim as to whether generated texts constitute peer reviews; any such implication through our word choice is unintended. 
 

In summary, \textbf{our contributions} are as follows:
\begin{enumerate}[topsep=0pt, left=0pt]
    
    \item We propose a simple and effective method for estimating the fraction of text in a large corpus that has been substantially modified or generated by AI (Section $\S$~\ref{sec: method}). The method uses historical data known to be human expert or AI-generated (Section $\S$~\ref{sec: data}), and leverages this data to compute an estimate for the fraction of AI-generated text in the target corpus via a maximum likelihood approach (Section $\S$~\ref{sec: dist}).

    \item We conduct a case study on reviews submitted to several top ML and scientific venues, including recent \textit{ICLR}, \textit{NeurIPS}, \textit{EMNLP}, \textit{CoRL} conferences, as well as papers published at \textit{Nature portfolio} journals (Section $\S$~\ref{sec:Experiments}). 
    Our method allows us to uncover trends in AI usage since the release of ChatGPT and corpus-level changes that occur when generated texts appear in an information ecosystem  (Section $\S$~\ref{subsec:fine-grained-analysis}).
    
\end{enumerate}

\begin{figure}[!ht]
\begin{tcolorbox}[enhanced, opacityback=0.1, opacityframe=0.1,      
    colback=lightgreen, 
    colframe=darkgreen, 
    colbacktitle=mediumgreen, 
    coltitle=white, 
    top=1mm, bottom=1mm, left=1mm, right=1mm, title=\textbf{Box 1: Summary of Main Findings}]
\begin{small}
\noindent
1. \textbf{Main Estimates:} Our estimates suggest that 10.6\% of \textit{ICLR} 2024 review sentences and 16.9\% for \textit{EMNLP} have been substantially modified by ChatGPT, with no significant evidence of ChatGPT usage in \textit{Nature portfolio} reviews (Section $\S$~\ref{sec: main-results}, Figure~\ref{fig: temporal}).\\
2. \textbf{Deadline Effect:} Estimated ChatGPT usage in reviews spikes significantly within 3 days of review deadlines (Section $\S$~\ref{subsec:fine-grained-analysis}, Figure~\ref{fig: deadline}).\\
3. \textbf{Reference Effect:} Reviews containing scholarly citations are less likely to be AI modified or generated than those lacking such citations (Section $\S$~\ref{subsec:fine-grained-analysis}, Figure~\ref{fig: et-al}). \\
4. \textbf{Lower Reply Rate Effect:} Reviewers who do not respond to \textit{ICLR}/\textit{NeurIPS} author rebuttals show a higher estimated usage of ChatGPT (Section $\S$~\ref{subsec:fine-grained-analysis}, Figure~\ref{fig: reply}). \\
5. \textbf{Homogenization Correlation:} 
Higher estimated AI modifications are correlated with homogenization of review content in the text embedding space (Section $\S$~\ref{subsec:fine-grained-analysis}, Figure~\ref{fig: homog}). \\
6. \textbf{Low Confidence Correlation:} Low self-reported confidence in reviews are associated with an increase of ChatGPT usage (Section $\S$~\ref{subsec:fine-grained-analysis}, Figure~\ref{fig: confidence}). 
\end{small}
\end{tcolorbox}
\end{figure}
\section{Related Work}

\paragraph{Zero-shot LLM detection.} 
Many approaches to LLM detection aim to detect AI-generated text at the level of individual documents. Zero-shot detection or ``model self-detection" represents a major approach family, utilizing the heuristic that text generated by an LLM will exhibit distinctive probabilistic or geometric characteristics within the very model that produced it. Early methods for LLM detection relied on metrics like entropy \cite{Lavergne2008DetectingFC}, log-probability scores \cite{Solaiman2019ReleaseSA}, perplexity \cite{Beresneva2016ComputerGeneratedTD}, and uncommon n-gram frequencies \cite{Badaskar2008IdentifyingRO} from language models to distinguish between human and machine text. More recently, DetectGPT \citep{Mitchell2023DetectGPTZM} suggests that AI-generated text typically occupies regions with negative log probability curvature. DNA-GPT \cite{Yang2023DNAGPTDN} improves performance by analyzing n-gram divergence between re-prompted and original texts. Fast-DetectGPT \cite{Bao2023FastDetectGPTEZ} enhances efficiency by leveraging conditional probability curvature over raw probability. \citet{Tulchinskii2023IntrinsicDE} show that machine text has lower intrinsic dimensionality than human writing, as measured by persistent homology for dimension estimation.
However, these methods are most effective when there is direct access to the internals of the specific LLM that generated the text. Since many commercial LLMs, including OpenAI's GPT-4, are not open-sourced, these approaches often rely on \textit{a proxy LLM} assumed to be mechanistically similar to the closed-source LLM. This reliance introduces compromises that, as studies by \cite{Sadasivan2023CanAT, Shi2023RedTL, Yang2023ASO, Zhang2023AssayingOT} demonstrate, limit the robustness of zero-shot detection methods across different scenarios. 

\paragraph{Training-based LLM detection.} 
An alternative LLM detection approach is to fine-tune a pretrained model on datasets with both human and AI-generated text examples in order to distinguish between the two types of text, bypassing the need for original model access. Earlier studies have used classifiers to detect synthetic text in peer review corpora \cite{Bhagat2013SquibsWI}, media outlets
 \cite{Zellers2019DefendingAN}, and other contexts \cite{Bakhtin2019RealOF, Uchendu2020AuthorshipAF}.  
More recently, GPT-Sentinel \cite{Chen2023GPTSentinelDH} train the RoBERTa \cite{Liu2019RoBERTaAR} and T5 \cite{raffel2020exploring} classifiers on the constructed dataset OpenGPTText. GPT-Pat \cite{Yu2023GPTPT} train a twin neural network to compute the similarity between original and re-decoded texts. \citet{Li2023DeepfakeTD} build a wild testbed by gathering texts from various human writings and deepfake texts generated by different LLMs. Notably, the application of contrastive and adversarial learning techniques has enhanced classifier robustness \cite{Liu2022CoCoCM, Bhattacharjee2023ConDACD, Hu2023RADARRA}. 
However, the recent development of several publicly available tools aimed at mitigating the risks associated with AI-generated content has sparked a debate about their effectiveness and reliability~\citep{OpenAIGPT2,jawahar2020automatic,fagni2021tweepfake,ippolito2019automatic,mitchell2023detectgpt,human-hard-to-detect-generated-text,mit-technology-review-how-to-spot-ai-generated-text,survey-2023, solaiman2019release}. This discussion gained further attention with OpenAI's 2023 decision to discontinue its AI-generated text classifier due to its “low rate of accuracy”~\cite{Kirchner2023,Kelly2023}. 

A major empirical challenge for training-based methods is their tendency to overfit to both training data and language models. Therefore, many classifiers show vulnerability to adversarial attacks \cite{Wolff2020AttackingNT} and display bias towards writers of non-dominant language varieties \cite{Liang2023GPTDA}.
The theoretical possibility of achieving accurate \textit{instance}-level detection has also been questioned by researchers, with debates exploring whether reliably distinguishing AI-generated content from human-created text on an individual basis is fundamentally impossible~\cite{Weber-Wulff2023,Sadasivan2023CanAT,chakraborty2023possibilities}. Unlike these approaches to detecting AI-generated text at the document, paragraph, or sentence level, our method estimates the fraction of an entire text corpus which is substantially AI-generated. Our extensive experiments demonstrate that by sidestepping the intermediate step of classifying individual documents or sentences, this method improves upon the stability, accuracy, and computational efficiency of existing approaches. 

\paragraph{LLM watermarking.} 
Text watermarking introduces a method to detect AI-generated text by embedding unique, algorithmically-detectable signals -known as watermarks- directly into the text. Early watermarking approaches modify pre-existing text by leveraging synonym substitution \citep{Chiang2003NaturalLW, Topkara2006TheHV}, syntactic structure restructuring \citep{Atallah2001NaturalLW, Topkara2006NaturalLW}, or paraphrasing \citep{Atallah2002NaturalLW}. Increasingly, scholars have focused on integrating a watermark directly into an LLM's decoding process.
\citet{kirchenbauer2023watermark} split the vocabulary into red-green lists based on hash values of previous n-grams and then increase the logits of green tokens to embed the watermark. \citet{Zhao2023ProvableRW} use a global red-green list to enhance robustness.  \citet{ Hu2023UnbiasedWF,Kuditipudi2023RobustDW, Wu2023DiPmarkAS} study watermarks that preserve the original token probability distributions. Meanwhile, semantic watermarks \cite{Hou2023SemStampAS, Fu2023WatermarkingCT, Liu2023ASI} using input sequences to find semantically related tokens and multi-bit watermarks \cite{Yoo2023RobustMN, Fernandez2023ThreeBT} to embed more complex information have been proposed to improve certain conditional generation tasks. 
However, watermarking requires the involvement of the model or service owner, such as OpenAI, to implant the watermark. Concerns have also been raised regarding the potential for watermarking to degrade text generation quality and to compromise the coherence and depth of LLM responses~\cite{singh2023new}. In contrast, our framework operates \textit{independently} of the model or service owner's intervention, allowing for the monitoring of AI-modified content without requiring their adoption.



\section{Method} \label{sec: method}

\subsection{Notation \& Problem Statement} \label{sec: notation}
Let $x$ represent a document or sentence, and let $t$ be a token. We write $t \in x$ if the token $t$ occurs in the document $x$. We will use the notation $X$ to refer to a \emph{corpus} (i.e., a collection of individual documents or sentences $x$) and $V$ to refer to a \emph{vocabulary} (i.e., a collection of tokens $t$).
In all of our experiments in the main body of the paper, we take the vocabulary $V$ to be the set of all \emph{adjectives}. Experiments comparing against these other possibilities such as adverbs, verbs, nouns can be found in the Appendix~\ref{Appendix:subsec:adverbs},\ref{Appendix:subsec:verbs},\ref{Appendix:subsec:nouns}. That is, all of our calculations depend only on the adjectives contained in each document. We found this vocabulary choice to exhibit greater stability than using other parts of speech such as adverbs, verbs, nouns, or all possible tokens.  
We removed technical terms by excluding the set of all technical keywords as self-reported by the authors during abstract submission on OpenReview. 

Let $P$ and $Q$ denote the probability distribution of documents written by scientists and generated by AI, respectively. Given a document $x$, we will use $P(x)$ (resp. $Q(x)$) to denote the likelihood of $x$ under $P$ (resp. $Q$). We assume that the documents in the target corpus are generated from the mixture distribution 
\begin{equation} \label{eq: mix}
(1-\a)P + \a Q
\end{equation}
and the goal is to estimate the fraction $\a$ which are AI-generated.

\subsection{Overview of Our Statistical Estimation Approach}
\label{subsec:overview}

\begin{figure*}[ht!]
\centering
\includegraphics[width=0.8\textwidth]{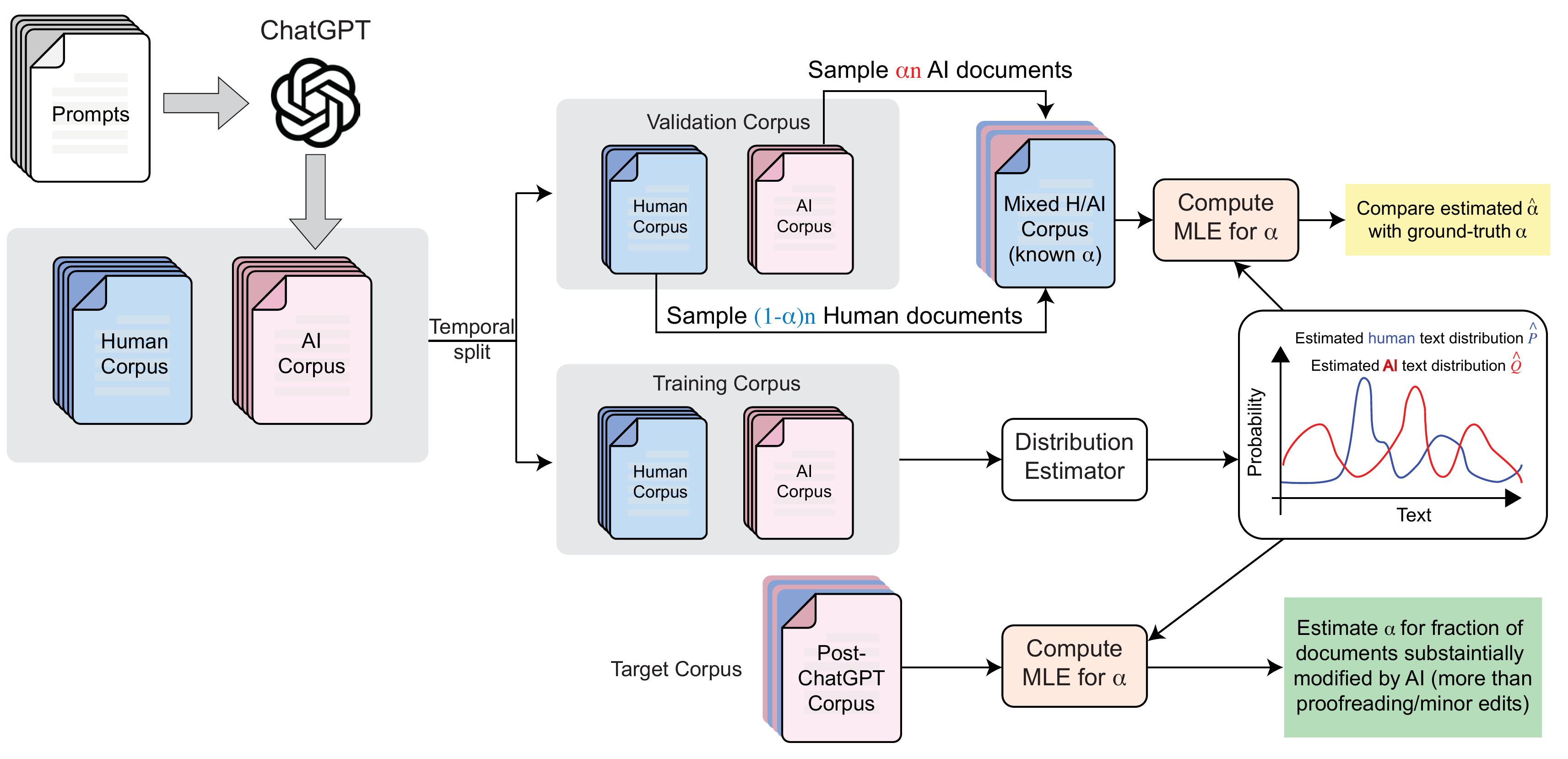}
\caption{\textbf{An overview of the method.} We begin by generating a corpus of documents with known scientist or AI authorship. Using this historical data, we can estimate the scientist-written and AI text distributions $P$ and $Q$ and validate our method's performance on held-out data. Finally, we can use the estimated $P$ and $Q$ to estimate the fraction of AI-generated text in a target corpus.
}
\label{fig: schematic}
\end{figure*}

LLM detectors are known to have unstable performance (Section $\S$~\ref{subsec:baseline}). Thus, rather than trying to classify each document in the corpus and directly count the number of occurrences in this manner, we take a maximum likelihood approach.
Our method has three components: training data generation, document probability distribution estimation, and computing the final estimate of the fraction of text that has been substantially modified or generated by AI. The method is summarized graphically in Figure~\ref{fig: schematic}. A non-graphical summary is as follows:
\begin{enumerate}[noitemsep,topsep=0pt]
    \item Collect the writing instructions given to (human) authors for the original corpus- in our case, peer review instructions. Give these instructions as prompts into an LLM to generate a corresponding corpus of AI-generated documents (Section $\S$~\ref{sec: data}).
    \item Using the human and AI document corpora, estimate the reference token usage distributions $P$ and $Q$ (Section $\S$~\ref{sec: dist}).
    \item Verify the method's performance on synthetic target corpora where the correct proportion of AI-generated documents is known (Section $\S$~\ref{sec: val}).
    \item Based on these estimates for $P$ and $Q$, use MLE to estimate the fraction $\a$ of AI-generated or modified documents in the target corpus (Section $\S$~\ref{sec: mle}).
\end{enumerate}
The following sections present each of these steps in more detail.

\subsection{MLE Framework} \label{sec: mle}
Given a collection of $n$ documents $\{x_i\}_{i=1}^n$ drawn independently from the mixture \eqref{eq: mix}, the log-likelihood of the corpus is given by
\begin{equation} \label{eq: log likelihood}
\cL(\a) = \sum_{i=1}^n \log\l( (1-\a) P(x_i) + \a  Q(x_i) \r).
\end{equation}
If $P$ and $Q$ are known, we can then estimate $\a$ via \emph{maximum likelihood estimation} (MLE) on \eqref{eq: log likelihood}. This is the final step in our method. It remains to construct accurate estimates for $P$ and $Q$.


\subsection{Generating the Training Data} \label{sec: data}
We require access to historical data for estimating $P$ and $Q$. Specifically, we assume that we have access to a collection of reviews which are known to contain only human-authored text, along with the associated review questions and the reviewed papers. We refer to the collection of such \emph{documents} as the \emph{human corpus}.

To generate the AI corpus, we prompt the LLM to generate a review given a paper. 
The texts output by the LLM are then collected into the \emph{AI corpus}. Empirically, we found that our framework exhibits moderate robustness to the distribution shift of LLM prompts. As discussed in Appendix~\ref{Appendix:subsec:LLM-prompt-shift}, training with one prompt and testing with a different prompt still yield accurate validation results (see Supp. Figure~\ref{fig: diff prompt val}).

\subsection{Estimating $P$ and $Q$ from Data} \label{sec: dist}
The space of all possible documents is too large to estimate $P(x), Q(x)$ directly. Thus, we make some simplifying assumptions on the document generation process to make the estimation tractable.

We represent each document $x_i$ as a list of \emph{occurrences} (i.e., a set) of tokens rather than a list of token \emph{counts}. While longer documents will tend to have more unique tokens (and thus a lower likelihood in this model), the number of additional unique tokens is likely sublinear in the document length, leading to a less exaggerated down-weighting of longer documents.\footnote{For the intuition behind this claim, one can consider the extreme case where the entire token vocabulary has been used in the first part of a document. As more text is added to the document, there will be no new token occurrences, so the number of unique tokens will remain constant regardless of how much length is added to the document. In general, even if the entire vocabulary of unique tokens has not been exhausted, as the document length increases, it is more likely that previously seen tokens will be re-used rather than introducing new ones. This can be seen as analogous to the \href{https://en.wikipedia.org/wiki/Coupon_collector\%27s_problem}{coupon collector problem}~\cite{newman1960double}.} 


The occurrence probabilities for the human document distribution can be estimated by
\begin{align*}
\hat{p}(t) &= \frac{\textrm{\# documents in which token } t \textrm{ appears}}{\textrm{total \# documents in the corpus}} \\[5pt]
&= \frac{\sum_{x \in X} \I\{t \in x\}}{|X|},
\end{align*}
where $X$ is the corpus of human-written documents. The estimate $\hat{q}(t)$ can be defined similarly for the AI distribution. Using the notation $t \in x$ to denote that token $t$ occurs in document $x$, we can then estimate $P$ via
\begin{equation} \label{eq: occur}
P(x_i) = \prod_{t\in x} \hat{p}(t) \times  \prod_{t\not\in x} (1-\hat{p}(t))
\end{equation}
and similarly for $Q$. Recall that our token vocabulary $V$ (defined in Section $\S$~\ref{sec: notation}) consists of all adjectives, so the product over $t\not\in x$ means the product only over all adjectives $t$ which were not in the document or sentence $x$.

We validated both approaches using either a document or a sentence as the unit of $x$, and both performed well (Appendix \ref{subsec:Results on Document-Level Analysis}). We used a sentence as our main unit for estimates, as sentences perform slightly better.

\subsection{Validating the Method} \label{sec: val}
The steps described above are sufficient for estimating the fraction $\a$ of documents in a target corpus which are AI-generated. We also provide a method for validating the system's performance.

We use the training partitions of the human and AI corpora to estimate $P$ and $Q$ as described above. To validate the system's performance, we do the following:
\begin{enumerate}[noitemsep,topsep=0pt]
    \item Choose a range of feasible values for $\a$, e.g. $\a \in \{0, 0.05, 0.1, 0.15, 0.2, 0.25\}$.
    \item Let $n$ be the size of the target corpus. For each of the selected $\a$ values, sample (with replacement) $\a n$ documents from the AI validation corpus and $(1-\a)n$ documents from the human validation corpus to create a \emph{target corpus}.
    \item Compute the MLE estimate $\hat{\a}$ on the target corpus. If $\hat{\a}\approx\a$ for each of the feasible $\a$ values, this provides evidence that the system is working correctly and the estimate can be trusted. 
\end{enumerate}
Step 2 can also be repeated multiple times to generate confidence intervals for the estimate $\hat{\a}$.

\section{Experiments}
\label{sec:Experiments}
In this section, we apply our method to a case study of peer reviews of academic machine learning (ML) and scientific papers. We report our results graphically; numerical results and the results for additional experiments can be found in Appendix~\ref{appendix: additional results}.

\subsection{Data}
\label{subsec:Data}
We collect review data for all major ML conferences available on OpenReview, including \textit{ICLR}, \textit{NeurIPS}, \textit{CoRL}, and \textit{EMNLP}, as detailed in Table~\ref{tab:data_split}. 
The Nature portfolio dataset encompasses 15 journals within the Nature portfolio, such as Nature,
Nature Biomedical Engineering, Nature Human Behaviour, and Nature Communications.
Additional information on the datasets can be found in Appendix~\ref{sec:Additional Dataset Information}.


\begin{table}[ht!]
\centering
\caption{
\textbf{Academic Peer Reviews Data from Major ML Conferences.}
All listed conferences except \textit{ICLR} '24, \textit{NeurIPS} '23, \textit{CoRL} '23, and \textit{EMNLP} '23 underwent peer review before the launch of ChatGPT on November 30, 2022. 
We use the \textit{ICLR} '23 conference data for in-distribution validation, and the \textit{NeurIPS} ('17–'22) and \textit{CoRL} ('21–'22) for out-of-distribution (OOD) validation.
}
\label{tab:data_split}
\resizebox{0.48\textwidth}{!}{
\setlength{\tabcolsep}{3.5pt}
\begin{tabular}{lccc}
\toprule
\bf Conference & \bf Post ChatGPT & \bf Data Split & \bf \# of Official Reviews \\
\midrule
\rowcolor{green!10} 
ICLR 2018 & \bf \textcolor{darkgreen!70}{Before} & \bf \cellcolor{green!20} \textcolor{black!85}{Training} & 2,930 \\
\rowcolor{green!10} 
ICLR 2019 & \bf \textcolor{darkgreen!70}{Before} & \bf \cellcolor{green!20} \textcolor{black!85}{Training} & 4,764 \\
\rowcolor{green!10} 
ICLR 2020 & \bf \textcolor{darkgreen!70}{Before} & \bf \cellcolor{green!20} \textcolor{black!85}{Training} & 7,772 \\
\rowcolor{green!10} 
ICLR 2021 & \bf \textcolor{darkgreen!70}{Before} & \bf \cellcolor{green!20} \textcolor{black!85}{Training} & 11,505 \\
\rowcolor{green!10} 
ICLR 2022 & \bf \textcolor{darkgreen!70}{Before} & \bf \cellcolor{green!20} \textcolor{black!85}{Training} & 13,161 \\
\cmidrule{1-4} 
\rowcolor{green!10} 
ICLR 2023 & \bf \textcolor{darkgreen!70}{Before} & \bf \cellcolor{blue!10} \textcolor{blue!85}{Validation} & 18,564 \\
\rowcolor{red!20} ICLR 2024 & \bf \textcolor{red!70}{After} & \bf \textcolor{black!85}{Inference} & 27,992 \\
\cmidrule{1-4} 
\rowcolor{green!10} 
NeurIPS 2017 & \bf \textcolor{darkgreen!70}{Before} & \bf \cellcolor{blue!10} \textcolor{blue!85}{OOD Validation} & 1,976 \\
\rowcolor{green!10} 
NeurIPS 2018 & \bf \textcolor{darkgreen!70}{Before} & \bf \cellcolor{blue!10} \textcolor{blue!85}{OOD Validation} & 3,096 \\
\rowcolor{green!10} 
NeurIPS 2019 & \bf \textcolor{darkgreen!70}{Before} & \bf \cellcolor{blue!10} \textcolor{blue!85}{OOD Validation} & 4,396 \\
\rowcolor{green!10} 
NeurIPS 2020 & \bf \textcolor{darkgreen!70}{Before} & \bf \cellcolor{blue!10} \textcolor{blue!85}{OOD Validation} & 7,271 \\
\rowcolor{green!10} 
NeurIPS 2021 & \bf \textcolor{darkgreen!70}{Before} & \bf \cellcolor{blue!10} \textcolor{blue!85}{OOD Validation} & 10,217 \\
\rowcolor{green!10} 
NeurIPS 2022 & \bf \textcolor{darkgreen!70}{Before} & \bf \cellcolor{blue!10} \textcolor{blue!85}{OOD Validation} & 9,780 \\
\rowcolor{red!20} NeurIPS 2023 & \bf \textcolor{red!70}{After} & \bf \textcolor{black!85}{Inference} & 14,389 \\
\cmidrule{1-4} 
\rowcolor{green!10} 
CoRL 2021 & \bf \textcolor{darkgreen!70}{Before} & \bf \cellcolor{blue!10} \textcolor{blue!85}{OOD Validation} & 558 \\
\rowcolor{green!10} 
CoRL 2022 & \bf \textcolor{darkgreen!70}{Before} & \bf \cellcolor{blue!10} \textcolor{blue!85}{OOD Validation} & 756 \\
\rowcolor{red!20} CoRL 2023 & \bf \textcolor{red!70}{After} & \bf \textcolor{black!85}{Inference} & 759 \\
\cmidrule{1-4} 
\rowcolor{red!20} EMNLP 2023 & \bf \textcolor{red!70}{After} & \bf \textcolor{black!85}{Inference} & 6,419 \\
\bottomrule
\end{tabular}
}
\end{table}

\subsection{Validation on Semi-Synthetic data}
\label{subsec:validation}
\begin{figure}[ht!]
    \centering
    \includegraphics[width=0.475\textwidth]{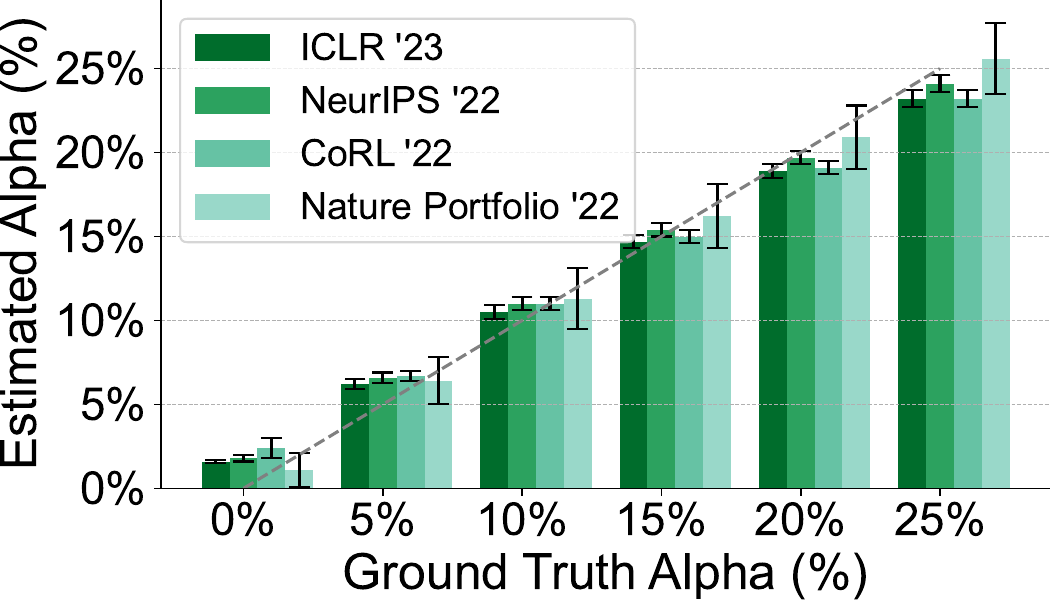}
\caption{
\textbf{Performance validation of our MLE estimator} across \textit{ICLR} '23, \textit{NeurIPS} '22, and \textit{CoRL} '22 reviews (all predating ChatGPT's launch) via the method described in Section~\ref{sec: val}.
Our algorithm demonstrates high accuracy with less than 2.4\% prediction error in identifying the proportion of LLM-generated feedback within the validation set. 
See {Supp. Table~\ref{tab:verification-adj-main},\ref{tab:verification-adj-main-nature}} for full results.
}
\label{fig: val}
\end{figure}

Next, we validate the efficacy of our method as described in Section~\ref{sec: val}. We find that our algorithm accurately estimates the proportion of LLM-generated texts in these mixed validation sets with a prediction error of less than 1.8\% at the population level across various ground truth $\alpha$ on \textit{ICLR} '23 (Figure~\ref{fig: val}, Supp. Table~\ref{tab:verification-adj-main}). 

Furthermore, despite being trained exclusively on \textit{ICLR} data from 2018 to 2022, our model displays robustness to moderate topic shifts observed in \textit{NeurIPS} and \textit{CoRL} papers. The prediction error remains below 1.8\% across various ground truth $\alpha$ for \textit{NeurIPS} '22 and under 2.4\% for \textit{CoRL} '22 (Figure~\ref{fig: val}, Supp. Table~\ref{tab:verification-adj-main}). This resilience against variation in paper content suggests that our model can reliably identify LLM alterations  
across different research areas and conference formats, underscoring its potential applicability in maintaining the integrity of the peer review process in the presence of continuously updated generative models. 


\subsection{Comparison to Instance-Based Detection Methods}
\label{subsec:baseline}

We compare our approach to a BERT classifier baseline, which we fine-tuned on identical training data, and two recently published, state-of-the-art AI text detection methods, all evaluated using the same protocol (Appendix~\ref{Appendix:subsec:baselines}).
Our method reduces the in-distribution estimation error by 3.4 times compared to the best-performing baseline (from 6.2\% to 1.8\%, Supp. Table~\ref{tab:BERT-and-other-baselines}), and the out-of-distribution estimation error by 4.6 times (from 11.2\% to 2.4\%, Supp. Table~\ref{tab:BERT-and-other-baselines}).  Additionally, our method is 
more than 10 million times (i.e., 7 orders of magnitude) more computationally efficient during inference time (68.09 FLOPS vs. 2.721 $\times 10^9$ FLOPS amortized per sentence, Supp. Table~\ref{table:baseline-computation-cost}), and the training cost is also negligible compared to any backpropagation-based algorithms as we are only counting word frequencies in the training corpora.



\subsection{Estimates on Real Reviews}\label{sec: main-results}
\begin{figure}[ht!] 
    \centering
    \includegraphics[width=0.475\textwidth]{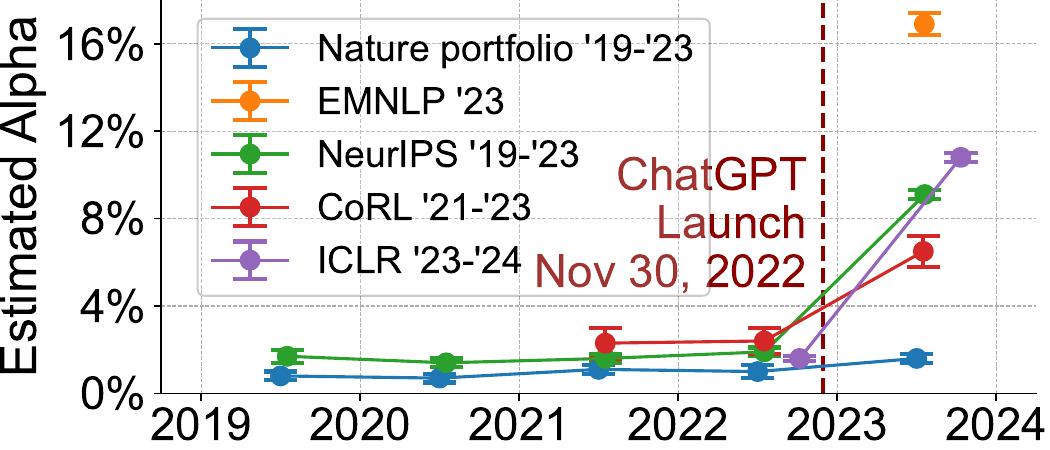}
    \caption{
    \textbf{Temporal changes in the estimated $\a$ for several ML conferences and \textit{Nature Portfolio} journals.} The estimated $\a$ for all ML conferences increases sharply after the release of ChatGPT (denoted by the dotted vertical line), 
    indicating that LLMs are being used in a small but significant way. Conversely, the $\a$ estimates for \textit{Nature Portfolio} reviews do not exhibit a significant increase or rise above the margin of error in our validation experiments for $\a=0$.
    See {Supp. Table~\ref{tab:main-result},\ref{tab: Nature trend}} for full results.}
    \label{fig: temporal}
\end{figure}

Next, we address the main question of our case study: what fraction of conference review text was substantially modified by LLMs, beyond simple grammar and spell checking? 
We find that there was a significant increase in AI-generated sentences after the release of ChatGPT for the ML venues, but not for \textit{Nature}(Appendix~\ref{sec:main-results}). 
The results are demonstrated in Figure~\ref{fig: temporal}, with error bars showing 95\% confidence intervals over 30,000 bootstrap samples.


Across all major ML conferences (\textit{NeurIPS}, \textit{CoRL}, and \textit{ICLR}), there was a sharp increase in the estimated $\alpha$ following the release of ChatGPT in late November 2022 (Figure~\ref{fig: temporal}). 
For instance, among the conferences with pre- and post-ChatGPT data, \textit{ICLR} experienced the most significant increase in estimated $\alpha$, from 1.6\% to 10.6\% (Figure~\ref{fig: temporal}, purple curve). 
\textit{NeurIPS} had a slightly lesser increase, from 1.9\% to 9.1\% (Figure~\ref{fig: temporal}, green curve), while \textit{CoRL}'s increase was the smallest, from 2.4\% to 6.5\% (Figure~\ref{fig: temporal}, red curve). Although data for \textit{EMNLP} reviews prior to ChatGPT's release are unavailable, this conference exhibited the highest estimated $\alpha$, at approximately 16.9\% (Figure~\ref{fig: temporal}, orange dot). This is perhaps unsurprising: NLP specialists may have had more exposure and knowledge of LLMs in the early days of its release. 

It should be noted that all of the post-ChatGPT $\a$ levels are significantly higher than the $\a$ estimated in the validation experiments with ground truth $\a=0$, and for \textit{ICLR} and \textit{NeurIPS}, the estimates are significantly higher than the validation estimates with ground truth $\a=5\%$. 
This suggests a modest yet noteworthy use of AI text-generation tools in conference review corpora.

\paragraph{Results on \textit{Nature Portfolio} journals}
We also train a separate model for \textit{Nature Portfolio} journals and validated its accuracy (Figure~\ref{fig: val}, \textit{Nature Portfolio} '22, Supp. Table~\ref{tab:verification-adj-main-nature}). 
Contrary to the ML conferences, the \textit{Nature Portfolio} journals do not exhibit a significant increase in the estimated $\alpha$ values following ChatGPT's release, with pre- and post-release $\alpha$ estimates remaining within the margin of error for the $\alpha=0$ validation experiment (Figure~\ref{fig: temporal}).
This consistency indicates a different response to AI tools within the broader scientific disciplines when compared to the specialized field of machine learning.

\subsection{Robustness to Proofreading}
\label{subsec:Proofreading}
\begin{figure}[ht!] 
\centering
\includegraphics[width=0.475\textwidth]{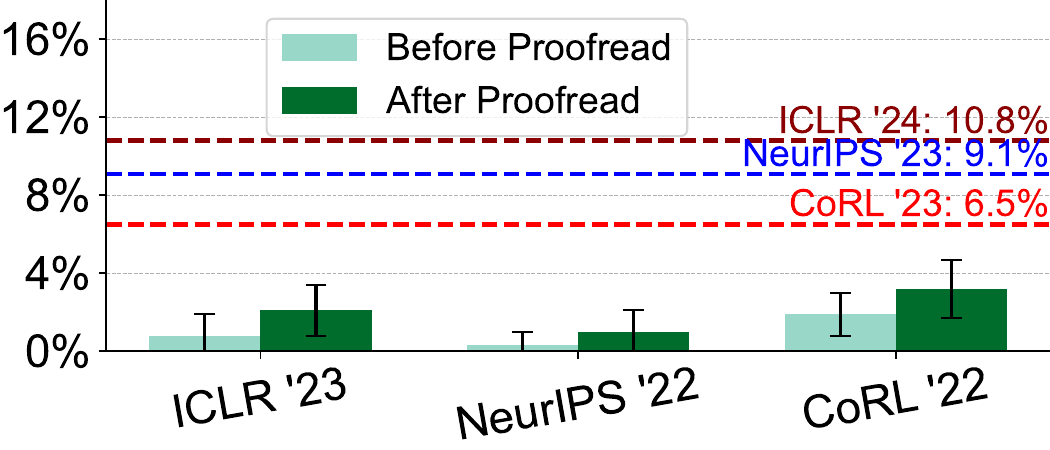}
\caption{
\textbf{Robustness of the estimations to proofreading.} 
Evaluating $\alpha$ after using LLMs for ``proof-reading" (non-substantial editing) of peer reviews shows a minor, non-significant increase across conferences, confirming our method's sensitivity to text which was generated in significant part by LLMs, beyond simple proofreading. See {Supp. Table~\ref{app: proofread}} for full results.
}
\label{fig: proofread}
\end{figure}

To verify that our method is detecting text which has been substantially modified by AI beyond simple grammatical edits, we conduct a robustness check by applying the method to peer reviews which were simply edited by ChatGPT for typos and grammar. The results are shown in Figure~\ref{fig: proofread}. While there is a slight increase in the estimated $\hat{\a}$, it is much smaller than the effect size seen in the real review corpus in the previous section (denoted with dashed lines in the figure).

\subsection{
Using LLMs to Substantially Expand Review Outline
}
\label{subsec:expand}

A reviewer might draft their review in two distinct stages: initially creating a brief outline of the review while reading the paper, followed by using LLMs to expand this outline into a detailed, comprehensive review. Consequently, we conduct an analysis to assess our algorithm's ability to detect such LLM usage.

To simulate this two-stage process retrospectively, we first condense a complete peer review into a structured, concise skeleton (outline) of key points (see Supp. Table~\ref{fig:skeleton-prompt-1}). Subsequently, rather than directly querying an LLM to generate feedback from papers, we instruct it to expand the skeleton into detailed, complete review feedback (see Supp. Table~\ref{fig:skeleton-prompt-2}). This mimics the two-stage scenario above.

\begin{figure}[ht!]
    \centering
    \includegraphics[width=0.475\textwidth]{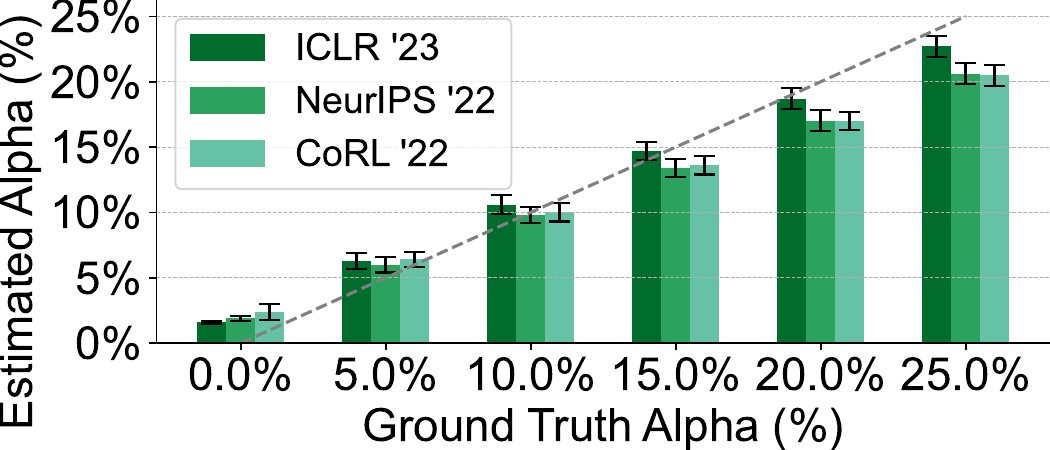}
\caption{
\textbf{Substantial modification and expansion of incomplete sentences using LLMs can largely account for the observed trend}. 
Rather than directly using LLMs to generate feedback, we expand a bullet-pointed skeleton of incomplete sentences into a full review using LLMs (see Supp. Table~\ref{fig:skeleton-prompt-1} and \ref{fig:skeleton-prompt-2} for prompts). The detected $\alpha$ may largely be attributed to this expansion. See {Supp. Table~\ref{tab: expand val}} for full results.
}
\label{fig:expand-verfication}
\end{figure}

We mix human peer reviews with the LLM-expanded feedback at various ground truth levels of $\alpha$, using our algorithm to predict these $\alpha$ values (Section $\S$~\ref{sec: val}). The results are presented in Figure~\ref{fig:expand-verfication}. The $\alpha$ estimated by our algorithm closely matches the ground truth $\alpha$. This suggests that our algorithm is sufficiently sensitive to detect the LLM use case of substantially expanding human-provided review outlines. The estimated $\alpha$ from our approach is consistent with reviewers using LLM to substantially expand their bullet points into full reviews.

\subsection{Factors that Correlate With Estimated LLM Usage}
\label{subsec:fine-grained-analysis}

\begin{figure}[ht!] 
\centering
\includegraphics[width=0.475\textwidth]{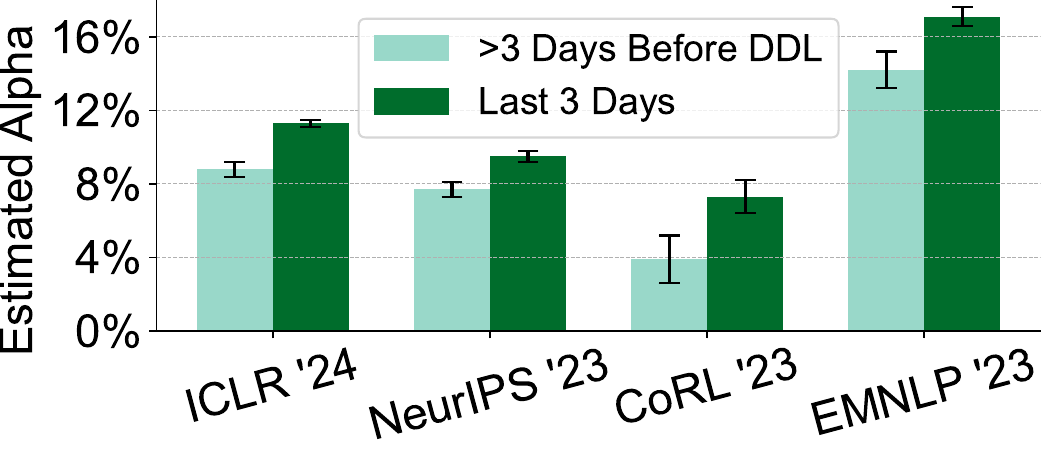}
\caption{
\textbf{The deadline effect.} Reviews submitted within 3 days of the review deadline tended to have a higher estimated $\a$. See {Supp. Table~\ref{app:timeline}} for full results.
}
\label{fig: deadline}
\end{figure}

\paragraph{Deadline Effect}
We see a small but consistent increase in the estimated $\a$ for reviews submitted 3 or fewer days before a deadline (Figure~\ref{fig: deadline}). As reviewers get closer to a looming deadline, they may try to save time by relying on LLMs. The following paragraphs explore some implications of this increased reliance.

\paragraph{Reference Effect}

\begin{figure}[ht!] 
\centering
\includegraphics[width=0.475\textwidth]{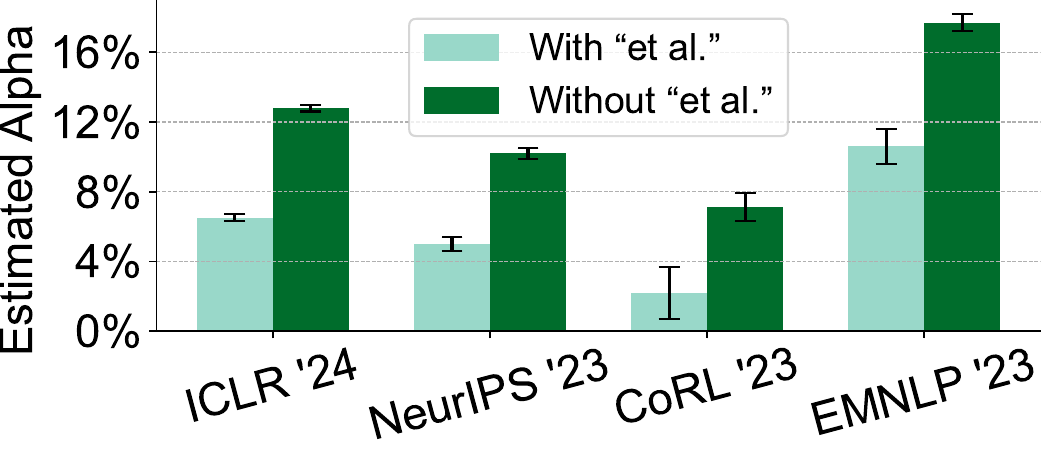}
\caption{
\textbf{The reference effect.} 
Our analysis demonstrates that reviews containing the term “et al.”, indicative of scholarly citations, are associated with a significantly lower estimated $\alpha$. 
See {Supp. Table~\ref{app:refere}} for full results.
}
\label{fig: et-al}
\end{figure}

Recognizing that LLMs often fail to accurately generate content and are less likely to include scholarly citations, as highlighted by recent studies~\cite{LLM-Research-Feedback-2023,walters2023fabrication}, we hypothesize that reviews containing scholarly citations might indicate lower LLM usage. To test this, we use the occurrence of the string “et al.” as a proxy for scholarly citations in reviews. We find that reviews featuring “et al.” consistently showed a lower estimated $\alpha$ than those lacking such references (see Figure~\ref{fig: et-al}). The lack of scholarly citations demonstrates one way that generated text does not include content that expert reviewers otherwise might. However, we lack a counterfactual- it could be that people who were more likely to use ChatGPT may also have been less likely to cite sources were ChatGPT not available. Future studies should examine the causal structure of this relationship.

\paragraph{Lower Reply Rate Effect}
We find a negative correlation between the number of author replies and estimated ChatGPT usage ($\a$), suggesting that authors who participated more actively in the discussion period were less likely to use ChatGPT to generate their reviews. There are a number of possible explanations, but we cannot make a causal claim. Reviewers may use LLMs as a quick-fix to avoid extra engagement, but if the role of the reviewer is to be a co-producer of better science, then this fix hinders that role. Alternatively, as AI conferences face a desperate shortage of reviewers, scholars may agree to participate in more reviews and rely on the tool to support the increased workload. Editors and conference organizers should carefully consider the relationship between ChatGPT-use and reply rate to ensure each paper receives an adequate level of feedback.
\begin{figure}[htb]
    \centering
    \begin{minipage}{0.23\textwidth}
        \centering
        \begin{overpic}[width=\textwidth]{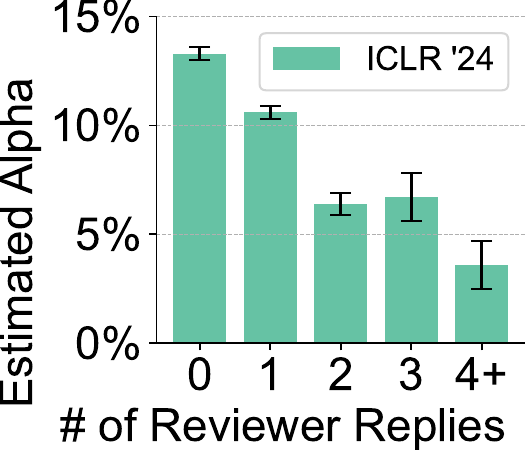}
            \put(-5,90){\textbf{(a)}} 
        \end{overpic}        
    \end{minipage}\hfill
    \begin{minipage}{0.23\textwidth}
        \centering
        \begin{overpic}[width=\textwidth]{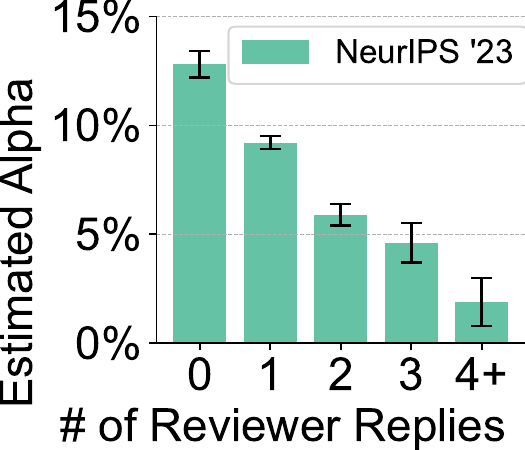}
            \put(-5,90){\textbf{(b)}} 
        \end{overpic}
    \end{minipage}
\caption{
\textbf{The lower reply rate effect.} We observe a negative correlation between number of reviewer replies in the review discussion period and the estimated $\a$ on these reviews. See {Supp. Table~\ref{app:replies}} for full results.
}
\label{fig: reply}
\end{figure}


\paragraph{Homogenization Effect}
There is growing evidence that the introduction of LLM content in information ecosystems can contribute to to \textit{output homogenization} 
\cite{liu2024chatgpt, bommasani2022picking, kleinberg2021algorithmic}. We examine this phenomenon in the context of text as a decrease in variation of linguistic features and epistemic content than would be expected in an unpolluted corpus \cite{christin2020data}. While it might be intuitive to expect that a standardization of text in peer reviews could be useful, empirical social studies of peer review demonstrate the important role of feedback variation from reviewers \cite{teplitskiy2018sociology, lamont2009professors, lamont2012toward, longino1990science, sulik2023scientists}. 

Here, we explore whether the presence of generated texts in a peer review corpus led to homogenization of feedback, using a new method to classify texts as ``convergent'' (similar to the other reviews) or ``divergent'' (dissimilar to the other reviews). 
For each paper, we obtained the OpenAI's text-embeddings for all reviews, followed by the calculation of their centroid (average). 
Among the assigned reviews, the one with its embedding closest to the centroid is labeled as convergent, and the one farthest as divergent. 
This process is repeated for each paper, generating a corpus of convergent and divergent reviews, to which we then apply our analysis method.

\begin{figure}[ht!] 
\centering
\includegraphics[width=0.475\textwidth]{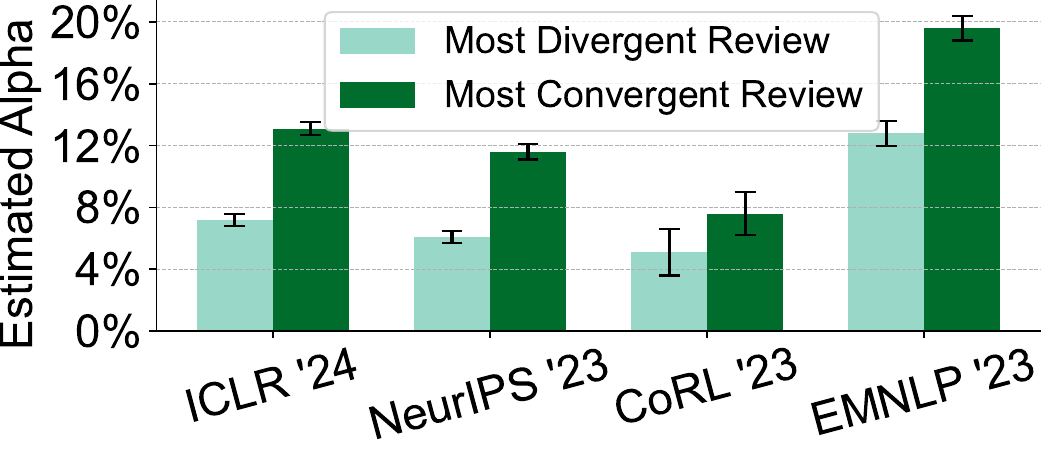}
\caption{
\textbf{The homogenization effect.} ``Convergent'' reviews (those most similar to other reviews of the same paper in the embedding space) tend to have a higher estimated $\a$ as compared to ``divergent'' reviews (those most dissimilar to other reviews).
See {Supp. Table~\ref{app:similarity}} for full results.
}
\label{fig: homog}
\end{figure}

The results, as shown in Figure~\ref{fig: homog}, suggest that convergent reviews, which align more closely with the centroid of review embeddings, tend to have a higher estimated $\alpha$. This finding aligns with previous observations that LLM-generated text often focuses on specific, recurring topics, such as research implications or suggestions for additional experiments, more consistently than expert peer reviewers do \cite{LLM-Research-Feedback-2023}. 

This corpus-level homogenization is potentially concerning for several reasons. First, if paper authors receive synthetically-generated text in place of an expert-written review, the scholars lose an opportunity to receive feedback from multiple, independent, diverse experts in their field. Instead, authors must contend with formulaic responses which may not capture the unique and creative ideas that a peer might present. Second, based on studies of representational harms in language model output, it is likely that this homogenization does not trend toward random, representative ways of knowing and producing language, but instead converges toward the practices of certain groups \cite{naous2024having, cao2023assessing, papadimitriou2023multilingual, arora2022probing, hofmann2024dialect}. 


\begin{figure}[t!] 
\centering
\includegraphics[width=0.475\textwidth]{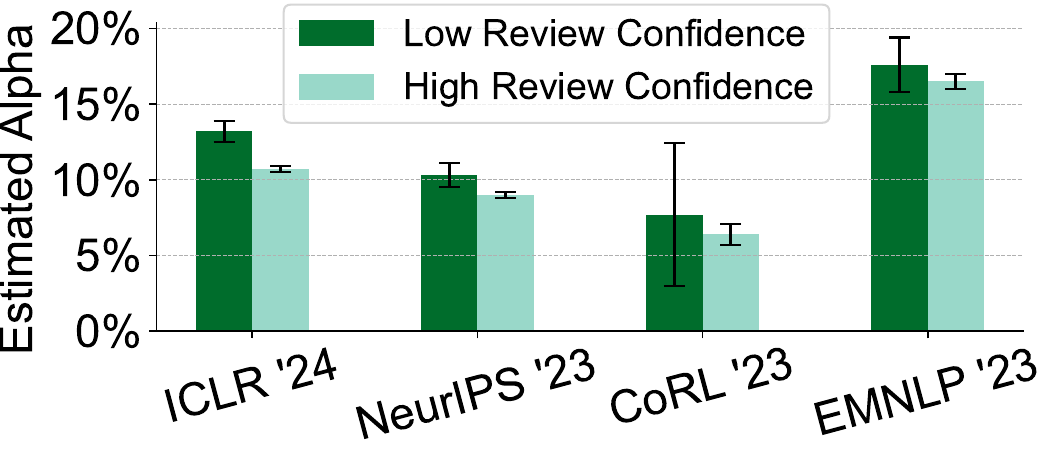}
\caption{
\textbf{The low confidence effect.} 
Reviews with low confidence, defined as self-rated confidence of 2 or lower on a 5-point scale, are correlated with higher alpha values than those with 3 or above, and are mostly identical across these major ML conferences. See the descriptions of the confidence rating scales in {Supp. Table~\ref{tab:appendix-confidence-scale}} and full results in {Supp. Table~\ref{app:confidence}}.
}
\label{fig: confidence}
\end{figure}

\paragraph{Low Confidence Effect} 
The correlation between reviewer confidence tends to be negatively correlated with ChatGPT usage -that is, the estimate for $\a$ (Figure~\ref{fig: confidence}). 
One possible interpretation of this phenomenon is that the integration of LMs into the review process introduces a layer of detachment for the reviewer from the generated content, which might make reviewers feel less personally invested or assured in the content's accuracy or relevance.

\section{Discussion}
In this work, we propose a method for estimating the fraction of documents in a large corpus which were generated primarily using AI tools. The method makes use of historical documents. The prompts from this historical corpus are then fed into an LLM (or LLMs) to produce a corresponding corpus of AI-generated texts. The written and AI-generated corpora are then used to estimate the distributions of AI-generated vs. written texts in a mixed corpus. Next, these estimated document distributions are used to compute the likelihood of the target corpus, and the estimate for $\a$ is produced by maximizing the likelihood. We also provide specific methods for estimating the text distributions by token frequency and occurrence, as well as a method for validating the performance of the system. 

Applying this method to conference and journal reviews written before and after the release of ChatGPT shows evidence that roughly 7-15\% of sentences in ML conference reviews were substantially modified by AI beyond a simple grammar check, while there does \textit{not} appear to be significant evidence of AI usage in reviews for \textit{Nature}. Finally, we demonstrate several ways this method can support social analysis. First, we show that reviewers are more likely to submit generated text for last-minute reviews, and that people who submit generated text offer fewer author replies than those who submit written reviews. Second, we show that generated texts include less specific feedback or citations of other work, in comparison to written reviews. Generated reviews also are associated with lower confidence ratings.
Third, we show how corpora with generated text appear to compress the linguistic variation and epistemic diversity that would be expected in unpolluted corpora. We should also note that other social concerns with ChatGPT presence in peer reviews extend beyond our scope, including the potential privacy and anonymity risks of providing unpublished work to a privately owned language model.
\paragraph{Limitations}

While our study focused on ChatGPT, which dominates the generative AI market with 76\% of global internet traffic in the category~\cite{vanrossum2024generative}, we acknowledge that there are other diverse LLMs used for generating or rephrasing text. However, recent studies have found that ChatGPT substantially outperforms other LLMs, including Bard, in the reviewing of scientific papers or proposals~\cite{liang2023can,liu2023reviewergpt}. 
We also found that our results are robust on the use of alternative LLMs such as GPT-3.5. For example, the model trained with only GPT-3.5 data provides consistent estimation results and findings, and demonstrates the ability to generalize, accurately detecting GPT-4 as well (see Supp. Table~\ref{table:GPT-3.5-validation} and \ref{table:GPT-3.5-validation-on-GPT-4}). 
However, we acknowledge that our framework's effectiveness may vary depending on the specific LLM used, and future practitioners should select the LLM that most closely mirrors the language model likely used to generate their target corpus, reflecting actual usage patterns at the time of creation.

Our findings are primarily based on datasets from major ML conferences (ICLR, NeurIPS, CoRL, EMNLP) and Nature Family Journals spanning 15 distinct journals across different disciplines such as medicine, biology, chemistry, and environmental sciences. While this demonstrates the applicability of our framework beyond these domains, further experimentation may be required to fully establish its generalizability to an even wider range of fields and publication venues. Factors such as field-specific writing styles and the prevalence of AI use could influence the effectiveness of our approach.

Moreover, the prompting techniques used in our study to simulate the process of revising, expanding, paraphrasing, and proofreading review texts (Section $\S$~\ref{subsec:Proofreading}) have limitations. The prompts we employed were designed based on our understanding of common practices, but they may not capture the full range of techniques used by reviewers or AI assistants. We emphasize that these techniques should be interpreted as a best-effort approximation rather than a definitive representation of how AI is used for review text modifications.

Although our validation experiments used real reviews from prior years, which included a significant fraction of non-native speaker-written texts, and our results remained accurate, we recognize that substantial shifts in the non-native speaker population over time could still impact the accuracy of our estimates~\cite{Liang2023GPTDA}. Future research should investigate the impact of evolving non-native speaker populations on the robustness of our framework.

In addition, the approximations made to the review generating process in Section $\S$~\ref{sec: method} in order to make estimation of the review likelihood tractable introduce an additional source of error, as does the temporal distribution shift in token frequencies due to, e.g., changes in topics, reviewers, etc. 

We emphasize here that we do not wish to pass a value judgement or claim that the use of AI tools for review papers is necessarily bad or good. 
We also do not claim (nor do we believe) that many reviewers are using ChatGPT to write entire reviews outright. Our method does not constitute direct evidence that reviewers are using ChatGPT to write reviews from scratch. For example, it is possible that a reviewer may sketch out several bullet points related to the paper and uses ChatGPT to formulate these bullet points into paragraphs. 
In this case, it is possible for the estimated $\alpha$ to be high; indeed our results in Appendix~\ref{subsec:expand} is consistent with this mode of using LLM to substantially modify and flesh out reviews. 

To enhance transparency and accountability, future work should focus on applying and extending our framework to estimate the extent of AI-generated text across various domains, including but not limited to peer review. We believe that our data and analyses can serve as a foundation for constructive discussions and further research by the community, ultimately contributing to the development of robust guidelines and best practices for the ethical use of generative AI.

\section*{Code Availability}
The code can be accessed at \href{https://github.com/Weixin-Liang/Mapping-the-Increasing-Use-of-LLMs-in-Scientific-Papers}{\texttt{https://github.com/}}\\\href{https://github.com/Weixin-Liang/Mapping-the-Increasing-Use-of-LLMs-in-Scientific-Papers}{\texttt{Weixin-Liang/Mapping-the-Increasing-Use}}\\\href{https://github.com/Weixin-Liang/Mapping-the-Increasing-Use-of-LLMs-in-Scientific-Papers}{\texttt{-of-LLMs-in-Scientific-Papers}} .

\section*{Impact Statement}
This work offers a method for the study of LLM use at scale. We apply this method on several corpora of peer reviews, demonstrating the potential ramifications of such use to scientific publishing. While our study has several limitations that we acknowledge throughout the manuscript, we believe that there is still value in providing transparent analysis of LLM use in the scientific community. We hope that our statistical analysis will inspire further social analysis, productive community reflection, and informed policy decisions about the extent and effects of LLM use in information ecosystems.

\subsubsection*{Acknowledgments}
We thank Christopher D. Manning, Dan Jurafsky, Diyi Yang, Yongchan Kwon, Federico Bianchi, and Mert Yuksekgonul for their helpful comments and discussions. J.Z. is supported by the National Science Foundation (CAREER 1942926) and grants from the Chan-Zuckerberg Initiative and Stanford HAI. D.M. and H.L. are supported by the National Science Foundation (2244804 and 2022435) and the Stanford Institute for Human-Centered Artificial Intelligence (HAI).

\bibliographystyle{icml2024}
\bibliography{main_paper}

@misc{vanrossum2024generative,
  author = {Van Rossum, Dann.},
  title = {{Generative AI Top 150: The World's Most Used AI Tools}},
  year = {2024},
  month = {February},
  howpublished = {\url{https://www.flexos.work/learn/generative-ai-top-150}},
  publisher = {FlexOS}
}

@article{liu2023reviewergpt,
  title={{Reviewergpt? an exploratory study on using large language models for paper reviewing}},
  author={Liu, Ryan and Shah, Nihar B},
  journal={arXiv preprint arXiv:2306.00622},
  year={2023}
}

@inproceedings{dycke-etal-2022-yes,
    title = "Yes-Yes-Yes: Proactive Data Collection for {ACL} Rolling Review and Beyond",
    author = "Dycke, Nils  and
      Kuznetsov, Ilia  and
      Gurevych, Iryna",
    editor = "Goldberg, Yoav  and
      Kozareva, Zornitsa  and
      Zhang, Yue",
    booktitle = "Findings of the Association for Computational Linguistics: EMNLP 2022",
    month = dec,
    year = "2022",
    address = "Abu Dhabi, United Arab Emirates",
    publisher = "Association for Computational Linguistics",
    url = "https://aclanthology.org/2022.findings-emnlp.23",
    doi = "10.18653/v1/2022.findings-emnlp.23",
    pages = "300--318",
}

@inproceedings{dycke-etal-2023-nlpeer,
    title = "{NLP}eer: A Unified Resource for the Computational Study of Peer Review",
    author = "Dycke, Nils  and
      Kuznetsov, Ilia  and
      Gurevych, Iryna",
    editor = "Rogers, Anna  and
      Boyd-Graber, Jordan  and
      Okazaki, Naoaki",
    booktitle = "Proceedings of the 61st Annual Meeting of the Association for Computational Linguistics (Volume 1: Long Papers)",
    month = jul,
    year = "2023",
    address = "Toronto, Canada",
    publisher = "Association for Computational Linguistics",
    url = "https://aclanthology.org/2023.acl-long.277",
    doi = "10.18653/v1/2023.acl-long.277",
    pages = "5049--5073",
}

@article{lin2023unlocking,
  title={{The unlocking spell on base llms: Rethinking alignment via in-context learning}},
  author={Lin, Bill Yuchen and Ravichander, Abhilasha and Lu, Ximing and Dziri, Nouha and Sclar, Melanie and Chandu, Khyathi and Bhagavatula, Chandra and Choi, Yejin},
  journal={arXiv preprint arXiv:2312.01552},
  year={2023}
}

@book{lamont2009professors,
  title={How professors think: Inside the curious world of academic judgment},
  author={Lamont, Mich{\`e}le},
  year={2009},
  publisher={Harvard University Press}
}

@article{lamont2012toward,
  title={Toward a comparative sociology of valuation and evaluation},
  author={Lamont, Mich{\`e}le},
  journal={Annual review of sociology},
  volume={38},
  pages={201--221},
  year={2012},
  publisher={Annual Reviews}
}

@misc{cao2023assessing,
      title={Assessing Cross-Cultural Alignment between ChatGPT and Human Societies: An Empirical Study}, 
      author={Yong Cao and Li Zhou and Seolhwa Lee and Laura Cabello and Min Chen and Daniel Hershcovich},
      year={2023},
      eprint={2303.17466},
      archivePrefix={arXiv},
      primaryClass={cs.CL}
}

@article{christin2020data,
  title={What data can do: A typology of mechanisms},
  author={Christin, Ang{\`e}le},
  journal={International Journal of Communication},
  volume={14},
  pages={20},
  year={2020}
}

@article{arora2022probing,
  title={Probing pre-trained language models for cross-cultural differences in values},
  author={Arora, Arnav and Kaffee, Lucie-Aim{\'e}e and Augenstein, Isabelle},
  journal={arXiv preprint arXiv:2203.13722},
  year={2022}
}

@misc{papadimitriou2023multilingual,
      title={Multilingual BERT has an accent: Evaluating English influences on fluency in multilingual models}, 
      author={Isabel Papadimitriou and Kezia Lopez and Dan Jurafsky},
      year={2023},
      eprint={2210.05619},
      archivePrefix={arXiv},
      primaryClass={cs.CL}
}

@misc{liu2024chatgpt,
      title={When ChatGPT is gone: Creativity reverts and homogeneity persists}, 
      author={Qinghan Liu and Yiyong Zhou and Jihao Huang and Guiquan Li},
      year={2024},
      eprint={2401.06816},
      archivePrefix={arXiv},
      primaryClass={cs.CL}
}

@article{sulik2023scientists,
  title={Why Do Scientists Disagree?},
  author={Sulik, Justin and Rim, Nakwon and Pontikes, Elizabeth and Evans, James and Lupyan, Gary},
  journal={PsyArXiv},
  year={2023},
  publisher={PsyArXiv},
  note={Preprint},
}

@book{longino1990science,
  title={Science as social knowledge: Values and objectivity in scientific inquiry},
  author={Longino, Helen E},
  year={1990},
  publisher={Princeton university press}
}

@article{teplitskiy2018sociology,
  title={The sociology of scientific validity: How professional networks shape judgement in peer review},
  author={Teplitskiy, Misha and Acuna, Daniel and Elamrani-Raoult, A{\"\i}da and K{\"o}rding, Konrad and Evans, James},
  journal={Research Policy},
  volume={47},
  number={9},
  pages={1825--1841},
  year={2018},
  publisher={Elsevier}
}

@article{kleinberg2021algorithmic,
  title={Algorithmic monoculture and social welfare},
  author={Kleinberg, Jon and Raghavan, Manish},
  journal={Proceedings of the National Academy of Sciences},
  volume={118},
  number={22},
  pages={e2018340118},
  year={2021},
  publisher={National Acad Sciences}
}

@inproceedings{LLM-Research-Feedback-2023,
  title={{Can large language models provide useful feedback on research papers? A large-scale empirical analysis}},
  author={Liang, Weixin and Zhang, Yuhui and Cao, Hancheng and Wang, Binglu and Ding, Daisy and Yang, Xinyu and Vodrahalli, Kailas and He, Siyu and Smith, Daniel and Yin, Yian and McFarland, Daniel and Zou, James},
  booktitle={arXiv preprint arXiv:2310.01783},
  year={2023}
}

@article{walters2023fabrication,
  title={{Fabrication and errors in the bibliographic citations generated by ChatGPT}},
  author={Walters, William H and Wilder, Esther Isabelle},
  journal={Scientific Reports},
  volume={13},
  number={1},
  pages={14045},
  year={2023},
  publisher={Nature Publishing Group UK London}
}

@article{Nature-news-Abstracts-written-by-ChatGPT-fool-scientists,
  title={{Abstracts written by ChatGPT fool scientists}},
  author={Else, Holly},
  journal={Nature},
  year={2023},
  month={Jan},
  day={12},
  url={https://www.nature.com/articles/d41586-023-00056-7}
}

@article{Abstracts-written-by-ChatGPT-fool-scientists,
  title={{Comparing scientific abstracts generated by ChatGPT to original abstracts using an artificial intelligence output detector, plagiarism detector, and blinded human reviewers}},
  author={Gao, Catherine A and Howard, Frederick M and Markov, Nikolay S and Dyer, Emma C and Ramesh, Siddhi and Luo, Yuan and Pearson, Alexander T},
  journal={bioRxiv},
  pages={2022--12},
  year={2022},
  publisher={Cold Spring Harbor Laboratory}
}

@article{bommasani2022picking,
  title={Picking on the Same Person: Does Algorithmic Monoculture lead to Outcome Homogenization?},
  author={Bommasani, Rishi and Creel, Kathleen A and Kumar, Ananya and Jurafsky, Dan and Liang, Percy S},
  journal={Advances in Neural Information Processing Systems},
  volume={35},
  pages={3663--3678},
  year={2022}
}

@article{messeri,
title = {Artificial intelligence and illusions ofunderstanding in scientific research},
author={Lisa Messeri and M. J. Crockett},
journal = {Nature},
volume={627}, 
pages={49–58},
year={2024}
}

@misc{naous2024having,
      title={Having Beer after Prayer? Measuring Cultural Bias in Large Language Models}, 
      author={Tarek Naous and Michael J. Ryan and Alan Ritter and Wei Xu},
      year={2024},
      eprint={2305.14456},
      archivePrefix={arXiv},
      primaryClass={cs.CL}
}

@article{van2023ai,
  title={AI and science: what 1,600 researchers think},
  author={Van Noorden, Richard and Perkel, Jeffrey M},
  journal={Nature},
  volume={621},
  number={7980},
  pages={672--675},
  year={2023},
  publisher={Nature}
}

@article{bearman2023discourses,
  title={Discourses of artificial intelligence in higher education: A critical literature review},
  author={Bearman, Margaret and Ryan, Juliana and Ajjawi, Rola},
  journal={Higher Education},
  volume={86},
  number={2},
  pages={369--385},
  year={2023},
  publisher={Springer}
}

@misc{hofmann2024dialect,
      title={{Dialect prejudice predicts AI decisions about people's character, employability, and criminality}}, 
      author={Valentin Hofmann and Pratyusha Ria Kalluri and Dan Jurafsky and Sharese King},
      year={2024},
      eprint={2403.00742},
      archivePrefix={arXiv},
      primaryClass={cs.CL}
}

@article{fake-news-66,
title={All the News That's Fit to Fabricate: AI-Generated Text as a Tool of Media Misinformation},
author={Kreps, Sarah and McCain, R and Brundage, Miles},
journal={Journal of Experimental Political Science},
volume={9},
number={1},
pages={104--117},
year={2022},
publisher={Cambridge University Press},
doi={10.1017/XPS.2020.37}
}

@inproceedings{huaman-detect-gpt3,
  title={{All That’s ‘Human’Is Not Gold: Evaluating Human Evaluation of Generated Text}},
  author={Clark, Elizabeth and August, Tal and Serrano, Sofia and Haduong, Nikita and Gururangan, Suchin and Smith, Noah A},
  booktitle={Proceedings of the 59th Annual Meeting of the Association for Computational Linguistics and the 11th International Joint Conference on Natural Language Processing (Volume 1: Long Papers)},
  pages={7282--7296},
  year={2021}
}

@article{mit-technology-review-how-to-spot-ai-generated-text,
title={How to spot AI-generated text},
author={Heikkil{\"a}, Melissa},
journal={MIT Technology Review},
year={2022},
month={Dec},
day={19},
url={https://www.technologyreview.com/2022/12/19/1065596/how-to-spot-ai-generated-text/}
}

@article{survey-2023,
  title={{Machine Generated Text: A Comprehensive Survey of Threat Models and Detection Methods}},
  author={Crothers, Evan and Japkowicz, Nathalie and Viktor, Herna},
  journal={arXiv preprint arXiv:2210.07321},
  year={2022}
}

@article{fagni2021tweepfake,
  title={{TweepFake: About detecting deepfake tweets}},
  author={Fagni, Tiziano and Falchi, Fabrizio and Gambini, Margherita and Martella, Antonio and Tesconi, Maurizio},
  journal={Plos one},
  volume={16},
  number={5},
  pages={e0251415},
  year={2021},
  publisher={Public Library of Science San Francisco, CA USA}
}

@misc{OpenAIGPT2,
  author = {OpenAI},
  title = {{GPT-2: 1.5B release}},
  year = {2019},
  howpublished = {\url{https://openai.com/research/gpt-2-1-5b-release}},
  note = {Accessed: 2019-11-05}
}

@article{jawahar2020automatic,
  title={{Automatic detection of machine generated text: A critical survey}},
  author={Jawahar, Ganesh and Abdul-Mageed, Muhammad and Lakshmanan, Laks VS},
  journal={arXiv preprint arXiv:2011.01314},
  year={2020}
}

@article{solaiman2019release,
  title={{Release strategies and the social impacts of language models}},
  author={Solaiman, Irene and Brundage, Miles and Clark, Jack and Askell, Amanda and Herbert-Voss, Ariel and Wu, Jeff and Radford, Alec and Krueger, Gretchen and Kim, Jong Wook and Kreps, Sarah and others},
  journal={arXiv preprint arXiv:1908.09203},
  year={2019}
}

@article{mitchell2023detectgpt,
  title={{DetectGPT}: Zero-shot machine-generated text detection using probability curvature},
  author={Mitchell, Eric and Lee, Yoonho and Khazatsky, Alexander and Manning, Christopher D and Finn, Chelsea},
  journal={arXiv preprint arXiv:2301.11305},
  year={2023}
}

@article{ippolito2019automatic,
  title={{Automatic detection of generated text is easiest when humans are fooled}},
  author={Ippolito, Daphne and Duckworth, Daniel and Callison-Burch, Chris and Eck, Douglas},
  journal={arXiv preprint arXiv:1911.00650},
  year={2019}
}

@inproceedings{human-hard-to-detect-generated-text,
  title={{GLTR: Statistical Detection and Visualization of Generated Text}},
  author={Gehrmann, Sebastian and Strobelt, Hendrik and Rush, Alexander M},
  booktitle={Proceedings of the 57th Annual Meeting of the Association for Computational Linguistics: System Demonstrations},
  pages={111--116},
  year={2019}
}

@misc{Kirchner2023,
  title={{New AI classifier for indicating AI-written text}},
  author={Jan Hendrik Kirchner and Lama Ahmad and Scott Aaronson and Jan Leike},
  year={2023},
  note = {{OpenAI}},
}

@article{Kelly2023,
  title={{ChatGPT creator pulls AI detection tool due to ‘low rate of accuracy’}},
  author={Samantha Murphy Kelly},
  journal={CNN Business},
  year={2023},
  month={Jul},
  day={25},
  url={https://www.cnn.com/2023/07/25/tech/openai-ai-detection-tool/index.html},
}

@misc{Cantor2023,
  author = {Matthew Cantor},
  title = {{Nearly 50 news websites are ‘AI-generated’, a study says. Would I be able to tell?}},
  year = {2023},
  url = {https://www.theguardian.com/technology/2023/may/08/ai-generated-news-websites-study},
  note = {Accessed: 2024-02-24},
  urldate = {2023-05-08}
}

@misc{NewsGuard2023,
  author = {{NewsGuard}},
  title = {{Tracking AI-enabled Misinformation: 713 ‘Unreliable AI-Generated News’ Websites (and Counting), Plus the Top False Narratives Generated by Artificial Intelligence Tools}},
  year = {2023},
  url = {https://www.newsguardtech.com/special-reports/ai-tracking-center/},
  note = {Accessed: 2024-02-24}
}

@article{chakraborty2023possibilities,
  title={{On the possibilities of ai-generated text detection}},
  author={Chakraborty, Souradip and Bedi, Amrit Singh and Zhu, Sicheng and An, Bang and Manocha, Dinesh and Huang, Furong},
  journal={arXiv preprint arXiv:2304.04736},
  year={2023}
}

@article{Weber-Wulff2023,
  author    = {Weber-Wulff, Debora and Anohina-Naumeca, Alla and Bjelobaba, Sonja and Foltýnek, Tomáš and Guerrero-Dib, Jean and Popoola, Olumide and Šigut, Petr and Waddington, Lorna},
  title     = {{Testing of detection tools for AI-generated text}},
  journal   = {International Journal for Educational Integrity},
  volume    = {19},
  number    = {1},
  pages     = {26},
  year      = {2023},
  doi       = {10.1007/s40979-023-00146-z},
  url       = {https://doi.org/10.1007/s40979-023-00146-z},
  issn      = {1833-2595},
  date      = {2023/12/25},
}

@article{kirchenbauer2023watermark,
  title={{A watermark for large language models}},
  author={Kirchenbauer, John and Geiping, Jonas and Wen, Yuxin and Katz, Jonathan and Miers, Ian and Goldstein, Tom},
  journal={International Conference on Machine Learning},
  year={2023}
}

@article{Mitchell2023DetectGPTZM,
  title={{DetectGPT: Zero-Shot Machine-Generated Text Detection using Probability Curvature}},
  author={Eric Mitchell and Yoonho Lee and Alexander Khazatsky and Christopher D. Manning and Chelsea Finn},
  journal={ArXiv},
  year={2023},
  volume={abs/2301.11305}
}

@article{Sadasivan2023CanAT,
  title={{Can AI-Generated Text be Reliably Detected?}},
  author={Vinu Sankar Sadasivan and Aounon Kumar and S. Balasubramanian and Wenxiao Wang and Soheil Feizi},
  journal={ArXiv},
  year={2023},
  volume={abs/2303.11156}
}

@article{Liu2019RoBERTaAR,
  title={{RoBERTa: A Robustly Optimized BERT Pretraining Approach}},
  author={Yinhan Liu and Myle Ott and Naman Goyal and Jingfei Du and Mandar Joshi and Danqi Chen and Omer Levy and Mike Lewis and Luke Zettlemoyer and Veselin Stoyanov},
  journal={ArXiv},
  year={2019},
  volume={abs/1907.11692}
}

@article{Liang2023GPTDA,
  title={{GPT detectors are biased against non-native English writers}},
  author={Weixin Liang and Mert Yuksekgonul and Yining Mao and Eric Wu and James Y. Zou},
  journal={ArXiv},
  year={2023},
  volume={abs/2304.02819}
}

@article{liang2023can,
  title={{Can large language models provide useful feedback on research papers? A large-scale empirical analysis}},
  author={Liang, Weixin and Zhang, Yuhui and Cao, Hancheng and Wang, Binglu and Ding, Daisy and Yang, Xinyu and Vodrahalli, Kailas and He, Siyu and Smith, Daniel and Yin, Yian and others},
  journal={arXiv preprint arXiv:2310.01783},
  year={2023}
}

@inproceedings{Topkara2006TheHV,
  title={{The hiding virtues of ambiguity: quantifiably resilient watermarking of natural language text through synonym substitutions}},
  author={Umut Topkara and Mercan Topkara and Mikhail J. Atallah},
  booktitle={Workshop on Multimedia \& Security},
  year={2006}
}

@inproceedings{Atallah2001NaturalLW,
  title={{{Natural Language Watermarking: Design, Analysis, and a Proof-of-Concept Implementation}}},
  author={Mikhail J. Atallah and Victor Raskin and Michael Crogan and Christian F. Hempelmann and Florian Kerschbaum and Dina Mohamed and Sanket Naik},
  booktitle={Information Hiding},
  year={2001}
}

@inproceedings{Atallah2002NaturalLW,
  title={{Natural Language Watermarking and Tamperproofing}},
  author={Mikhail J. Atallah and Victor Raskin and Christian F. Hempelmann and Mercan Topkara and Radu Sion and Umut Topkara and Katrina E. Triezenberg},
  booktitle={Information Hiding},
  year={2002}
}

@article{Wolff2020AttackingNT,
  title={{Attacking Neural Text Detectors}},
  author={Max Wolff},
  journal={ArXiv},
  year={2020},
  volume={abs/2002.11768}
}

@article{raffel2020exploring,
  title={{Exploring the limits of transfer learning with a unified text-to-text transformer}},
  author={Raffel, Colin and Shazeer, Noam and Roberts, Adam and Lee, Katherine and Narang, Sharan and Matena, Michael and Zhou, Yanqi and Li, Wei and Liu, Peter J},
  journal={The Journal of Machine Learning Research},
  volume={21},
  number={1},
  pages={5485--5551},
  year={2020},
  publisher={JMLRORG}
}

@inproceedings{Chiang2003NaturalLW,
  title={{Natural Language Watermarking Using Semantic Substitution for Chinese Text}},
  author={Yuei-Lin Chiang and Lu-Ping Chang and Wen-Tai Hsieh and Wen-Chih Chen},
  booktitle={International Workshop on Digital Watermarking},
  year={2003},
  url={https://api.semanticscholar.org/CorpusID:40971354}
}

@inproceedings{Topkara2006NaturalLW,
  title={{Natural language watermarking: challenges in building a practical system}},
  author={Mercan Topkara and Giuseppe Riccardi and Dilek Z. Hakkani-T{\"u}r and Mikhail J. Atallah},
  booktitle={Electronic imaging},
  year={2006},
  url={https://api.semanticscholar.org/CorpusID:16650373}
}

@article{Zhao2023ProvableRW,
  title={{Provable Robust Watermarking for AI-Generated Text}},
  author={Xuandong Zhao and Prabhanjan Vijendra Ananth and Lei Li and Yu-Xiang Wang},
  journal={ArXiv},
  year={2023},
  volume={abs/2306.17439},
  url={https://api.semanticscholar.org/CorpusID:259308864}
}

@article{Kuditipudi2023RobustDW,
  title={{Robust Distortion-free Watermarks for Language Models}},
  author={Rohith Kuditipudi and John Thickstun and Tatsunori Hashimoto and Percy Liang},
  journal={ArXiv},
  year={2023},
  volume={abs/2307.15593},
  url={https://api.semanticscholar.org/CorpusID:260315804}
}

@article{Hu2023UnbiasedWF,
  title={{Unbiased Watermark for Large Language Models}},
  author={Zhengmian Hu and Lichang Chen and Xidong Wu and Yihan Wu and Hongyang Zhang and Heng Huang},
  journal={ArXiv},
  year={2023},
  volume={abs/2310.10669},
  url={https://api.semanticscholar.org/CorpusID:264172471}
}

@article{Wu2023DiPmarkAS,
  title={{DiPmark: A Stealthy, Efficient and Resilient Watermark for Large Language Models}},
  author={Yihan Wu and Zhengmian Hu and Hongyang Zhang and Heng Huang},
  journal={ArXiv},
  year={2023},
  volume={abs/2310.07710},
  url={https://api.semanticscholar.org/CorpusID:263834753}
}

@article{Hou2023SemStampAS,
  title={{SemStamp: A Semantic Watermark with Paraphrastic Robustness for Text Generation}},
  author={Abe Bohan Hou and Jingyu Zhang and Tianxing He and Yichen Wang and Yung-Sung Chuang and Hongwei Wang and Lingfeng Shen and Benjamin Van Durme and Daniel Khashabi and Yulia Tsvetkov},
  journal={ArXiv},
  year={2023},
  volume={abs/2310.03991},
  url={https://api.semanticscholar.org/CorpusID:263831179}
}

@article{Fu2023WatermarkingCT,
  title={{Watermarking Conditional Text Generation for AI Detection: Unveiling Challenges and a Semantic-Aware Watermark Remedy}},
  author={Yu Fu and Deyi Xiong and Yue Dong},
  journal={ArXiv},
  year={2023},
  volume={abs/2307.13808},
  url={https://api.semanticscholar.org/CorpusID:260164516}
}

@article{Liu2023ASI,
  title={{A Semantic Invariant Robust Watermark for Large Language Models}},
  author={Aiwei Liu and Leyi Pan and Xuming Hu and Shiao Meng and Lijie Wen},
  journal={ArXiv},
  year={2023},
  volume={abs/2310.06356},
  url={https://api.semanticscholar.org/CorpusID:263830310}
}

@inproceedings{Yoo2023RobustMN,
  title={{Robust Multi-bit Natural Language Watermarking through Invariant Features}},
  author={Kiyoon Yoo and Wonhyuk Ahn and Jiho Jang and No Jun Kwak},
  booktitle={Annual Meeting of the Association for Computational Linguistics},
  year={2023},
  url={https://api.semanticscholar.org/CorpusID:259129912}
}

@article{Fernandez2023ThreeBT,
  title={{Three Bricks to Consolidate Watermarks for Large Language Models}},
  author={Pierre Fernandez and Antoine Chaffin and Karim Tit and Vivien Chappelier and Teddy Furon},
  journal={2023 IEEE International Workshop on Information Forensics and Security (WIFS)},
  year={2023},
  pages={1-6},
  url={https://api.semanticscholar.org/CorpusID:260351507}
}

@inproceedings{Lavergne2008DetectingFC,
  title={{Detecting Fake Content with Relative Entropy Scoring}},
  author={Thomas Lavergne and Tanguy Urvoy and François Yvon},
  booktitle={Pan},
  year={2008},
  url={https://api.semanticscholar.org/CorpusID:12098535}
}

@inproceedings{Beresneva2016ComputerGeneratedTD,
  title={{Computer-Generated Text Detection Using Machine Learning: A Systematic Review}},
  author={Daria Beresneva},
  booktitle={International Conference on Applications of Natural Language to Data Bases},
  year={2016},
  url={https://api.semanticscholar.org/CorpusID:1175726}
}

@inproceedings{Badaskar2008IdentifyingRO,
  title={{Identifying Real or Fake Articles: Towards better Language Modeling}},
  author={Sameer Badaskar and Sachin Agarwal and Shilpa Arora},
  booktitle={International Joint Conference on Natural Language Processing},
  year={2008},
  url={https://api.semanticscholar.org/CorpusID:4324753}
}

@article{Yang2023DNAGPTDN,
  title={{DNA-GPT: Divergent N-Gram Analysis for Training-Free Detection of GPT-Generated Text}},
  author={Xianjun Yang and Wei Cheng and Linda Petzold and William Yang Wang and Haifeng Chen},
  journal={ArXiv},
  year={2023},
  volume={abs/2305.17359},
  url={https://api.semanticscholar.org/CorpusID:258960101}
}

@article{Bao2023FastDetectGPTEZ,
  title={{Fast-DetectGPT: Efficient Zero-Shot Detection of Machine-Generated Text via Conditional Probability Curvature}},
  author={Guangsheng Bao and Yanbin Zhao and Zhiyang Teng and Linyi Yang and Yue Zhang},
  journal={ArXiv},
  year={2023},
  volume={abs/2310.05130},
  url={https://api.semanticscholar.org/CorpusID:263831345}
}

@article{Tulchinskii2023IntrinsicDE,
  title={{Intrinsic Dimension Estimation for Robust Detection of AI-Generated Texts}},
  author={Eduard Tulchinskii and Kristian Kuznetsov and Laida Kushnareva and Daniil Cherniavskii and S. Barannikov and Irina Piontkovskaya and Sergey I. Nikolenko and Evgeny Burnaev},
  journal={ArXiv},
  year={2023},
  volume={abs/2306.04723},
  url={https://api.semanticscholar.org/CorpusID:259108779}
}

@article{Yang2023ASO,
  title={{A Survey on Detection of LLMs-Generated Content}},
  author={Xianjun Yang and Liangming Pan and Xuandong Zhao and Haifeng Chen and Linda Ruth Petzold and William Yang Wang and Wei Cheng},
  journal={ArXiv},
  year={2023},
  volume={abs/2310.15654},
  url={https://api.semanticscholar.org/CorpusID:264439179}
}

@article{Shi2023RedTL,
  title={{Red Teaming Language Model Detectors with Language Models}},
  author={Zhouxing Shi and Yihan Wang and Fan Yin and Xiangning Chen and Kai-Wei Chang and Cho-Jui Hsieh},
  journal={ArXiv},
  year={2023},
  volume={abs/2305.19713},
  url={https://api.semanticscholar.org/CorpusID:258987266}
}

@article{Zhang2023AssayingOT,
  title={{Assaying on the Robustness of Zero-Shot Machine-Generated Text Detectors}},
  author={Yi-Fan Zhang and Zhang Zhang and Liang Wang and Tien-Ping Tan and Rong Jin},
  journal={ArXiv},
  year={2023},
  volume={abs/2312.12918},
  url={https://api.semanticscholar.org/CorpusID:266375086}
}

@article{Bhagat2013SquibsWI,
  title={{Squibs: What Is a Paraphrase?}},
  author={Rahul Bhagat and Eduard H. Hovy},
  journal={Computational Linguistics},
  year={2013},
  volume={39},
  pages={463-472},
  url={https://api.semanticscholar.org/CorpusID:32452685}
}

@article{Zellers2019DefendingAN,
  title={{Defending Against Neural Fake News}},
  author={Rowan Zellers and Ari Holtzman and Hannah Rashkin and Yonatan Bisk and Ali Farhadi and Franziska Roesner and Yejin Choi},
  journal={ArXiv},
  year={2019},
  volume={abs/1905.12616},
  url={https://api.semanticscholar.org/CorpusID:168169824}
}

@article{Solaiman2019ReleaseSA,
  title={{Release Strategies and the Social Impacts of Language Models}},
  author={Irene Solaiman and Miles Brundage and Jack Clark and Amanda Askell and Ariel Herbert-Voss and Jeff Wu and Alec Radford and Jasmine Wang},
  journal={ArXiv},
  year={2019},
  volume={abs/1908.09203},
  url={https://api.semanticscholar.org/CorpusID:201666234}
}

@article{Bakhtin2019RealOF,
  title={{Real or Fake? Learning to Discriminate Machine from Human Generated Text}},
  author={Anton Bakhtin and Sam Gross and Myle Ott and Yuntian Deng and Marc'Aurelio Ranzato and Arthur Szlam},
  journal={ArXiv},
  year={2019},
  volume={abs/1906.03351},
  url={https://api.semanticscholar.org/CorpusID:182952342}
}

@inproceedings{Uchendu2020AuthorshipAF,
  title={{Authorship Attribution for Neural Text Generation}},
  author={Adaku Uchendu and Thai Le and Kai Shu and Dongwon Lee},
  booktitle={Conference on Empirical Methods in Natural Language Processing},
  year={2020},
  url={https://api.semanticscholar.org/CorpusID:221835708}
}

@article{Li2023DeepfakeTD,
  title={{Deepfake Text Detection in the Wild}},
  author={Yafu Li and Qintong Li and Leyang Cui and Wei Bi and Longyue Wang and Linyi Yang and Shuming Shi and Yue Zhang},
  journal={ArXiv},
  year={2023},
  volume={abs/2305.13242},
  url={https://api.semanticscholar.org/CorpusID:258832454}
}

@article{Chen2023GPTSentinelDH,
  title={{GPT-Sentinel: Distinguishing Human and ChatGPT Generated Content}},
  author={Yutian Chen and Hao Kang and Vivian Zhai and Liang Li and Rita Singh and Bhiksha Ramakrishnan},
  journal={ArXiv},
  year={2023},
  volume={abs/2305.07969},
  url={https://api.semanticscholar.org/CorpusID:258686680}
}

@article{Liu2022CoCoCM,
  title={{CoCo: Coherence-Enhanced Machine-Generated Text Detection Under Data Limitation With Contrastive Learning}},
  author={Xiaoming Liu and Zhaohan Zhang and Yichen Wang and Yu Lan and Chao Shen},
  journal={ArXiv},
  year={2022},
  volume={abs/2212.10341},
  url={https://api.semanticscholar.org/CorpusID:254877728}
}

@article{Bhattacharjee2023ConDACD,
  title={{ConDA: Contrastive Domain Adaptation for AI-generated Text Detection}},
  author={Amrita Bhattacharjee and Tharindu Kumarage and Raha Moraffah and Huan Liu},
  journal={ArXiv},
  year={2023},
  volume={abs/2309.03992},
  url={https://api.semanticscholar.org/CorpusID:261660497}
}

@article{Hu2023RADARRA,
  title={{RADAR: Robust AI-Text Detection via Adversarial Learning}},
  author={Xiaobing Hu and Pin-Yu Chen and Tsung-Yi Ho},
  journal={ArXiv},
  year={2023},
  volume={abs/2307.03838},
  url={https://api.semanticscholar.org/CorpusID:259501842}
}

@article{Yu2023GPTPT,
  title={{GPT Paternity Test: GPT Generated Text Detection with GPT Genetic Inheritance}},
  author={Xiao Yu and Yuang Qi and Kejiang Chen and Guoqiang Chen and Xi Yang and Pengyuan Zhu and Weiming Zhang and Neng H. Yu},
  journal={ArXiv},
  year={2023},
  volume={abs/2305.12519},
  url={https://api.semanticscholar.org/CorpusID:258833423}
}

@article{singh2023new,
  title={{New Evaluation Metrics Capture Quality Degradation due to LLM Watermarking}},
  author={Singh, Karanpartap and Zou, James},
  journal={arXiv preprint arXiv:2312.02382},
  year={2023}
}

@article{newman1960double,
  title={The double dixie cup problem},
  author={Newman, Donald J},
  journal={The American Mathematical Monthly},
  volume={67},
  number={1},
  pages={58--61},
  year={1960},
  publisher={JSTOR}
}

\newpage
\appendix
\onecolumn

\section{Top 100 adjectives that are disproportionately used more frequently by AI}

\begin{table}[ht!]
\centering
\caption{Top 100 adjectives disproportionately used more frequently by AI. }
\begin{tabular}{lllll}
\hline
commendable & innovative & meticulous & intricate & notable \\
versatile & noteworthy & invaluable & pivotal & potent \\
fresh & ingenious & cogent & ongoing & tangible \\
profound & methodical & laudable & lucid & appreciable \\
fascinating & adaptable & admirable & refreshing & proficient \\
intriguing & thoughtful & credible & exceptional & digestible \\
prevalent & interpretative & remarkable & seamless & economical \\
proactive & interdisciplinary & sustainable & optimizable & comprehensive \\
vital & pragmatic & comprehensible & unique & fuller \\
authentic & foundational & distinctive & pertinent & valuable \\
invasive & speedy & inherent & considerable & holistic \\
insightful & operational & substantial & compelling & technological \\
beneficial & excellent & keen & cultural & unauthorized \\
strategic & expansive & prospective & vivid & consequential \\
manageable & unprecedented & inclusive & asymmetrical & cohesive \\
replicable & quicker & defensive & wider & imaginative \\
traditional & competent & contentious & widespread & environmental \\
instrumental & substantive & creative & academic & sizeable \\
extant & demonstrable & prudent & practicable & signatory \\
continental & unnoticed & automotive & minimalistic & intelligent \\
\hline
\end{tabular}
\label{table:word_adj_list}
\end{table}

\begin{figure}[ht!]
    \centering
    \includegraphics[width=1\textwidth]{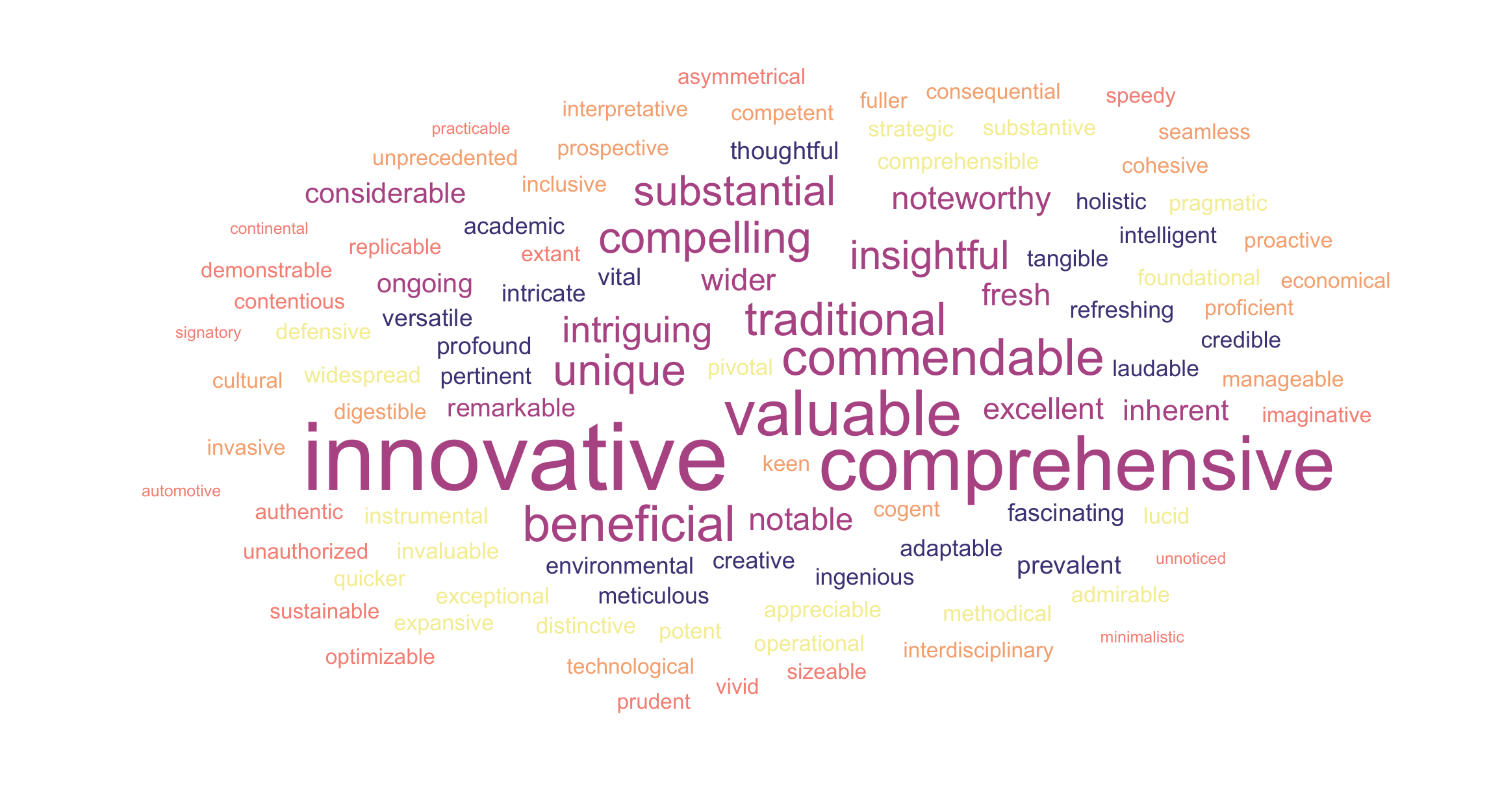}
    \caption{
    Word cloud of top 100 adjectives in LLM feedback, with font size indicating frequency.
    }
    \label{fig:word-cloud-adj}
\end{figure}


\clearpage
\newpage 
\section{Top 100 adverbs that are disproportionately used more frequently by AI}

\begin{table}[ht!]
\centering
\caption{Top 100 adverbs disproportionately used more frequently by AI. }
\begin{tabular}{lllll}
\hline
meticulously & reportedly & lucidly & innovatively & aptly \\
methodically & excellently & compellingly & impressively & undoubtedly \\
scholarly & strategically & intriguingly & competently & intelligently \\
hitherto & thoughtfully & profoundly & undeniably & admirably \\
creatively & logically & markedly & thereby & contextually \\
distinctly & judiciously & cleverly & invariably & successfully \\
chiefly & refreshingly & constructively & inadvertently & effectively \\
intellectually & rightly & convincingly & comprehensively & seamlessly \\
predominantly & coherently & evidently & notably & professionally \\
subtly & synergistically & productively & purportedly & remarkably \\
traditionally & starkly & promptly & richly & nonetheless \\
elegantly & smartly & solidly & inadequately & effortlessly \\
forth & firmly & autonomously & duly & critically \\
immensely & beautifully & maliciously & finely & succinctly \\
further & robustly & decidedly & conclusively & diversely \\
exceptionally & concurrently & appreciably & methodologically & universally \\
thoroughly & soundly & particularly & elaborately & uniquely \\
neatly & definitively & substantively & usefully & adversely \\
primarily & principally & discriminatively & efficiently & scientifically \\
alike & herein & additionally & subsequently & potentially \\
\hline
\end{tabular}
\label{table:word_adv_list}
\end{table}
\begin{figure}[ht!]
    \centering
    \includegraphics[width=1\textwidth]{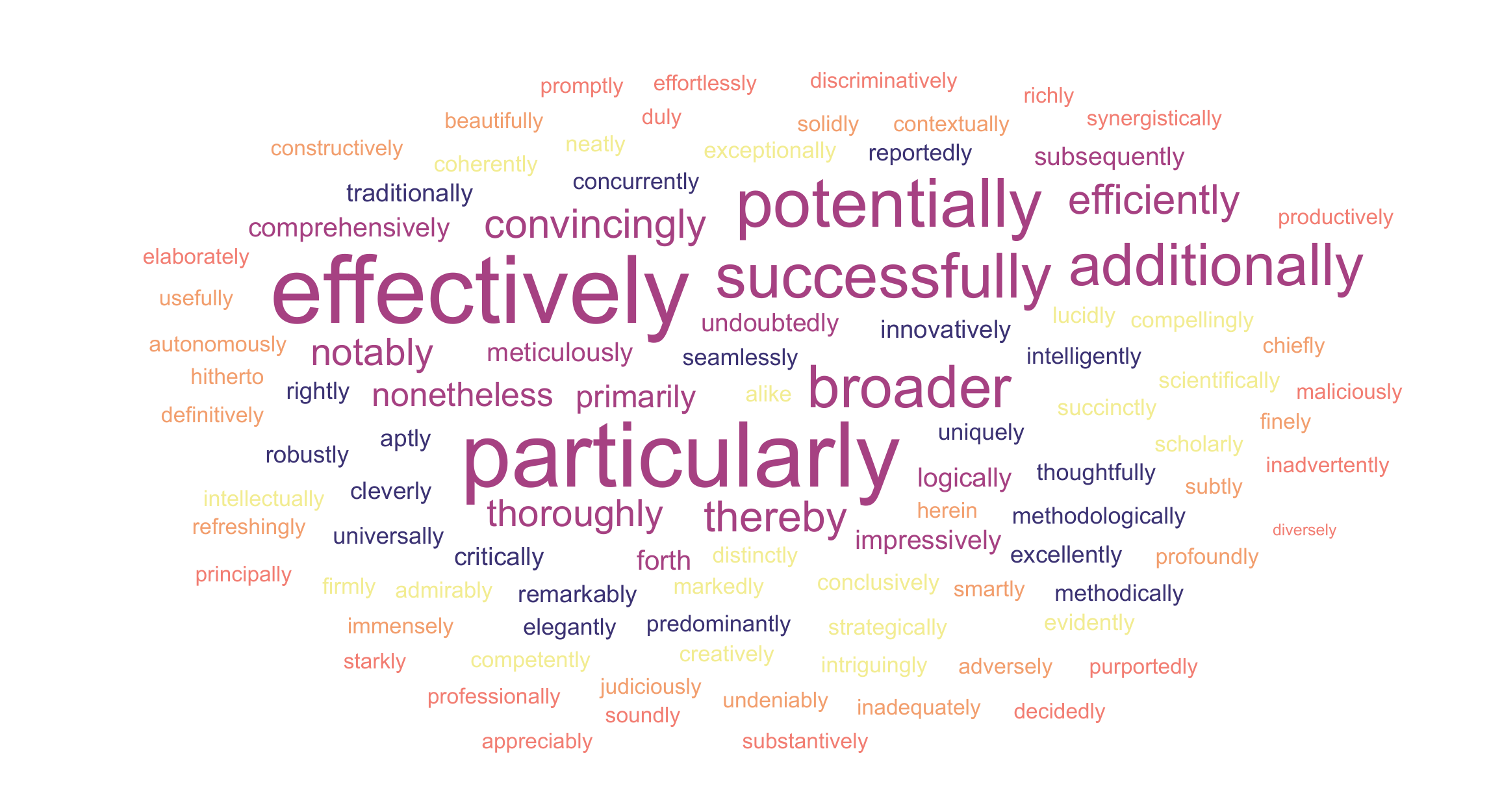}
    \caption{
        Word cloud of top 100 adverbs in LLM feedback, with font size indicating frequency.
    }
    \label{fig:word-cloud-adv}
\end{figure}

\newpage 
\clearpage

\section{Additional Details on Major ML Conferences Reviewer Confidence Scale} \label{appendix: data}
Here we include additional details on the datasets used for our experiments. Table~\ref{tab:appendix-confidence-scale} includes the descriptions of the reviewer confidence scales for each conference.

\begin{table}[ht!]
\centering
\caption{Confidence Scale Description for Major ML Conferences}
\setlength{\tabcolsep}{3.5pt}
\begin{tabular}{r p{12cm}}
\toprule
\textbf{Conference} & \textbf{Confidence Scale Description} \\
\midrule
ICLR 2024 &
1: You are unable to assess this paper and have alerted the ACs to seek an opinion from different reviewers.  
\\
& 
2: You are willing to defend your assessment, but it is quite likely that you did not understand the central parts of the submission or that you are unfamiliar with some pieces of related work. Math/other details were not carefully checked.
\\
& 
3: You are fairly confident in your assessment. It is possible that you did not understand some parts of the submission or that you are unfamiliar with some pieces of related work. Math/other details were not carefully checked. 
\\
& 
4: You are confident in your assessment, but not absolutely certain. It is unlikely, but not impossible, that you did not understand some parts of the submission or that you are unfamiliar with some pieces of related work.                            
\\
& 
5: You are absolutely certain about your assessment. You are very familiar with the related work and checked the math/other details carefully. 
\\
\midrule
NeurIPS 2023 &
1: Your assessment is an educated guess. The submission is not in your area or the submission was difficult to understand. Math/other details were not carefully checked.                       
\\
& 
2: You are willing to defend your assessment, but it is quite likely that you did not understand the central parts of the submission or that you are unfamiliar with some pieces of related work. Math/other details were not carefully checked.
\\
& 
3: You are fairly confident in your assessment. It is possible that you did not understand some parts of the submission or that you are unfamiliar with some pieces of related work. Math/other details were not carefully checked.   
\\
& 
4: You are confident in your assessment, but not absolutely certain. It is unlikely, but not impossible, that you did not understand some parts of the submission or that you are unfamiliar with some pieces of related work.  
\\
& 
5: You are absolutely certain about your assessment. You are very familiar with the related work and checked the math/other details carefully.  
\\
\midrule
CoRL 2023 &
1: The reviewer's evaluation is an educated guess
\\
& 
2: The reviewer is willing to defend the evaluation, but it is quite likely that the reviewer did not understand central parts of the paper
\\
& 
3: The reviewer is fairly confident that the evaluation is correct  
\\
& 
4: The reviewer is confident but not absolutely certain that the evaluation is correct 
\\
& 
5: The reviewer is absolutely certain that the evaluation is correct and very familiar with the relevant literature
\\
\midrule
EMNLP 2023 &
1: Not my area, or paper was hard for me to understand. My evaluation is just an educated guess.                    
\\
& 
2: Willing to defend my evaluation, but it is fairly likely that I missed some details, didn't understand some central points, or can't be sure about the novelty of the work. 
\\
& 
3: Pretty sure, but there's a chance I missed something. Although I have a good feel for this area in general, I did not carefully check the paper's details, e.g., the math, experimental design, or novelty.
\\
& 
4: Quite sure. I tried to check the important points carefully. It's unlikely, though conceivable, that I missed something that should affect my ratings.    
\\
& 
5: Positive that my evaluation is correct. I read the paper very carefully and I am very familiar with related work.  
\\
\bottomrule
\end{tabular}
\label{tab:appendix-confidence-scale}
\end{table}

\newpage 
\clearpage

\section{Additional Results} \label{appendix: additional results}
In this appendix, we collect additional experimental results. This includes tables of the exact numbers used to produce the figures in the main text, as well as results for additional experiments not reported in the main text.


\subsection{Validation Accuracy Tables}
Here we present the numerical results for validating our method in Section~\ref{sec: val}. Table~\ref{tab: adj val}, ~\ref{tab:verification-adj-main-nature} shows the numerical values used in Figure~\ref{fig: val}.

We also trained a separate model for \textit{Nature} family journals using official review data for papers accepted between 2021-09-13 and 2022-08-03. 
We validated the model's accuracy on reviews for papers accepted between 2022-08-04 and 2022-11-29 (Figure~\ref{fig: val}, \textit{Nature Portfolio} '22, Table~\ref{tab:verification-adj-main-nature}).

\begin{figure}[ht!] 
\centering
\includegraphics[width=1\textwidth]{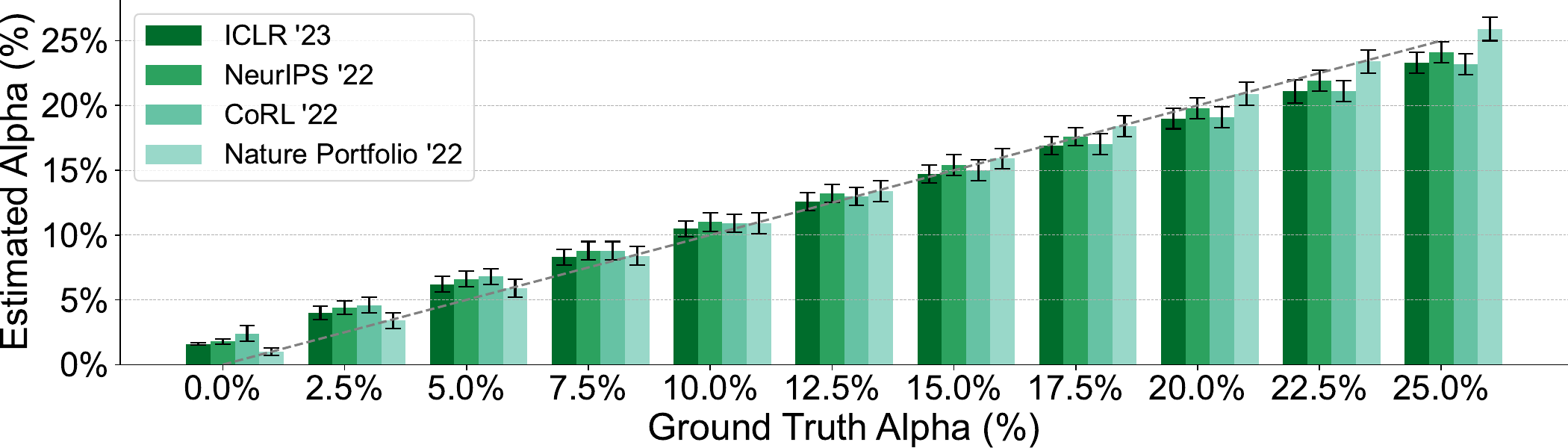}
\caption{
Full Results of the validation procedure from Section~\ref{sec: val} using adjectives.
}
\label{fig: full adj val}
\end{figure}

\begin{table}[htb!]
\small
\begin{center}
\caption{
\textbf{Performance validation of our model} across \textit{ICLR} '23, \textit{NeurIPS} '22, and \textit{CoRL} '22 reviews (all predating ChatGPT's launch), using a blend of official human and LLM-generated reviews. 
Our algorithm demonstrates high accuracy with less than 2.4\% prediction error in identifying the proportion of LLM reviews within the validation set.
This table presents the results data for Figure ~\ref{fig: val}. 
}
\label{tab: adj val}
\begin{tabular}{lrcllc}
\cmidrule[\heavyrulewidth]{1-6}
\multirow{2}{*}{\bf No.} 
& \multirow{2}{*}{\bf \begin{tabular}[c]{@{}c@{}} Validation \\ Data Source 
\end{tabular} } 
& \multirow{2}{*}{\bf \begin{tabular}[c]{@{}c@{}} Ground \\ Truth $\alpha$
\end{tabular}}  
&\multicolumn{2}{l}{\bf Estimated} 
& \multirow{2}{*}{\bf \begin{tabular}[c]{@{}c@{}} Prediction \\ Error 
\end{tabular} } 
\\
\cmidrule{4-5}
 & & & $\alpha$ & $CI$ ($\pm$) & \\
\cmidrule{1-6}
(1) & \emph{ICLR} 2023 & 0.0\% & 1.6\% & 0.1\% & 1.6\% \\
(2) & \emph{ICLR} 2023 & 2.5\% & 4.0\% & 0.5\% & 1.5\% \\
(3) & \emph{ICLR} 2023 & 5.0\% & 6.2\% & 0.6\% & 1.2\% \\
(4) & \emph{ICLR} 2023 & 7.5\% & 8.3\% & 0.6\% & 0.8\% \\
(5) & \emph{ICLR} 2023 & 10.0\% & 10.5\% & 0.6\% & 0.5\% \\
(6) & \emph{ICLR} 2023 & 12.5\% & 12.6\% & 0.7\% & 0.1\% \\
(7) & \emph{ICLR} 2023 & 15.0\% & 14.7\% & 0.7\% & 0.3\% \\
(8) & \emph{ICLR} 2023 & 17.5\% & 16.9\% & 0.7\% & 0.6\% \\
(9) & \emph{ICLR} 2023 & 20.0\% & 19.0\% & 0.8\% & 1.0\% \\
(10) & \emph{ICLR} 2023 & 22.5\% & 21.1\% & 0.9\% & 1.4\% \\
(11) & \emph{ICLR} 2023 & 25.0\% & 23.3\% & 0.8\% & 1.7\% \\
\cmidrule{1-6}
(12) & \emph{NeurIPS} 2022 & 0.0\% & 1.8\% & 0.2\% & 1.8\% \\
(13) & \emph{NeurIPS} 2022 & 2.5\% & 4.4\% & 0.5\% & 1.9\% \\
(14) & \emph{NeurIPS} 2022 & 5.0\% & 6.6\% & 0.6\% & 1.6\% \\
(15) & \emph{NeurIPS} 2022 & 7.5\% & 8.8\% & 0.7\% & 1.3\% \\
(16) & \emph{NeurIPS} 2022 & 10.0\% & 11.0\% & 0.7\% & 1.0\% \\
(17) & \emph{NeurIPS} 2022 & 12.5\% & 13.2\% & 0.7\% & 0.7\% \\
(18) & \emph{NeurIPS} 2022 & 15.0\% & 15.4\% & 0.8\% & 0.4\% \\
(19) & \emph{NeurIPS} 2022 & 17.5\% & 17.6\% & 0.7\% & 0.1\% \\
(20) & \emph{NeurIPS} 2022 & 20.0\% & 19.8\% & 0.8\% & 0.2\% \\
(21) & \emph{NeurIPS} 2022 & 22.5\% & 21.9\% & 0.8\% & 0.6\% \\
(22) & \emph{NeurIPS} 2022 & 25.0\% & 24.1\% & 0.8\% & 0.9\% \\
\cmidrule{1-6}
(23) & \emph{CoRL} 2022 & 0.0\% & 2.4\% & 0.6\% & 2.4\% \\
(24) & \emph{CoRL} 2022 & 2.5\% & 4.6\% & 0.6\% & 2.1\% \\
(25) & \emph{CoRL} 2022 & 5.0\% & 6.8\% & 0.6\% & 1.8\% \\
(26) & \emph{CoRL} 2022 & 7.5\% & 8.8\% & 0.7\% & 1.3\% \\
(27) & \emph{CoRL} 2022 & 10.0\% &10.9\% & 0.7\% & 0.9\% \\
(28) & \emph{CoRL} 2022 & 12.5\% & 13.0\% & 0.7\% & 0.5\% \\
(29) & \emph{CoRL} 2022 & 15.0\% & 15.0\% & 0.8\% & 0.0\% \\
(30) & \emph{CoRL} 2022 & 17.5\% & 17.0\% & 0.8\% & 0.5\% \\
(31) & \emph{CoRL} 2022 & 20.0\% & 19.1\% & 0.8\% & 0.9\% \\
(32) & \emph{CoRL} 2022 & 22.5\% & 21.1\% & 0.8\% & 1.4\% \\
(33) & \emph{CoRL} 2022 & 25.0\% & 23.2\% & 0.8\% & 1.8\% \\
\cmidrule[\heavyrulewidth]{1-6}
\end{tabular}
\label{tab:verification-adj-main}
\end{center}
\vspace{-5mm}
\end{table}

\begin{table}[htb!]
\small
\begin{center}
\caption{
\textbf{Performance validation of our model} across \textit{Nature} family journals (all predating ChatGPT's launch), using a blend of official human and LLM-generated reviews. 
This table presents the results data for Figure ~\ref{fig: val}. 
}
\begin{tabular}{lrcllc}
\cmidrule[\heavyrulewidth]{1-6}
\multirow{2}{*}{\bf No.} 
& \multirow{2}{*}{\bf \begin{tabular}[c]{@{}c@{}} Validation \\ Data Source 
\end{tabular} } 
& \multirow{2}{*}{\bf \begin{tabular}[c]{@{}c@{}} Ground \\ Truth $\alpha$
\end{tabular}}  
&\multicolumn{2}{l}{\bf Estimated} 
& \multirow{2}{*}{\bf \begin{tabular}[c]{@{}c@{}} Prediction \\ Error 
\end{tabular} } 
\\
\cmidrule{4-5}
 & & & $\alpha$ & $CI$ ($\pm$) & \\
\cmidrule{1-6}
(1) & \emph{Nature Portfolio} 2022 & 0.0\% & 1.0\% & 0.3\% & 1.0\% \\
(2) & \emph{Nature Portfolio} 2022 & 2.5\% & 3.4\% & 0.6\% & 0.9\% \\
(3) & \emph{Nature Portfolio} 2022 & 5.0\% & 5.9\% & 0.7\% & 0.9\% \\
(4) & \emph{Nature Portfolio} 2022 & 7.5\% & 8.4\% & 0.7\% & 0.9\% \\
(5) & \emph{Nature Portfolio} 2022 & 10.0\% & 10.9\% & 0.8\% & 0.9\% \\
(6) & \emph{Nature Portfolio} 2022 & 12.5\% & 13.4\% & 0.8\% & 0.9\% \\
(7) & \emph{Nature Portfolio} 2022 & 15.0\% & 15.9\% & 0.8\% & 0.9\% \\
(8) & \emph{Nature Portfolio} 2022 & 17.5\% & 18.4\% & 0.8\% & 0.9\% \\
(9) & \emph{Nature Portfolio} 2022 & 20.0\% & 20.9\% & 0.9\% & 0.9\% \\
(10) & \emph{Nature Portfolio} 2022 & 22.5\% & 23.4\% & 0.9\% & 0.9\% \\
(11) & \emph{Nature Portfolio} 2022 & 25.0\% & 25.9\% & 0.9\% & 0.9\% \\
\cmidrule[\heavyrulewidth]{1-6}
\end{tabular}
\label{tab:verification-adj-main-nature}
\end{center}
\vspace{-5mm}
\end{table}

\newpage 

\subsection{Main Results Tables}
\label{sec:main-results}
Here we present the numerical results for estimating on real reviews in Section~\ref{sec: main-results}. Table~\ref{tab:main-result}, ~\ref{tab: Nature trend} shows the numerical values used in Figure~\ref{fig: temporal}.
We still use our separately trained model for \textit{Nature} family journals in main results estimation.

\begin{table}[htb!]
\small
\begin{center}
\caption{Temporal trends of ML conferences in the $\a$ estimate on official reviews using adjectives. $\a$ estimates pre-ChatGPT are close to 0, and there is a sharp increase after the release of ChatGPT.
This table presents the results data for Figure ~\ref{fig: temporal}.}
\begin{tabular}{lrcll}
\cmidrule[\heavyrulewidth]{1-4}
\multirow{2}{*}{\bf No.} 
& \multirow{2}{*}{\bf \begin{tabular}[c]{@{}c@{}} Validation \\ Data Source 
\end{tabular} } 
&\multicolumn{2}{l}{\bf Estimated} 
\\
\cmidrule{3-4}
 & & $\alpha$ & $CI$ ($\pm$) \\
\cmidrule{1-4}
(1) & \emph{NeurIPS} 2019 & 1.7\% & 0.3\% \\
(2) & \emph{NeurIPS} 2020 & 1.4\% & 0.1\% \\
(3) & \emph{NeurIPS} 2021 & 1.6\% & 0.2\% \\
(4) & \emph{NeurIPS} 2022 & 1.9\% & 0.2\% \\
(5) & \emph{NeurIPS} 2023 & 9.1\%  & 0.2\% \\

\cmidrule{1-4} 
(6) & \emph{ICLR} 2023 & 1.6\% & 0.1\% \\
(7) & \emph{ICLR} 2024 & 10.6\% & 0.2\% \\

\cmidrule{1-4} 
(8) & \emph{CoRL} 2021 & 2.4\% & 0.7\% \\
(9) & \emph{CoRL} 2022 & 2.4\% & 0.6\% \\
(10) & \emph{CoRL} 2023 & 6.5\% & 0.7\% \\

\cmidrule{1-4} 
(11) & \emph{EMNLP} 2023 & 16.9\% & 0.5\% \\

\cmidrule[\heavyrulewidth]{1-4}
\end{tabular}
\label{tab:main-result}
\end{center}
\vspace{-5mm}
\end{table}

\begin{table}[htb!]
\small
\begin{center}
\caption{
Temporal trends of the \textit{Nature} family journals in the $\a$ estimate on official reviews using adjectives. 
Contrary to the ML conferences, the \textit{Nature} family journals did not exhibit a significant increase in the estimated $\alpha$ values following ChatGPT's release, with pre- and post-release $\alpha$ estimates remaining within the margin of error for the $\alpha=0$ validation experiment.
This table presents the results data for Figure ~\ref{fig: temporal}.}
\label{tab: Nature trend}
\begin{tabular}{lrcll}
\cmidrule[\heavyrulewidth]{1-4}
\multirow{2}{*}{\bf No.} 
& \multirow{2}{*}{\bf \begin{tabular}[c]{@{}c@{}} Validation \\ Data Source 
\end{tabular} } 
&\multicolumn{2}{l}{\bf Estimated} 
\\
\cmidrule{3-4}
 & & $\alpha$ & $CI$ ($\pm$) \\
\cmidrule{1-4}
(1) &   \emph{Nature portfolio} 2019    & 0.8\% & 0.2\% \\
(2) &   \emph{Nature portfolio} 2020    & 0.7\% & 0.2\% \\
(3) &   \emph{Nature portfolio} 2021    & 1.1\% & 0.2\% \\
(4) &   \emph{Nature portfolio} 2022    & 1.0\% & 0.3\% \\
(5) &   \emph{Nature portfolio} 2023    & 1.6\% & 0.2\% \\

\cmidrule[\heavyrulewidth]{1-4}
\end{tabular}
\end{center}
\vspace{-5mm}
\end{table}


\clearpage
\newpage 

\subsection{Sensitivity to LLM Prompt}
\label{Appendix:subsec:LLM-prompt-shift}

\begin{figure}[ht!] 
\centering
\includegraphics[width=1\textwidth]{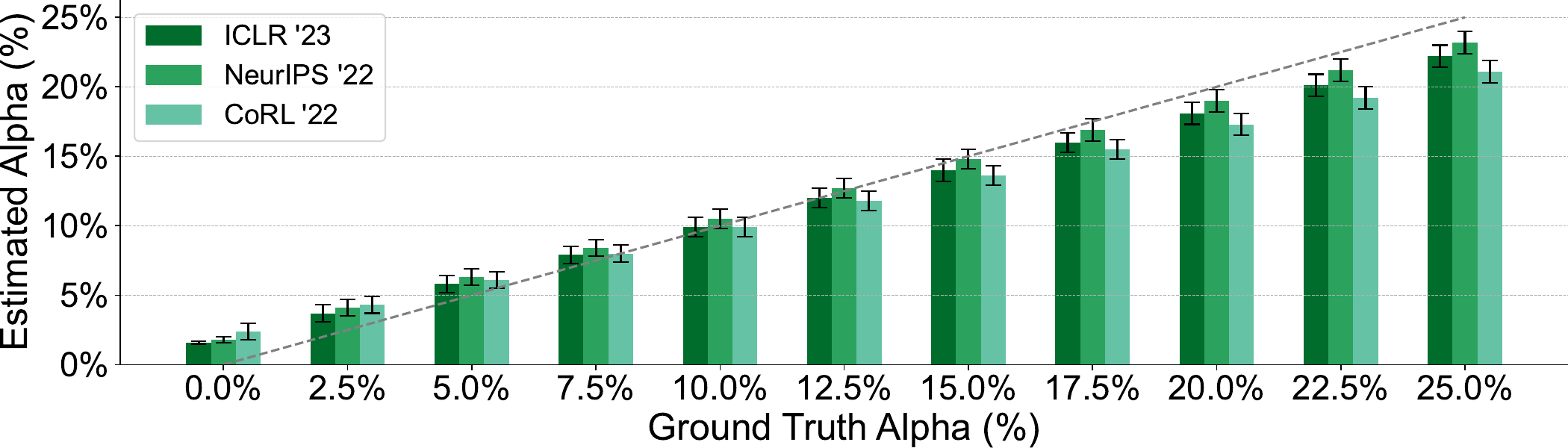}
\caption{
Results of the validation procedure from Section~\ref{sec: val} using a different prompt.
}
\label{fig: diff prompt val}
\end{figure}

Empirically, we found that our framework exhibits moderate robustness to the distribution shift of LLM prompts. Training with one prompt and testing on a different prompt still yields accurate validation results (Supp. Figure~\ref{fig: diff prompt val}). 
Figure~\ref{fig:training-prompt} shows the prompt for generating training data with GPT-4 June. Figure~\ref{fig:validation-prompt-shift-prompt} shows the prompt for generating validation data on prompt shift.

Table~\ref{tab: diff prompt val} shows the results using a different prompt than that in the main text.

\begin{table}[htb!]
\small
\begin{center}
\caption{
\textbf{Validation accuracy for our method using a different prompt.} The model was trained using data from \textit{ICLR} 2018-2022, and OOD verification was performed on \textit{NeurIPS} and \textit{CoRL} (moderate distribution shift). The method is robust  to changes in the prompt and still exhibits accurate and stable performance.
}
\label{tab: diff prompt val}
\begin{tabular}{lrcllc}
\cmidrule[\heavyrulewidth]{1-6}
\multirow{2}{*}{\bf No.} 
& \multirow{2}{*}{\bf \begin{tabular}[c]{@{}c@{}} Validation \\ Data Source 
\end{tabular} } 
& \multirow{2}{*}{\bf \begin{tabular}[c]{@{}c@{}} Ground \\ Truth $\alpha$
\end{tabular}}  
&\multicolumn{2}{l}{\bf Estimated} 
& \multirow{2}{*}{\bf \begin{tabular}[c]{@{}c@{}} Prediction \\ Error 
\end{tabular} } 
\\
\cmidrule{4-5}
 & & & $\alpha$ & $CI$ ($\pm$) & \\
\cmidrule{1-6}
(1) & \emph{ICLR} 2023 & 0.0\% & 1.6\% & 0.1\% & 1.6\% \\
(2) & \emph{ICLR} 2023 & 2.5\% & 3.7\% & 0.6\% & 1.2\% \\
(3) & \emph{ICLR} 2023 & 5.0\% & 5.8\% & 0.6\% & 0.8\% \\
(4) & \emph{ICLR} 2023 & 7.5\% & 7.9\% & 0.6\% & 0.4\% \\
(5) & \emph{ICLR} 2023 & 10.0\% & 9.9\% & 0.7\% & 0.1\% \\
(6) & \emph{ICLR} 2023 & 12.5\% & 12.0\% & 0.7\% & 0.5\% \\
(7) & \emph{ICLR} 2023 & 15.0\% & 14.0\% & 0.8\% & 1.0\% \\
(8) & \emph{ICLR} 2023 & 17.5\% & 16.0\% & 0.7\% & 1.5\% \\
(9) & \emph{ICLR} 2023 & 20.0\% & 18.1\% & 0.8\% & 1.9\% \\
(10) & \emph{ICLR} 2023 & 22.5\% & 20.1\% & 0.8\% & 2.4\% \\
(11) & \emph{ICLR} 2023 & 25.0\% & 22.2\% & 0.8\% & 2.8\% \\
\cmidrule{1-6}
(12) & \emph{NeurIPS} 2022 & 0.0\% & 1.8\% & 0.2\%  & 1.8\%\\
(13) & \emph{NeurIPS} 2022 & 2.5\% & 4.1\% & 0.6\%  & 1.6\%\\
(14) & \emph{NeurIPS} 2022 & 5.0\% & 6.3\% & 0.6\%  & 1.3\%\\
(15) & \emph{NeurIPS} 2022 & 7.5\% & 8.4\% & 0.6\% & 0.9\%\\
(16) & \emph{NeurIPS} 2022 & 10.0\% & 10.5\% & 0.7\% & 0.5\%\\
(17) & \emph{NeurIPS} 2022 & 12.5\% & 12.7\% & 0.7\% & 0.2\%\\
(18) & \emph{NeurIPS} 2022 & 15.0\% & 14.8\% & 0.7\% & 0.2\%\\
(19) & \emph{NeurIPS} 2022 & 17.5\% & 16.9\% & 0.8\% & 0.6\%\\
(20) & \emph{NeurIPS} 2022 & 20.0\% & 19.0\% & 0.8\% & 1.0\%\\
(21) & \emph{NeurIPS} 2022 & 22.5\% & 21.2\% & 0.8\% & 1.3\%\\
(22) & \emph{NeurIPS} 2022 & 25.0\% & 23.2\% & 0.8\% & 1.8\%\\
\cmidrule{1-6}
(23) & \emph{CoRL} 2022 & 0.0\% & 2.4\% & 0.6\% & 2.4\% \\
(24) & \emph{CoRL} 2022 & 2.5\% & 4.3\% & 0.6\% & 1.8\% \\
(25) & \emph{CoRL} 2022 & 5.0\% & 6.1\% & 0.6\% & 1.1\% \\
(26) & \emph{CoRL} 2022 & 7.5\% & 8.0\% & 0.6\% & 0.5\% \\
(27) & \emph{CoRL} 2022 & 10.0\% &9.9\% & 0.7\% & 0.1\% \\
(28) & \emph{CoRL} 2022 & 12.5\% & 11.8\% & 0.7\% & 0.7\% \\
(29) & \emph{CoRL} 2022 & 15.0\% & 13.6\% & 0.7\% & 1.4\% \\
(30) & \emph{CoRL} 2022 & 17.5\% & 15.5\% & 0.7\% & 2.0\% \\
(31) & \emph{CoRL} 2022 & 20.0\% & 17.3\% & 0.8\% & 2.7\% \\
(32) & \emph{CoRL} 2022 & 22.5\% & 19.2\% & 0.8\% & 3.3\% \\
(33) & \emph{CoRL} 2022 & 25.0\% & 21.1\% & 0.8\% & 3.9\% \\
\cmidrule[\heavyrulewidth]{1-6}
\end{tabular}
\end{center}
\vspace{-5mm}
\end{table}

\newpage 
\clearpage

\subsection{Tables for Stratification by Paper Topic (\textit{ICLR})}
Here, we provide the numerical results for various fields in the \textit{ICLR} 2024 conference. The results are shown in Table ~\ref{tab: fields}.
\begin{table}[htb]
\small
\begin{center}
\caption{
Changes in the estimated $\a$ for different fields of ML (sorted according to a paper's designated primary area in \textit{ICLR} 2024).
}
\label{tab: fields}
\resizebox{0.97\textwidth}{!}{
\begin{tabular}{lrcll}
%
\cmidrule[\heavyrulewidth]{1-5}
\multirow{2}{*}{\bf No.} 
& \multirow{2}{*}{\bf \begin{tabular}[c]{@{}c@{}} ICLR 2024 Primary Area 
\end{tabular} } 
& \multirow{2}{*}{\bf \begin{tabular}[c]{@{}c@{}} \# of \\ Papers
\end{tabular}}  
&\multicolumn{2}{l}{\bf Estimated} 
\\
\cmidrule{4-5}
 & & & $\alpha$ & $CI$ ($\pm$) \\
\cmidrule{1-5}
(1) & Datasets and Benchmarks  & 271 & 20.9\% &1.0\% \\
(2) & Transfer Learning, Meta Learning, and Lifelong Learning  & 375 & 14.0\% &0.8\%\\
(3) & Learning on Graphs and Other Geometries \& Topologies  & 189 & 12.6\% &1.0\%\\
(4) & Applications to Physical Sciences (Physics, Chemistry, Biology, etc.) & 312 & 12.4\% &0.8\%\\
(5) & Representation Learning for Computer Vision, Audio, Language, and Other Modalities  & 1037 &12.3\% &0.5\%\\
(6) & Unsupervised, Self-supervised, Semi-supervised, and Supervised Representation Learning  & 856 & 11.9\% &0.5\%\\
(7) & Infrastructure, Software Libraries, Hardware, etc.   & 47 & 11.5\% &2.0\%\\
(8) & Societal Considerations including Fairness, Safety, Privacy   & 535 & 11.4\% &0.6\%\\
(9) & General Machine Learning (i.e., None of the Above)  & 786 & 11.3\% &0.5\%\\
(10) & Applications to Neuroscience \& Cognitive Science   & 133 & 10.9\% &1.1\%\\
(11) & Generative Models  & 777 & 10.4\% &0.5\%\\
(12) & Applications to Robotics, Autonomy, Planning   & 177 & 10.0\% &0.9\%\\
(13) & Visualization or Interpretation of Learned Representations  & 212 & 8.4\% &0.8\%\\
(14) & Reinforcement Learning  & 654 & 8.2\% &0.4\%\\
(15) & Neurosymbolic \& Hybrid AI Systems (Physics-informed, Logic \& Formal Reasoning, etc.) & 101 & 7.7\% &1.3\% \\
(16) & Learning Theory  & 211 & 7.3\% &0.8\%\\
(17) & Metric learning, Kernel learning, and Sparse coding   & 36 & 7.2\% &2.1\%\\
(18) & Probabilistic Methods  (Bayesian Methods, Variational Inference, Sampling, UQ, etc.) & 184 & 6.0\% &0.8\%\\
(19) & Optimization  & 312 & 5.8\%\ &0.6\% \\
(20) & Causal Reasoning & 99  & 5.0\% &1.0\%\\
\cmidrule[\heavyrulewidth]{1-5}
\end{tabular}
}
\end{center}
\vspace{-5mm}
\end{table}


\newpage 
\clearpage

\subsection{Results with Adverbs}
\label{Appendix:subsec:adverbs}
For our results in the main paper, we only considered adjectives for the space of all possible tokens. 
We found this vocabulary choice to exhibit greater stability than using other parts of speech such as adverbs, verbs, nouns, or all possible tokens.  
This remotely aligns with the findings in the literature~\cite{lin2023unlocking}, which indicate that \textit{stylistic} words are the most impacted during alignment fine-tuning.

Here, we conducted experiments using adverbs. The results for adverbs are shown in Table~\ref{tab: adv val}.

\begin{figure}[htb!] 
\centering
\includegraphics[width=1\textwidth]{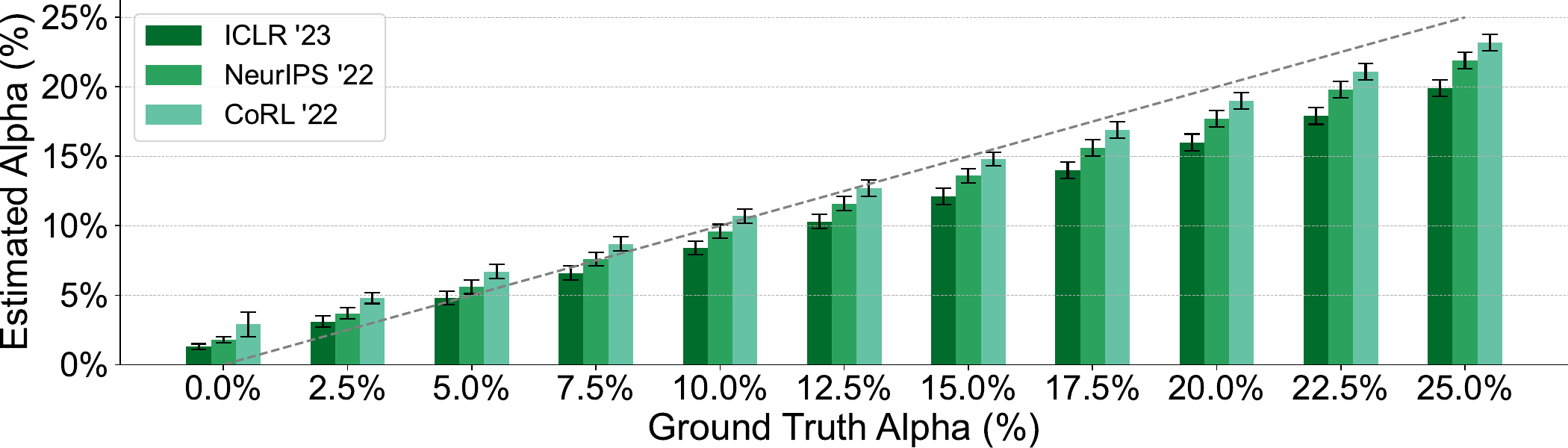}
\caption{
Results of the validation procedure from Section~\ref{sec: val} using adverbs (instead of adjectives).
}
\label{fig: adv val}
\end{figure}
\begin{table}[htb!]
\small
\begin{center}
\caption{
Validation results when adverbs are used. The performance degrades compared to using adjectives.
}
\label{tab: adv val}
\begin{tabular}{lrcllc}
\cmidrule[\heavyrulewidth]{1-6}
\multirow{2}{*}{\bf No.} 
& \multirow{2}{*}{\bf \begin{tabular}[c]{@{}c@{}} Validation \\ Data Source 
\end{tabular} } 
& \multirow{2}{*}{\bf \begin{tabular}[c]{@{}c@{}} Ground \\ Truth $\alpha$
\end{tabular}}  
&\multicolumn{2}{l}{\bf Estimated} 
& \multirow{2}{*}{\bf \begin{tabular}[c]{@{}c@{}} Prediction \\ Error 
\end{tabular} } 
\\
\cmidrule{4-5}
 & & & $\alpha$ & $CI$ ($\pm$) & \\
\cmidrule{1-6}
(1) & \emph{ICLR} 2023 & 0.0\% & 1.3\% & 0.2\% & 1.3\% \\
(2) & \emph{ICLR} 2023 & 2.5\% & 3.1\% & 0.4\% & 0.6\% \\
(3) & \emph{ICLR} 2023 & 5.0\% & 4.8\% & 0.5\% & 0.2\% \\
(4) & \emph{ICLR} 2023 & 7.5\% & 6.6\% & 0.5\% & 0.9\% \\
(5) & \emph{ICLR} 2023 & 10.0\% & 8.4\% & 0.5\% & 1.6\% \\
(6) & \emph{ICLR} 2023 & 12.5\% & 10.3\% & 0.5\% & 2.2\% \\
(7) & \emph{ICLR} 2023 & 15.0\% & 12.1\% & 0.6\% & 2.9\% \\
(8) & \emph{ICLR} 2023 & 17.5\% & 14.0\% & 0.6\% & 3.5\% \\
(9) & \emph{ICLR} 2023 & 20.0\% & 16.0\% & 0.6\% & 4.0\% \\
(10) & \emph{ICLR} 2023 & 22.5\% & 17.9\% & 0.6\% & 4.6\% \\
(11) & \emph{ICLR} 2023 & 25.0\% & 19.9\% & 0.6\% & 5.1\% \\
\cmidrule{1-6}
(12) & \emph{NeurIPS} 2022 & 0.0\% & 1.8\% & 0.2\%  & 1.8\%\\
(13) & \emph{NeurIPS} 2022 & 2.5\% & 3.7\% & 0.4\%  & 1.2\%\\
(14) & \emph{NeurIPS} 2022 & 5.0\% & 5.6\% & 0.5\%  & 0.6\%\\
(15) & \emph{NeurIPS} 2022 & 7.5\% & 7.6\% & 0.5\% & 0.1\%\\
(16) & \emph{NeurIPS} 2022 & 10.0\% & 9.6\% & 0.5\% & 0.4\%\\
(17) & \emph{NeurIPS} 2022 & 12.5\% & 11.6\% & 0.5\% & 0.9\%\\
(18) & \emph{NeurIPS} 2022 & 15.0\% & 13.6\% & 0.5\% & 1.4\%\\
(19) & \emph{NeurIPS} 2022 & 17.5\% & 15.6\% & 0.6\% & 1.9\%\\
(20) & \emph{NeurIPS} 2022 & 20.0\% & 17.7\% & 0.6\% & 2.3\%\\
(21) & \emph{NeurIPS} 2022 & 22.5\% & 19.8\% & 0.6\% & 2.7\%\\
(22) & \emph{NeurIPS} 2022 & 25.0\% & 21.9\% & 0.6\% & 3.1\%\\
\cmidrule{1-6}
(23) & \emph{CoRL} 2022 & 0.0\% & 2.9\% & 0.9\% & 2.9\% \\
(24) & \emph{CoRL} 2022 & 2.5\% & 4.8\% & 0.4\% & 2.3\% \\
(25) & \emph{CoRL} 2022 & 5.0\% & 6.7\% & 0.5\% & 1.7\% \\
(26) & \emph{CoRL} 2022 & 7.5\% & 8.7\% & 0.5\% & 1.2\% \\
(27) & \emph{CoRL} 2022 & 10.0\% &10.7\% & 0.5\% & 0.7\% \\
(28) & \emph{CoRL} 2022 & 12.5\% & 12.7\% & 0.6\% & 0.2\% \\
(29) & \emph{CoRL} 2022 & 15.0\% & 14.8\% & 0.5\% & 0.2\% \\
(30) & \emph{CoRL} 2022 & 17.5\% & 16.9\% & 0.6\% & 0.6\% \\
(31) & \emph{CoRL} 2022 & 20.0\% & 19.0\% & 0.6\% & 1.0\% \\
(32) & \emph{CoRL} 2022 & 22.5\% & 21.1\% & 0.6\% & 1.4\% \\
(33) & \emph{CoRL} 2022 & 25.0\% & 23.2\% & 0.6\% & 1.8\% \\
\cmidrule[\heavyrulewidth]{1-6}
\end{tabular}
\end{center}
\vspace{-5mm}
\end{table}
\begin{figure}[htb!] 
    \centering
    \includegraphics[width=0.475\textwidth]{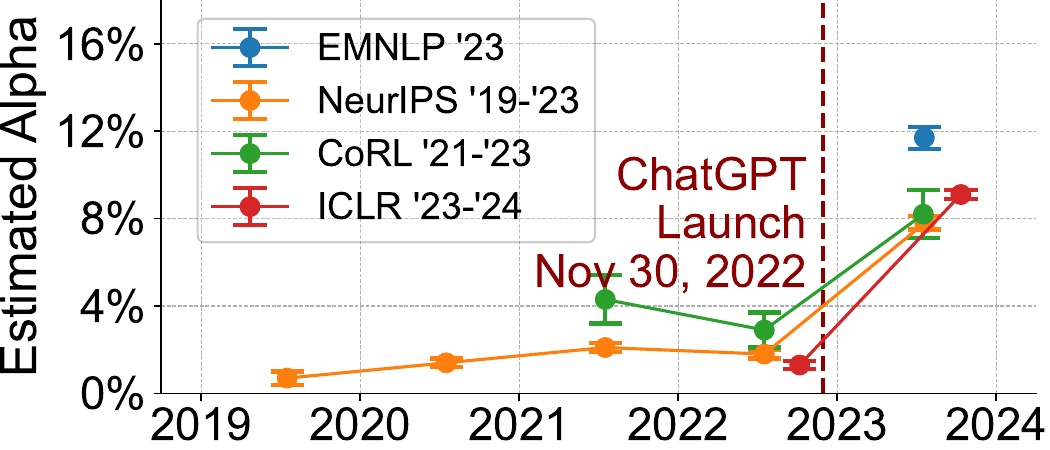}
\caption{
\textbf{Temporal changes in the estimated $\a$ for several ML conferences using adverbs.}}
\label{fig: adv temporal}
\end{figure}
\begin{table}[htb!]
\small
\begin{center}
\caption{Temporal trends in the $\a$ estimate on official reviews using adverbs. The same qualitative trend is observed: $\a$ estimates pre-ChatGPT are close to 0, and there is a sharp increase after the release of ChatGPT.}
\label{tab: adv main}
\begin{tabular}{lrcll}
\cmidrule[\heavyrulewidth]{1-4}
\multirow{2}{*}{\bf No.} 
& \multirow{2}{*}{\bf \begin{tabular}[c]{@{}c@{}} Validation \\ Data Source 
\end{tabular} } 
&\multicolumn{2}{l}{\bf Estimated} 
\\
\cmidrule{3-4}
 & & $\alpha$ & $CI$ ($\pm$) \\
\cmidrule{1-4}
(1) & \emph{NeurIPS} 2019 & 0.7\%  & 0.3\% \\
(2) & \emph{NeurIPS} 2020 & 1.4\%  & 0.2\% \\
(3) & \emph{NeurIPS} 2021 & 2.1\%  & 0.2\% \\
(4) & \emph{NeurIPS} 2022 & 1.8\%  & 0.2\% \\
(5) & \emph{NeurIPS} 2023 & 7.8\% & 0.3\% \\
\cmidrule{1-4} 
(6) & \emph{ICLR} 2023 & 1.3\% & 0.2\% \\
(7) & \emph{ICLR} 2024 & 9.1\% & 0.2\% \\
\cmidrule{1-4} 
(8) & \emph{CoRL} 2021 & 4.3\% & 1.1\% \\
(9) & \emph{CoRL} 2022 & 2.9\% & 0.8\% \\
(10) & \emph{CoRL} 2023 & 8.2\% & 1.1\% \\
\cmidrule{1-4} 
(11) & \emph{EMNLP} 2023 & 11.7\% & 0.5\% \\
\cmidrule[\heavyrulewidth]{1-4}
\end{tabular}
\end{center}
\vspace{-5mm}
\end{table}

\newpage 
\clearpage

\subsection{Results with Verbs}
\label{Appendix:subsec:verbs}
Here, we conducted experiments using verbs. The results for verbs are shown in Table~\ref{tab: verb val}.

\begin{figure}[htb!] 
\centering
\includegraphics[width=1\textwidth]{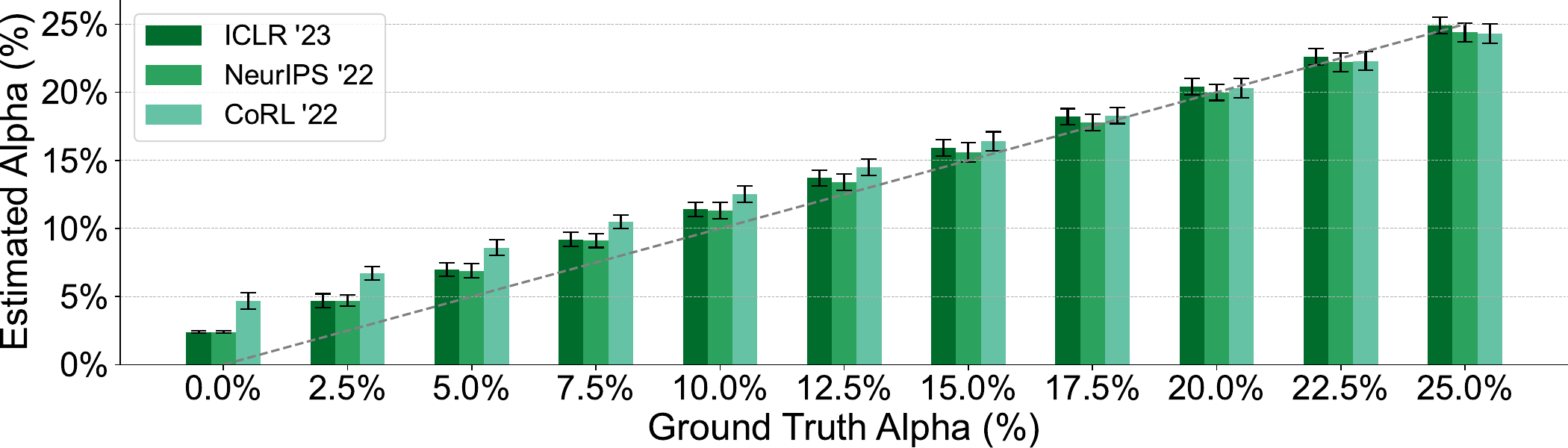}
\caption{
Results of the validation procedure from Section~\ref{sec: val} using verbs (instead of adjectives).
}
\label{fig: verb val}
\end{figure}

\begin{table}[htb!]
\small
\begin{center}
\caption{
Validation accuracy when verbs are used. The performance degrades slightly as compared to using adjectives.
}
\label{tab: verb val}
\begin{tabular}{lrcllc}
\cmidrule[\heavyrulewidth]{1-6}
\multirow{2}{*}{\bf No.} 
& \multirow{2}{*}{\bf \begin{tabular}[c]{@{}c@{}} Validation \\ Data Source 
\end{tabular} } 
& \multirow{2}{*}{\bf \begin{tabular}[c]{@{}c@{}} Ground \\ Truth $\alpha$
\end{tabular}}  
&\multicolumn{2}{l}{\bf Estimated} 
& \multirow{2}{*}{\bf \begin{tabular}[c]{@{}c@{}} Prediction \\ Error 
\end{tabular} } 
\\
\cmidrule{4-5}
 & & & $\alpha$ & $CI$ ($\pm$) & \\
\cmidrule{1-6}
(1) & \emph{ICLR} 2023 & 0.0\% & 2.4\% & 0.1\% & 2.4\% \\
(2) & \emph{ICLR} 2023 & 2.5\% & 4.7\% & 0.5\% & 2.2\% \\
(3) & \emph{ICLR} 2023 & 5.0\% & 7.0\% & 0.5\% & 2.0\% \\
(4) & \emph{ICLR} 2023 & 7.5\% & 9.2\% & 0.5\% & 1.7\% \\
(5) & \emph{ICLR} 2023 & 10.0\% & 11.4\% & 0.5\% & 1.4\% \\
(6) & \emph{ICLR} 2023 & 12.5\% & 13.7\% & 0.6\% & 1.2\% \\
(7) & \emph{ICLR} 2023 & 15.0\% & 15.9\% & 0.6\% & 0.9\% \\
(8) & \emph{ICLR} 2023 & 17.5\% & 18.2\% & 0.6\% & 0.7\% \\
(9) & \emph{ICLR} 2023 & 20.0\% & 20.4\% & 0.6\% & 0.4\% \\
(10) & \emph{ICLR} 2023 & 22.5\% & 22.6\% & 0.6\% & 0.1\% \\
(11) & \emph{ICLR} 2023 & 25.0\% & 24.9\% & 0.6\% & 0.1\% \\
\cmidrule{1-6}
(12) & \emph{NeurIPS} 2022 & 0.0\% & 2.4\% & 0.1\%  & 2.4\%\\
(13) & \emph{NeurIPS} 2022 & 2.5\% & 4.7\% & 0.4\%  & 2.2\%\\
(14) & \emph{NeurIPS} 2022 & 5.0\% & 6.9\% & 0.5\%  & 1.9\%\\
(15) & \emph{NeurIPS} 2022 & 7.5\% & 9.1\% & 0.5\% & 1.6\%\\
(16) & \emph{NeurIPS} 2022 & 10.0\% & 11.3\% & 0.6\% & 1.3\%\\
(17) & \emph{NeurIPS} 2022 & 12.5\% & 13.4\% & 0.6\% & 0.9\%\\
(18) & \emph{NeurIPS} 2022 & 15.0\% & 15.6\% & 0.7\% & 0.6\%\\
(19) & \emph{NeurIPS} 2022 & 17.5\% & 17.8\% & 0.6\% & 0.3\%\\
(20) & \emph{NeurIPS} 2022 & 20.0\% & 20.0\% & 0.6\% & 0.0\%\\
(21) & \emph{NeurIPS} 2022 & 22.5\% & 22.2\% & 0.7\% & 0.3\%\\
(22) & \emph{NeurIPS} 2022 & 25.0\% & 24.4\% & 0.7\% & 0.6\%\\
\cmidrule{1-6}
(23) & \emph{CoRL} 2022 & 0.0\% & 4.7\% & 0.6\% & 4.7\% \\
(24) & \emph{CoRL} 2022 & 2.5\% & 6.7\% & 0.5\% & 4.2\% \\
(25) & \emph{CoRL} 2022 & 5.0\% & 8.6\% & 0.6\% & 3.6\% \\
(26) & \emph{CoRL} 2022 & 7.5\% & 10.5\% & 0.5\% & 3.0\% \\
(27) & \emph{CoRL} 2022 & 10.0\% &12.5\% & 0.6\% & 2.5\% \\
(28) & \emph{CoRL} 2022 & 12.5\% & 14.5\% & 0.6\% & 2.0\% \\
(29) & \emph{CoRL} 2022 & 15.0\% & 16.4\% & 0.7\% & 1.4\% \\
(30) & \emph{CoRL} 2022 & 17.5\% & 18.3\% & 0.6\% & 0.8\% \\
(31) & \emph{CoRL} 2022 & 20.0\% & 20.3\% & 0.7\% & 0.3\% \\
(32) & \emph{CoRL} 2022 & 22.5\% & 22.3\% & 0.7\% & 0.2\% \\
(33) & \emph{CoRL} 2022 & 25.0\% & 24.3\% & 0.7\% & 0.7\% \\
\cmidrule[\heavyrulewidth]{1-6}
\end{tabular}
\end{center}
\vspace{-5mm}
\end{table}

\begin{figure}[ht!] 
    \centering
    \includegraphics[width=0.475\textwidth]{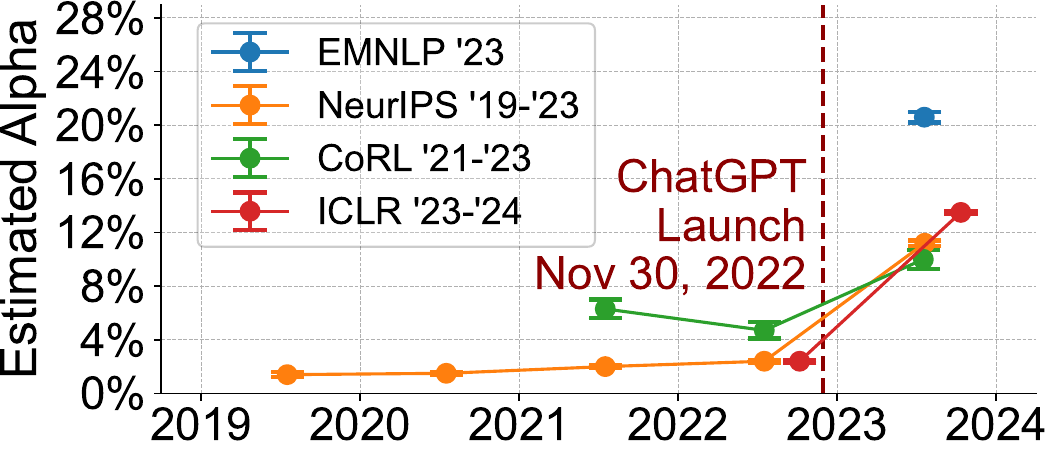}
    \caption{
    \textbf{Temporal changes in the estimated $\a$ for several ML conferences using verbs.}}
    \label{fig: verb temporal}
\end{figure}

\begin{table}[htb!]
\small
\begin{center}
\caption{Temporal trends in the $\a$ estimate on official reviews using verbs. The same qualitative trend is observed: $\a$ estimates pre-ChatGPT are close to 0, and there is a sharp increase after the release of ChatGPT.}
\label{tab: verb main}
\begin{tabular}{lrcll}
\cmidrule[\heavyrulewidth]{1-4}
\multirow{2}{*}{\bf No.} 
& \multirow{2}{*}{\bf \begin{tabular}[c]{@{}c@{}} Validation \\ Data Source 
\end{tabular} } 
&\multicolumn{2}{l}{\bf Estimated} 
\\
\cmidrule{3-4}
 & & $\alpha$ & $CI$ ($\pm$) \\
\cmidrule{1-4}
(1) & \emph{NeurIPS} 2019 & 1.4\%  & 0.2\% \\
(2) & \emph{NeurIPS} 2020 & 1.5\%  & 0.1\% \\
(3) & \emph{NeurIPS} 2021 & 2.0\%  & 0.1\% \\
(4) & \emph{NeurIPS} 2022 & 2.4\%  & 0.1\% \\
(5) & \emph{NeurIPS} 2023 & 11.2\% & 0.2\% \\
\cmidrule{1-4} 
(6) & \emph{ICLR} 2023 & 2.4\% & 0.1\% \\
(7) & \emph{ICLR} 2024 & 13.5\% & 0.1\% \\
\cmidrule{1-4} 
(8) & \emph{CoRL} 2021 & 6.3\% & 0.7\% \\
(9) & \emph{CoRL} 2022 & 4.7\% & 0.6\% \\
(10) & \emph{CoRL} 2023 & 10.0\% & 0.7\% \\
\cmidrule{1-4} 
(11) & \emph{EMNLP} 2023 & 20.6\% & 0.4\% \\
\cmidrule[\heavyrulewidth]{1-4}
\end{tabular}
\end{center}
\vspace{-5mm}
\end{table}

\newpage 
\clearpage

\subsection{Results with Nouns}
\label{Appendix:subsec:nouns}
Here, we conducted experiments using nouns. The results for nouns in Table~\ref{tab: noun val}.

\begin{figure}[ht!] 
\centering
\includegraphics[width=1\textwidth]{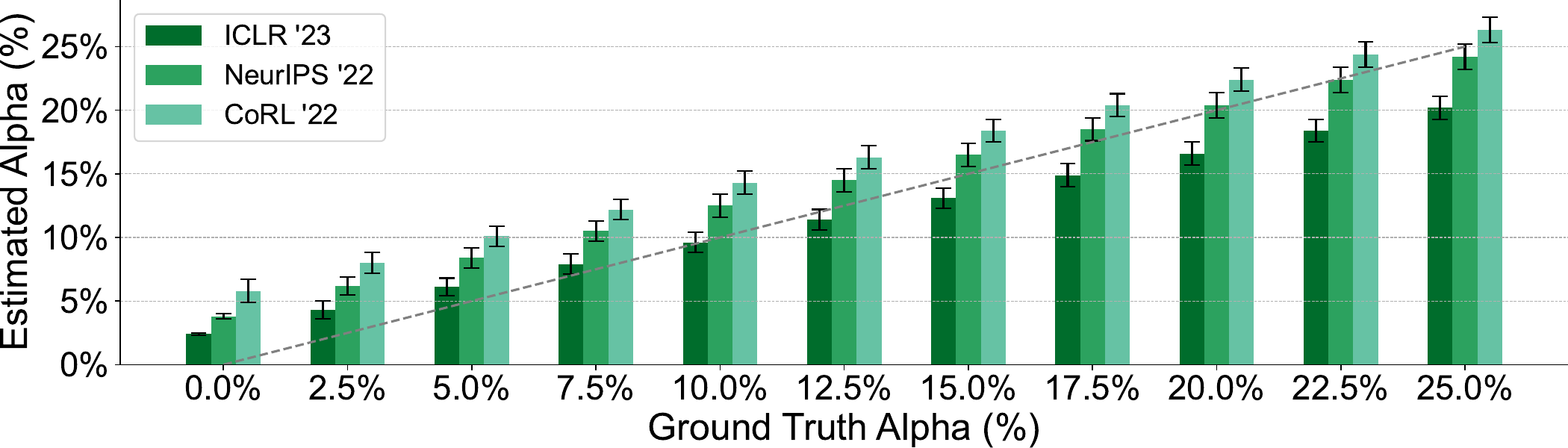}
\caption{
Results of the validation procedure from Section~\ref{sec: val} using nouns (instead of adjectives).
}
\label{fig: noun val}
\end{figure}

\begin{table}[htb!]
\small
\begin{center}
\caption{
Validation accuracy when nouns are used. The performance degrades as compared to using adjectives.
}
\label{tab: noun val}
\begin{tabular}{lrcllc}
\cmidrule[\heavyrulewidth]{1-6}
\multirow{2}{*}{\bf No.} 
& \multirow{2}{*}{\bf \begin{tabular}[c]{@{}c@{}} Validation \\ Data Source 
\end{tabular} } 
& \multirow{2}{*}{\bf \begin{tabular}[c]{@{}c@{}} Ground \\ Truth $\alpha$
\end{tabular}}  
&\multicolumn{2}{l}{\bf Estimated} 
& \multirow{2}{*}{\bf \begin{tabular}[c]{@{}c@{}} Prediction \\ Error 
\end{tabular} } 
\\
\cmidrule{4-5}
 & & & $\alpha$ & $CI$ ($\pm$) & \\
\cmidrule{1-6}
(1) & \emph{ICLR} 2023 & 0.0\% & 2.4\% & 0.1\% & 2.4\% \\
(2) & \emph{ICLR} 2023 & 2.5\% & 4.3\% & 0.7\% & 1.8\% \\
(3) & \emph{ICLR} 2023 & 5.0\% & 6.1\% & 0.7\% & 1.1\% \\
(4) & \emph{ICLR} 2023 & 7.5\% & 7.9\% & 0.8\% & 0.4\% \\
(5) & \emph{ICLR} 2023 & 10.0\% & 9.6\% & 0.8\% & 0.4\% \\
(6) & \emph{ICLR} 2023 & 12.5\% & 11.4\% & 0.8\% & 1.1\% \\
(7) & \emph{ICLR} 2023 & 15.0\% & 13.1\% & 0.8\% & 1.9\% \\
(8) & \emph{ICLR} 2023 & 17.5\% & 14.9\% & 0.9\% & 2.6\% \\
(9) & \emph{ICLR} 2023 & 20.0\% & 16.6\% & 0.9\% & 3.4\% \\
(10) & \emph{ICLR} 2023 & 22.5\% & 18.4\% & 0.9\% & 4.1\% \\
(11) & \emph{ICLR} 2023 & 25.0\% & 20.2\% & 0.9\% & 4.8\% \\
\cmidrule{1-6}
(12) & \emph{NeurIPS} 2022 & 0.0\% & 3.8\% & 0.2\%  & 3.8\%\\
(13) & \emph{NeurIPS} 2022 & 2.5\% & 6.2\% & 0.7\%  & 3.7\%\\
(14) & \emph{NeurIPS} 2022 & 5.0\% & 8.4\% & 0.8\%  & 3.4\%\\
(15) & \emph{NeurIPS} 2022 & 7.5\% & 10.5\% & 0.8\% & 3.0\%\\
(16) & \emph{NeurIPS} 2022 & 10.0\% & 12.5\% & 0.9\% & 2.5\%\\
(17) & \emph{NeurIPS} 2022 & 12.5\% & 14.5\% & 0.9\% & 2.0\%\\
(18) & \emph{NeurIPS} 2022 & 15.0\% & 16.5\% & 0.9\% & 1.5\%\\
(19) & \emph{NeurIPS} 2022 & 17.5\% & 18.5\% & 0.9\% & 1.0\%\\
(20) & \emph{NeurIPS} 2022 & 20.0\% & 20.4\% & 1.0\% & 0.4\%\\
(21) & \emph{NeurIPS} 2022 & 22.5\% & 22.4\% & 1.0\% & 0.1\%\\
(22) & \emph{NeurIPS} 2022 & 25.0\% & 24.2\% & 1.0\% & 0.8\%\\
\cmidrule{1-6}
(23) & \emph{CoRL} 2022 & 0.0\% & 5.8\% & 0.9\% & 5.8\% \\
(24) & \emph{CoRL} 2022 & 2.5\% & 8.0\% & 0.8\% & 5.5\% \\
(25) & \emph{CoRL} 2022 & 5.0\% & 10.1\% & 0.8\% & 5.1\% \\
(26) & \emph{CoRL} 2022 & 7.5\% & 12.2\% & 0.8\% & 4.7\% \\
(27) & \emph{CoRL} 2022 & 10.0\% &14.3\% & 0.9\% & 4.3\% \\
(28) & \emph{CoRL} 2022 & 12.5\% & 16.3\% & 0.9\% & 3.8\% \\
(29) & \emph{CoRL} 2022 & 15.0\% & 18.4\% & 0.9\% & 3.4\% \\
(30) & \emph{CoRL} 2022 & 17.5\% & 20.4\% & 0.9\% & 2.9\% \\
(31) & \emph{CoRL} 2022 & 20.0\% & 22.4\% & 0.9\% & 2.4\% \\
(32) & \emph{CoRL} 2022 & 22.5\% & 24.4\% & 1.0\% & 1.9\% \\
(33) & \emph{CoRL} 2022 & 25.0\% & 26.3\% & 1.0\% & 1.3\% \\
\cmidrule[\heavyrulewidth]{1-6}
\end{tabular}
\end{center}
\vspace{-5mm}
\end{table}

\begin{figure}[ht!] 
    \centering
    \includegraphics[width=0.475\textwidth]{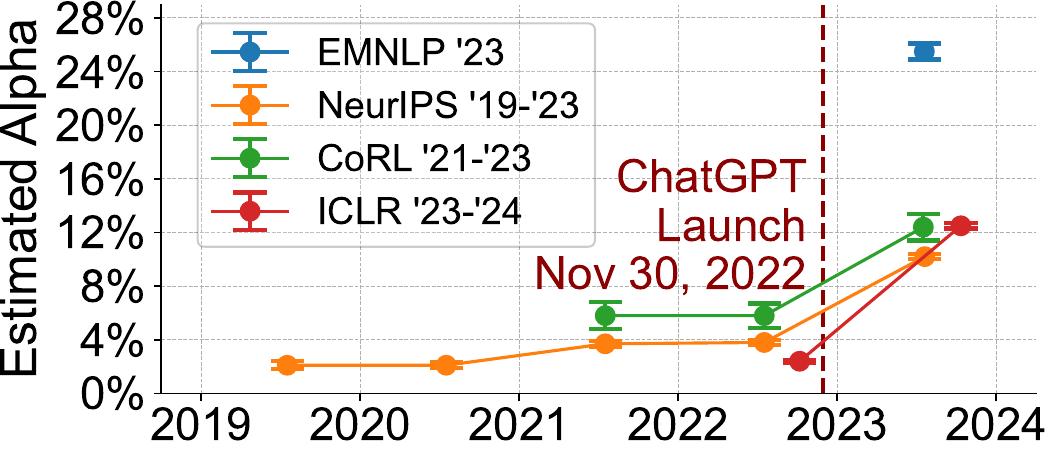}
    \caption{
    \textbf{Temporal changes in the estimated $\a$ for several ML conferences using nouns.}}
    \label{fig: noun temporal}
\end{figure}

\begin{table}[htb!]
\small
\begin{center}
\caption{Temporal trends in the $\a$ estimate on official reviews using nouns. The same qualitative trend is observed: $\a$ estimates pre-ChatGPT are close to 0, and there is a sharp increase after the release of ChatGPT.}
\label{tab: noun main}
\begin{tabular}{lrcll}
\cmidrule[\heavyrulewidth]{1-4}
\multirow{2}{*}{\bf No.} 
& \multirow{2}{*}{\bf \begin{tabular}[c]{@{}c@{}} Validation \\ Data Source 
\end{tabular} } 
&\multicolumn{2}{l}{\bf Estimated} 
\\
\cmidrule{3-4}
 & & $\alpha$ & $CI$ ($\pm$) \\
\cmidrule{1-4}
(1) & \emph{NeurIPS} 2019 & 2.1\%  & 0.3\% \\
(2) & \emph{NeurIPS} 2020 & 2.1\%  & 0.2\% \\
(3) & \emph{NeurIPS} 2021 & 3.7\%  & 0.2\% \\
(4) & \emph{NeurIPS} 2022 & 3.8\%  & 0.2\% \\
(5) & \emph{NeurIPS} 2023 & 10.2\% & 0.2\% \\
\cmidrule{1-4} 
(6) & \emph{ICLR} 2023 & 2.4\% & 0.1\% \\
(7) & \emph{ICLR} 2024 & 12.5\% & 0.2\% \\
\cmidrule{1-4} 
(8) & \emph{CoRL} 2021 & 5.8\% & 1.0\% \\
(9) & \emph{CoRL} 2022 & 5.8\% & 0.9\% \\
(10) & \emph{CoRL} 2023 & 12.4\% & 1.0\% \\
\cmidrule{1-4} 
(11) & \emph{EMNLP} 2023 & 25.5\% & 0.6\% \\
\cmidrule[\heavyrulewidth]{1-4}
\end{tabular}
\end{center}
\vspace{-5mm}
\end{table}


\subsection{Results on Document-Level Analysis}
\label{subsec:Results on Document-Level Analysis}

Our results in the main paper analyzed the data at a sentence level. That is, we assumed that each sentence in a review was drawn from the mixture model \eqref{eq: mix}, and estimated the fraction $\a$ of sentences which were AI generated. We can perform the same analysis on entire \emph{documents} (i.e., complete reviews) to check the robustness of our method to this design choice. Here, $P$ should be interpreted as the distribution of reviews generated without AI assistance, while $Q$ should be interpreted as reviews for which a significant fraction of the content is AI generated. (We do not expect any reviews to be 100\% AI-generated, so this distinction is important.)

The results of the document-level analysis are similar to that at the sentence level. Table~\ref{tab: doc val} shows the validation results corresponding to Section~\ref{sec: val}.

\begin{figure}[ht!] 
\centering
\includegraphics[width=1\textwidth]{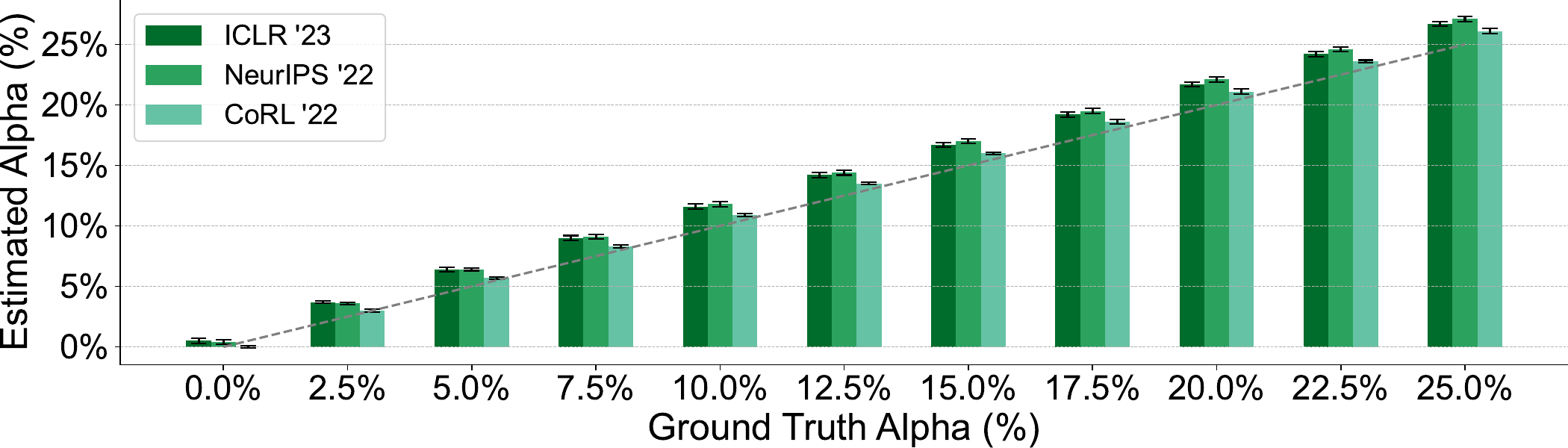}
\caption{
Results of the validation procedure from Section~\ref{sec: val} at a document (rather than sentence) level.
}
\label{fig: doc level val}
\end{figure}

\begin{table}[htb!]
\small
\begin{center}
\caption{
Validation accuracy applying the method at a document (rather than sentence) level. There is a slight degradation in performance compared to the sentence-level approach, and the method tends to slightly over-estimate the true $\a$. We prefer the sentence-level method since it is unlikely that any reviewer will generate an entire review using ChatGPT, as opposed to generating individual sentences or parts of the review using AI.
}
\label{tab: doc val}
\begin{tabular}{lrcllc}
\cmidrule[\heavyrulewidth]{1-6}
\multirow{2}{*}{\bf No.} 
& \multirow{2}{*}{\bf \begin{tabular}[c]{@{}c@{}} Validation \\ Data Source 
\end{tabular} } 
& \multirow{2}{*}{\bf \begin{tabular}[c]{@{}c@{}} Ground \\ Truth $\alpha$
\end{tabular}}  
&\multicolumn{2}{l}{\bf Estimated} 
& \multirow{2}{*}{\bf \begin{tabular}[c]{@{}c@{}} Prediction \\ Error 
\end{tabular} } 
\\
\cmidrule{4-5}
 & & & $\alpha$ & $CI$ ($\pm$) & \\
\cmidrule{1-6}
(1) & \emph{ICLR} 2023 & 0.0\% & 0.5\% & 0.2\% & 0.5\% \\
(2) & \emph{ICLR} 2023 & 2.5\% & 3.7\% & 0.1\% & 1.2\% \\
(3) & \emph{ICLR} 2023 & 5.0\% & 6.4\% & 0.2\% & 1.4\% \\
(4) & \emph{ICLR} 2023 & 7.5\% & 9.0\% & 0.2\% & 1.5\% \\
(5) & \emph{ICLR} 2023 & 10.0\% & 11.6\% & 0.2\% & 1.6\% \\
(6) & \emph{ICLR} 2023 & 12.5\% & 14.2\% & 0.2\% & 1.7\% \\
(7) & \emph{ICLR} 2023 & 15.0\% & 16.7\% & 0.2\% & 1.7\% \\
(8) & \emph{ICLR} 2023 & 17.5\% & 19.2\% & 0.2\% & 1.7\% \\
(9) & \emph{ICLR} 2023 & 20.0\% & 21.7\% & 0.2\% & 1.7\% \\
(10) & \emph{ICLR} 2023 & 22.5\% & 24.2\% & 0.2\% & 1.7\% \\
(11) & \emph{ICLR} 2023 & 25.0\% & 26.7\% & 0.2\% & 1.7\% \\
\cmidrule{1-6}
(12) & \emph{NeurIPS} 2022 & 0.0\% & 0.4\% & 0.2\%  & 0.4\%\\
(13) & \emph{NeurIPS} 2022 & 2.5\% & 3.6\% & 0.1\%  & 1.1\%\\
(14) & \emph{NeurIPS} 2022 & 5.0\% & 6.4\% & 0.1\%  & 1.4\%\\
(15) & \emph{NeurIPS} 2022 & 7.5\% & 9.1\% & 0.2\% & 1.6\%\\
(16) & \emph{NeurIPS} 2022 & 10.0\% & 11.8\% & 0.2\% & 1.8\%\\
(17) & \emph{NeurIPS} 2022 & 12.5\% & 14.4\% & 0.2\% & 1.9\%\\
(18) & \emph{NeurIPS} 2022 & 15.0\% & 17.0\% & 0.2\% & 2.0\%\\
(19) & \emph{NeurIPS} 2022 & 17.5\% & 19.5\% & 0.2\% & 2.0\%\\
(20) & \emph{NeurIPS} 2022 & 20.0\% & 22.1\% & 0.2\% & 2.1\%\\
(21) & \emph{NeurIPS} 2022 & 22.5\% & 24.6\% & 0.2\% & 2.1\%\\
(22) & \emph{NeurIPS} 2022 & 25.0\% & 27.1\% & 0.2\% & 2.1\%\\
\cmidrule{1-6}
(23) & \emph{CoRL} 2022 & 0.0\% & 0.0\% & 0.1\% & 0.0\% \\
(24) & \emph{CoRL} 2022 & 2.5\% & 3.0\% & 0.1\% & 0.5\% \\
(25) & \emph{CoRL} 2022 & 5.0\% & 5.7\% & 0.1\% & 0.7\% \\
(26) & \emph{CoRL} 2022 & 7.5\% & 8.3\% & 0.1\% & 0.8\% \\
(27) & \emph{CoRL} 2022 & 10.0\% &10.9\% & 0.1\% & 0.9\% \\
(28) & \emph{CoRL} 2022 & 12.5\% & 13.5\% & 0.1\% & 1.0\% \\
(29) & \emph{CoRL} 2022 & 15.0\% & 16.0\% & 0.1\% & 1.0\% \\
(30) & \emph{CoRL} 2022 & 17.5\% & 18.6\% & 0.2\% & 1.1\% \\
(31) & \emph{CoRL} 2022 & 20.0\% & 21.1\% & 0.2\% & 1.1\% \\
(32) & \emph{CoRL} 2022 & 22.5\% & 23.6\% & 0.1\% & 1.1\% \\
(33) & \emph{CoRL} 2022 & 25.0\% & 26.1\% & 0.2\% & 1.1\% \\
\cmidrule[\heavyrulewidth]{1-6}
\end{tabular}
\label{app:doc level}
\end{center}
\vspace{-5mm}
\end{table}

\begin{figure}[htb!] 
    \centering
    \includegraphics[width=0.475\textwidth]{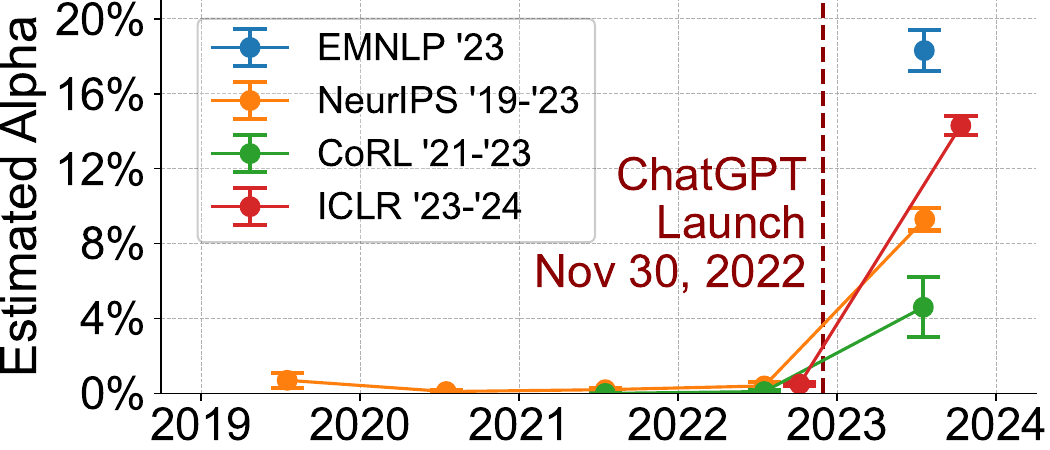}
\caption{
\textbf{Temporal changes in the estimated $\a$ for several ML conferences at the document level.}}
\label{fig: doc temporal}
\end{figure}

\begin{table}[htb!]
\small
\begin{center}
\caption{Temporal trends in the $\a$ estimate on official reviews using the model trained at the document level. The same qualitative trend is observed: $\a$ estimates pre-ChatGPT are close to 0, and there is a sharp increase after the release of ChatGPT.}
\label{tab: document main}
\begin{tabular}{lrcll}
\cmidrule[\heavyrulewidth]{1-4}
\multirow{2}{*}{\bf No.} 
& \multirow{2}{*}{\bf \begin{tabular}[c]{@{}c@{}} Validation \\ Data Source 
\end{tabular} } 
&\multicolumn{2}{l}{\bf Estimated} 
\\
\cmidrule{3-4}
 & & $\alpha$ & $CI$ ($\pm$) \\
\cmidrule{1-4}
(1) & \emph{NeurIPS} 2019 & 0.3\%  & 0.3\% \\
(2) & \emph{NeurIPS} 2020 & 1.1\%  & 0.3\% \\
(3) & \emph{NeurIPS} 2021 & 2.1\%  & 0.2\% \\
(4) & \emph{NeurIPS} 2022 & 3.7\%  & 0.3\% \\
(5) & \emph{NeurIPS} 2023 & 13.7\% & 0.3\% \\
\cmidrule{1-4} 
(6) & \emph{ICLR} 2023 & 3.6\% & 0.2\% \\
(7) & \emph{ICLR} 2024 & 16.3\% & 0.2\% \\
\cmidrule{1-4} 
(8) & \emph{CoRL} 2021 & 2.8\% & 1.1\% \\
(9) & \emph{CoRL} 2022 & 2.9\% & 1.0\% \\
(10) & \emph{CoRL} 2023 & 8.5\% & 1.1\% \\
\cmidrule{1-4} 
(11) & \emph{EMNLP} 2023 & 24.0\% & 0.6\% \\
\cmidrule[\heavyrulewidth]{1-4}
\end{tabular}
\end{center}
\vspace{-5mm}
\end{table}

\newpage 
\clearpage

\subsection{Comparison to State-of-the-art GPT Detection Methods}
\label{Appendix:subsec:baselines}

We conducted experiments using the traditional classification approach to AI text detection. That is, we used two off-the-shelf AI text detectors (RADAR and Deepfake Text Detect) to classify each sentence as AI- or human-generated. Our estimate for $\a$ is the fraction of sentences which the classifier believes are AI-generated. We used the same validation procedure as in Section~\ref{sec: val}. The results are shown in Table~\ref{tab: classifiers}. Two off-the-shelf classifiers predict that either almost all (RADAR) or none (Deepfake) of the text are AI-generated, regardless of the true $\a$ level. With the exception of the BERT-based method, the predictions made by all of the classifiers remain nearly constant across all $\a$ levels, leading to poor performance for all of them. This may be due to a distribution shift between the data used to train the classifier (likely general text scraped from the internet) vs. text found in conference reviews. While BERT's estimates for $\a$ seem at least positively correlated with the correct $\a$ value, the error in the estimate is still large compared to the high accuracy obtained by our method (see Figure~\ref{fig: val} and Table~\ref{tab: adj val}).

\begin{table}[ht!]
\small
\begin{center}
\caption{
Validation accuracy for classifier-based methods. RADAR, Deepfake, and DetectGPT all produce estimates which remain almost constant, independent of the true $\alpha$. The BERT estimates are correlated with the true $\alpha$, but the estimates are still far off.
}
\label{tab: classifiers}
\begin{tabular}{lrcccccr}
\cmidrule[\heavyrulewidth]{1-8}
\multirow{2}{*}{\bf No.} 
& \multirow{2}{*}{\bf \begin{tabular}[c]{@{}c@{}} Validation \\ Data Source 
\end{tabular} } 
& \multirow{2}{*}{\bf \begin{tabular}[c]{@{}c@{}} Ground \\ Truth $\alpha$
\end{tabular}}  
& \multirow{2}{*}{\bf \begin{tabular}[c]{@{}c@{}} RADAR \\ Estimated $\alpha$ 
\end{tabular}} 
& \multirow{2}{*}{\bf \begin{tabular}[c]{@{}c@{}} Deepfake  \\ Estimated $\alpha$ 
\end{tabular}}
& \multirow{2}{*}{\bf \begin{tabular}[c]{@{}c@{}} Fast-DetectGPT \\ Estimated $\alpha$
\end{tabular}}
& \multicolumn{2}{c}{\bf BERT} 
\\
 & & & & & & \bf Estimated $\alpha$ &  \bf Predictor Error\\
\cmidrule{1-8}
(1) & \emph{ICLR} 2023 & 0.0\% & 99.3\% & 0.2\% & 11.3\% & 1.1\% & 1.1\% \\
(2) & \emph{ICLR} 2023 & 2.5\% & 99.4\% & 0.2\% & 11.2\% & 2.9\% & 0.4\% \\
(3) & \emph{ICLR} 2023 & 5.0\% & 99.4\% & 0.3\% & 11.2\% & 4.7\% & 0.3\% \\
(4) & \emph{ICLR} 2023 & 7.5\% & 99.4\% & 0.2\% & 11.4\% & 6.4\% & 1.1\% \\
(5) & \emph{ICLR} 2023 & 10.0\% & 99.4\% & 0.2\% & 11.6\% & 8.0\% & 2.0\% \\
(6) & \emph{ICLR} 2023 & 12.5\% & 99.4\% & 0.3\% & 11.6\% & 9.9\% & 2.6\% \\
(7) & \emph{ICLR} 2023 & 15.0\% & 99.4\% & 0.3\% & 11.8\% & 11.6\% & 3.4\% \\
(8) & \emph{ICLR} 2023 & 17.5\% & 99.4\% & 0.2\% & 11.9\% & 13.4\% & 4.1\% \\
(9) & \emph{ICLR} 2023 & 20.0\% & 99.4\% & 0.3\% & 12.2\% & 15.3\% & 4.7\% \\
(10) & \emph{ICLR} 2023 & 22.5\% & 99.4\% & 0.2\% & 12.0\% & 17.0\% & 5.5\% \\
(11) & \emph{ICLR} 2023 & 25.0\% & 99.4\% & 0.3\% & 12.1\% & 18.8\% & 6.2\% \\
\cmidrule{1-8} 
(12) & \emph{NeurIPS} 2022 & 0.0\% & 99.2\% & 0.2\% & 10.5\% & 1.1\% & 1.1\% \\
(13) & \emph{NeurIPS} 2022 & 2.5\% & 99.2\% & 0.2\% & 10.5\% & 2.3\% & 0.2\% \\
(14) & \emph{NeurIPS} 2022 & 5.0\% & 99.2\% & 0.3\% & 10.7\% & 3.6\% & 1.4\% \\
(15) & \emph{NeurIPS} 2022 & 7.5\% & 99.2\% & 0.2\% & 10.9\% & 5.0\% & 2.5\% \\
(16) & \emph{NeurIPS} 2022 & 10.0\% & 99.2\% & 0.2\% & 10.9\% & 6.1\% & 3.9\% \\
(17) & \emph{NeurIPS} 2022 & 12.5\% & 99.2\% & 0.3\% & 11.1\% & 7.2\% & 5.3\% \\
(18) & \emph{NeurIPS} 2022 & 15.0\% & 99.2\% & 0.3\% & 11.0\% & 8.6\% & 6.4\% \\
(19) & \emph{NeurIPS} 2022 & 17.5\% & 99.3\% & 0.2\% & 11.0\% & 9.9\% & 7.6\% \\
(20) & \emph{NeurIPS} 2022 & 20.0\% & 99.2\% & 0.3\% & 11.3\% & 11.3\% & 8.7\% \\
(21) & \emph{NeurIPS} 2022 & 22.5\% & 99.3\% & 0.2\% & 11.4\% & 12.5\% & 10.0\% \\
(22) & \emph{NeurIPS} 2022 & 25.0\% & 99.2\% & 0.3\% & 11.5\% & 13.8\% & 11.2\% \\
\cmidrule{1-8}
(23) & \emph{CoRL} 2022 & 0.0\% & 99.5\% & 0.2\% & 10.2\% & 1.5\% & 1.5\% \\
(24) & \emph{CoRL} 2022 & 2.5\% & 99.5\% & 0.2\% & 10.4\% & 3.3\% & 0.8\% \\
(25) & \emph{CoRL} 2022 & 5.0\% & 99.5\% & 0.2\% & 10.4\% & 5.0\% & 0.0\% \\
(26) & \emph{CoRL} 2022 & 7.5\% & 99.5\% & 0.3\% & 10.8\% & 6.8\% & 0.7\% \\
(27) & \emph{CoRL} 2022 & 10.0\% &99.5\% & 0.3\% & 11.0\% & 8.4\% & 1.6\% \\
(28) & \emph{CoRL} 2022 & 12.5\% & 99.5\% & 0.3\% & 10.9\% & 10.2\% & 2.3\% \\
(29) & \emph{CoRL} 2022 & 15.0\% & 99.5\% & 0.3\% & 11.1\% & 11.8\% & 3.2\% \\
(30) & \emph{CoRL} 2022 & 17.5\% & 99.5\% & 0.3\% & 11.1\% & 13.8\% & 3.7\% \\
(31) & \emph{CoRL} 2022 & 20.0\% & 99.5\% & 0.3\% & 11.4\% & 15.5\% & 4.5\% \\
(32) & \emph{CoRL} 2022 & 22.5\% & 99.5\% & 0.2\% & 11.6\% & 17.4\% & 5.1\% \\
(33) & \emph{CoRL} 2022 & 25.0\% & 99.5\% & 0.3\% & 11.7\% & 18.9\% & 6.1\% \\
\cmidrule[\heavyrulewidth]{1-8}
\end{tabular}
\label{tab:BERT-and-other-baselines}
\end{center}
\vspace{-5mm}
\end{table}

\begin{table}[htb]
\small
\begin{center}
\caption{
Amortized inference computation cost per 32-token sentence in GFLOPs (total number of floating point operations; $1$ GFLOPs = $10^9$ FLOPs). 
}
\label{tab: simplified_classifiers}
\begin{tabular}{cccccr}
\cmidrule[\heavyrulewidth]{1-5}
\bf \begin{tabular}[c]{@{}c@{}} Ours
\end{tabular}  
& \bf \begin{tabular}[c]{@{}c@{}} RADAR(RoBERTa)
\end{tabular} 
& \bf \begin{tabular}[c]{@{}c@{}} Deepfake(Longformer)
\end{tabular}
& \bf \begin{tabular}[c]{@{}c@{}} Fast-DetectGPT(Zero-shot)
\end{tabular}
& \multicolumn{1}{l}{\bf BERT} \\
\cmidrule{1-5}
6.809 $\times 10^-8$ & 9.671 & 50.781 & 84.669 & 2.721 \\
\cmidrule[\heavyrulewidth]{1-5}
\end{tabular}
\label{table:baseline-computation-cost}
\end{center}
\vspace{-5mm}
\end{table}

\newpage 
\clearpage

\subsection{Robustness to Proofreading}
\begin{table}[ht!]
\centering
\caption{
Proofreading with ChatGPT alone cannot explain the increase. 
}
\label{app: proofread}
\resizebox{0.48\textwidth}{!}{
\setlength{\tabcolsep}{3.5pt}
\begin{tabular}{lrrrr}
\cmidrule[\heavyrulewidth]{1-5}
\textbf{Conferences} & \multicolumn{2}{c}{\textbf{Before Proofread}} & \multicolumn{2}{c}{\textbf{After Proofread}} \\
\cmidrule(lr){2-3} \cmidrule(lr){4-5}
 & $\alpha$ & $CI$ ($\pm$) & $\alpha$ & $CI$ ($\pm$) \\
\cmidrule[\heavyrulewidth]{1-5}
ICLR2023  & 1.5\% & 0.7\% & 2.2\% & 0.8\% \\
NeurIPS2022 & 0.9\% & 0.6\% & 1.5\% & 0.7\% \\
CoRL2022  & 2.3\% & 0.7\% & 3.0\% & 0.8\% \\
\bottomrule
\end{tabular}
}
\end{table}

\subsection{Using LLMs to Substantially Expand Incomplete Sentences}
\begin{table}[htb!]
\small
\begin{center}
\caption{
Validation accuracy using a blend of official human and LLM-expanded review.
}
\label{tab: expand val}
\begin{tabular}{lrcllc}
\cmidrule[\heavyrulewidth]{1-6}
\multirow{2}{*}{\bf No.} 
& \multirow{2}{*}{\bf \begin{tabular}[c]{@{}c@{}} Validation \\ Data Source 
\end{tabular} } 
& \multirow{2}{*}{\bf \begin{tabular}[c]{@{}c@{}} Ground \\ Truth $\alpha$
\end{tabular}}  
&\multicolumn{2}{l}{\bf Estimated} 
& \multirow{2}{*}{\bf \begin{tabular}[c]{@{}c@{}} Prediction \\ Error 
\end{tabular} } 
\\
\cmidrule{4-5}
 & & & $\alpha$ & $CI$ ($\pm$) & \\
\cmidrule{1-6}
(1) & \emph{ICLR} 2023 & 0.0\% & 1.6\% & 0.1\% & 1.6\% \\
(2) & \emph{ICLR} 2023 & 2.5\% & 4.1\% & 0.5\% & 1.6\% \\
(3) & \emph{ICLR} 2023 & 5.0\% & 6.3\% & 0.6\% & 1.3\% \\
(4) & \emph{ICLR} 2023 & 7.5\% & 8.5\% & 0.6\% & 1.0\% \\
(5) & \emph{ICLR} 2023 & 10.0\% & 10.6\% & 0.7\% & 0.6\% \\
(6) & \emph{ICLR} 2023 & 12.5\% & 12.6\% & 0.7\% & 0.1\% \\
(7) & \emph{ICLR} 2023 & 15.0\% & 14.7\% & 0.7\% & 0.3\% \\
(8) & \emph{ICLR} 2023 & 17.5\% & 16.7\% & 0.7\% & 0.8\% \\
(9) & \emph{ICLR} 2023 & 20.0\% & 18.7\% & 0.8\% & 1.3\% \\
(10) & \emph{ICLR} 2023 & 22.5\% & 20.7\% & 0.8\% & 1.8\% \\
(11) & \emph{ICLR} 2023 & 25.0\% & 22.7\% & 0.8\% & 2.3\% \\
\cmidrule{1-6}
(12) & \emph{NeurIPS} 2022 & 0.0\% & 1.9\% & 0.2\%  & 1.9\%\\
(13) & \emph{NeurIPS} 2022 & 2.5\% & 4.0\% & 0.6\%  & 1.5\%\\
(14) & \emph{NeurIPS} 2022 & 5.0\% & 6.0\% & 0.6\%  & 1.0\%\\
(15) & \emph{NeurIPS} 2022 & 7.5\% & 7.9\% & 0.6\% & 0.4\%\\
(16) & \emph{NeurIPS} 2022 & 10.0\% & 9.8\% & 0.6\% & 0.2\%\\
(17) & \emph{NeurIPS} 2022 & 12.5\% & 11.6\% & 0.7\% & 0.9\%\\
(18) & \emph{NeurIPS} 2022 & 15.0\% & 13.4\% & 0.7\% & 1.6\%\\
(19) & \emph{NeurIPS} 2022 & 17.5\% & 15.2\% & 0.8\% & 2.3\%\\
(20) & \emph{NeurIPS} 2022 & 20.0\% & 17.0\% & 0.8\% & 3.0\%\\
(21) & \emph{NeurIPS} 2022 & 22.5\% & 18.8\% & 0.8\% & 3.7\%\\
(22) & \emph{NeurIPS} 2022 & 25.0\% & 20.6\% & 0.8\% & 4.4\%\\
\cmidrule{1-6}
(23) & \emph{CoRL} 2022 & 0.0\% & 2.4\% & 0.6\% & 2.4\% \\
(24) & \emph{CoRL} 2022 & 2.5\% & 4.5\% & 0.5\% & 2.0\% \\
(25) & \emph{CoRL} 2022 & 5.0\% & 6.4\% & 0.6\% & 1.4\% \\
(26) & \emph{CoRL} 2022 & 7.5\% & 8.2\% & 0.6\% & 0.7\% \\
(27) & \emph{CoRL} 2022 & 10.0\% &10.0\% & 0.7\% & 0.0\% \\
(28) & \emph{CoRL} 2022 & 12.5\% & 11.8\% & 0.7\% & 0.7\% \\
(29) & \emph{CoRL} 2022 & 15.0\% & 13.6\% & 0.7\% & 1.4\% \\
(30) & \emph{CoRL} 2022 & 17.5\% & 15.3\% & 0.7\% & 2.2\% \\
(31) & \emph{CoRL} 2022 & 20.0\% & 17.0\% & 0.7\% & 3.0\% \\
(32) & \emph{CoRL} 2022 & 22.5\% & 18.7\% & 0.8\% & 3.8\% \\
(33) & \emph{CoRL} 2022 & 25.0\% & 20.5\% & 0.8\% & 4.5\% \\
\cmidrule[\heavyrulewidth]{1-6}
\end{tabular}
\label{app:LLM-expanded }
\end{center}
\vspace{-5mm}
\end{table}
\newpage
\clearpage

\subsection{Factors that Correlate With Estimated LLM Usage}
\label{sec:factors}
\begin{table}[ht!]
\centering
\caption{Numerical results for the deadline effect (Figure~\ref{fig: deadline}).}
\setlength{\tabcolsep}{3.5pt}
\begin{tabular}{lrrrr}
\cmidrule[\heavyrulewidth]{1-5}
\textbf{Conferences} & \multicolumn{2}{c}{\textbf{\begin{tabular}{@{}c@{}}More than 3 Days \\ Before Review Deadline\end{tabular}}} & \multicolumn{2}{c}{\textbf{\begin{tabular}{@{}c@{}}Within 3 Days 
 \\ of Review Deadline\end{tabular}}} \\
\cmidrule(lr){2-3} \cmidrule(lr){4-5}
 & $\alpha$ & $CI$ ($\pm$) & $\alpha$ & $CI$ ($\pm$) \\
\cmidrule[\heavyrulewidth]{1-5}
ICLR2024  & 8.8\% & 0.4\% & 11.3\% & 0.2\% \\
NeurIPS2023 & 7.7\% & 0.4\% & 9.5\% & 0.3\% \\
CoRL2023  & 3.9\% & 1.3\% & 7.3\% & 0.9\% \\
EMNLP2023  & 14.2\% & 1.0\% & 17.1\% & 0.5\% \\
\bottomrule
\end{tabular}
\label{app:timeline}
\end{table}

\begin{table}[ht!]
\centering
\caption{Numerical results for the reference effect (Figure~\ref{fig: et-al})}
\resizebox{0.48\textwidth}{!}{
\setlength{\tabcolsep}{3.5pt}
\begin{tabular}{lrrrr}
\cmidrule[\heavyrulewidth]{1-5}
\textbf{Conferences} & \multicolumn{2}{c}{\textbf{\begin{tabular}{@{}c@{}}With Reference \end{tabular}}} & \multicolumn{2}{c}{\textbf{\begin{tabular}{@{}c@{}}No Reference\end{tabular}}} \\
\cmidrule(lr){2-3} \cmidrule(lr){4-5}
 & $\alpha$ & $CI$ ($\pm$) & $\alpha$ & $CI$ ($\pm$) \\
\cmidrule[\heavyrulewidth]{1-5}
ICLR2024  & 6.5\% & 0.2\% & 12.8\% & 0.2\% \\
NeurIPS2023 & 5.0\% & 0.4\% & 10.2\% & 0.3\% \\
CoRL2023  & 2.2\% & 1.5\% & 7.1\% & 0.8\% \\
EMNLP2023  & 10.6\% & 1.0\% & 17.7\% & 0.5\% \\
\bottomrule
\end{tabular}
}
\label{app:refere}
\end{table}

\begin{table}[ht!]
\centering
\caption{Numerical results for the low reply effect (Figure~\ref{fig: reply}).}
\setlength{\tabcolsep}{3.5pt}
\begin{tabular}{lrrrr}
\cmidrule[\heavyrulewidth]{1-5}
\textbf{\# of Replies} & \multicolumn{2}{c}{\textbf{ICLR 2024}} & \multicolumn{2}{c}{\textbf{NeurIPS 2023}}  \\
\cmidrule(lr){2-3} \cmidrule(lr){4-5}
 & $\alpha$ & $CI$ ($\pm$) & $\alpha$ & $CI$ ($\pm$) \\
\cmidrule[\heavyrulewidth]{1-5}
0       & 13.3\% & 0.3\% & 12.8\% & 0.6\%\\
1       & 10.6\%  & 0.3\%  & 9.2\% & 0.3\%\\
2       & 6.4\%  & 0.5\% & 5.9\% & 0.5\%\\
3       & 6.7\%  & 1.1\% & 4.6\% & 0.9\%\\
4+      & 3.6\% & 1.1\% & 1.9\% & 1.1\%\\
\bottomrule
\end{tabular}
\label{app:replies}
\end{table}

\begin{table}[ht!]
\centering
\caption{Numerical results for the homogenization effect (Figure~\ref{fig: homog}).}
\setlength{\tabcolsep}{3.5pt}
\begin{tabular}{lrrrr}
\cmidrule[\heavyrulewidth]{1-5}
\textbf{Conferences} & \multicolumn{2}{c}{\textbf{Heterogeneous Reviews}} & \multicolumn{2}{c}{\textbf{Homogeneous Reviews}} \\
\cmidrule(lr){2-3} \cmidrule(lr){4-5}
 & $\alpha$ & $CI$ ($\pm$) & $\alpha$ & $CI$ ($\pm$) \\
\cmidrule[\heavyrulewidth]{1-5}
ICLR2024  & 7.2\% & 0.4\% & 13.1\% & 0.4\% \\
NeurIPS2023 & 6.1\% & 0.4\% & 11.6\% & 0.5\% \\
CoRL2023  & 5.1\% & 1.5\% & 7.6\% & 1.4\% \\
EMNLP2023  & 12.8\% & 0.8\% & 19.6\% & 0.8\% \\
\bottomrule
\end{tabular}
\label{app:similarity}
\end{table}

\begin{table}[ht!]
\centering
\caption{Numerical results for the low confidence effect (Figure~\ref{fig: confidence}).}
\setlength{\tabcolsep}{3.5pt}
\begin{tabular}{lrrrr}
\cmidrule[\heavyrulewidth]{1-5}
\textbf{Conferences} & \multicolumn{2}{c}{\textbf{ Reviews with Low Confidence}} & \multicolumn{2}{c}{\textbf{Reviews with High Confidence}} \\
\cmidrule(lr){2-3} \cmidrule(lr){4-5}
 & $\alpha$ & $CI$ ($\pm$) & $\alpha$ & $CI$ ($\pm$) \\
\cmidrule[\heavyrulewidth]{1-5}
ICLR2024  & 13.2\% & 0.7\% & 10.7\% & 0.2\% \\
NeurIPS2023 & 10.3\% & 0.8\% & 8.9\% & 0.2\% \\
CoRL2023  & 7.8\% & 4.8\% & 6.5\% & 0.7\% \\
EMNLP2023  & 17.6\% & 1.8\% & 16.6\% & 0.5\% \\
\bottomrule
\end{tabular}
\label{app:confidence}
\end{table}

\newpage 
\clearpage

\subsection{Additional Results on GPT-3.5}

Here we chose to focus on ChatGPT because it is by far the most popular in general usage. According to a comprehensive analysis by FlexOS in early 2024, ChatGPT dominates the generative AI market, with 76\% of global internet traffic in the category. Bing AI follows with 16\%, Bard with 7\%, and Claude with 1\%~\cite{vanrossum2024generative}. Recent studies have also found that GPT-4 substantially outperforms other LLMs, including Bard, in the reviewing of scientific papers or proposals \cite{liu2023reviewergpt}.

\begin{figure}[ht!] 
\centering
\includegraphics[width=1\textwidth]{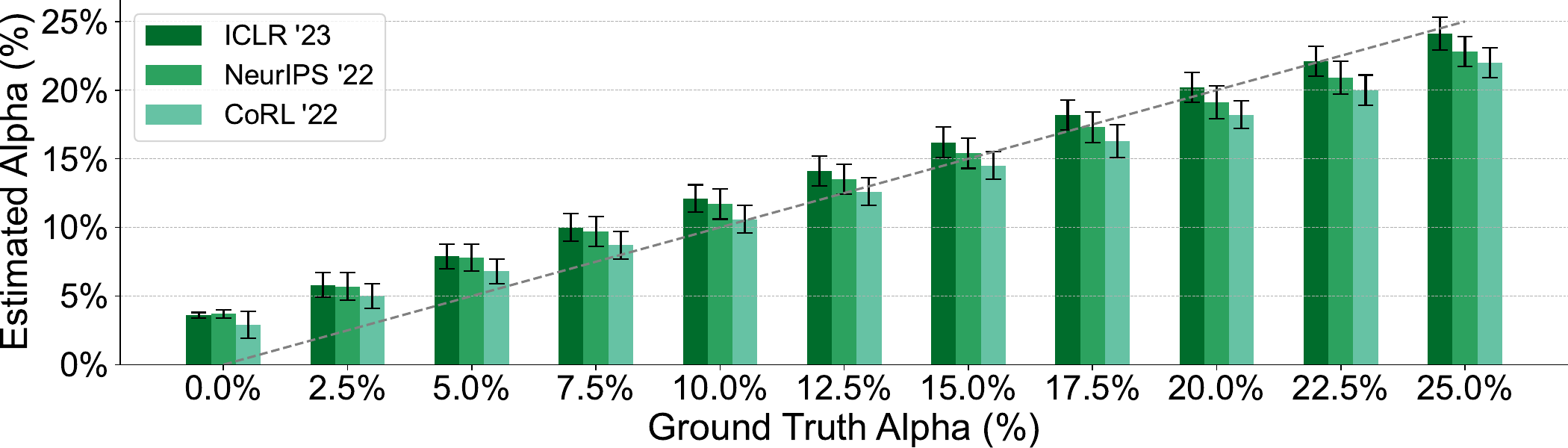}
\caption{
Results of the validation procedure from Section~\ref{sec: val}(model trained on reviews generated by GPT-3.5 and tested on reviews generated by GPT-3.5).
}
\label{fig: 3.5-3.5 val}
\end{figure}

\begin{table}[htb!]
\small
\begin{center}
\caption{
\textbf{Performance validation of the model trained on reviews generated by GPT-3.5.}
}
\label{tab: Performance validation of GPT-3.5 by 3.5}
\begin{tabular}{lrcllc}
\cmidrule[\heavyrulewidth]{1-6}
\multirow{2}{*}{\bf No.} 
& \multirow{2}{*}{\bf \begin{tabular}[c]{@{}c@{}} Validation \\ Data Source 
\end{tabular} } 
& \multirow{2}{*}{\bf \begin{tabular}[c]{@{}c@{}} Ground \\ Truth $\alpha$
\end{tabular}}  
&\multicolumn{2}{l}{\bf Estimated} 
& \multirow{2}{*}{\bf \begin{tabular}[c]{@{}c@{}} Prediction \\ Error 
\end{tabular} } 
\\
\cmidrule{4-5}
 & & & $\alpha$ & $CI$ ($\pm$) & \\
\cmidrule{1-6}
(1) & \emph{ICLR} 2023 & 0.0\% & 3.6\% & 0.2\% & 3.6\% \\
(2) & \emph{ICLR} 2023 & 2.5\% & 5.8\% & 0.9\% & 3.3\% \\
(3) & \emph{ICLR} 2023 & 5.0\% & 7.9\% & 0.9\% & 2.9\% \\
(4) & \emph{ICLR} 2023 & 7.5\% & 10.0\% & 1.0\% & 2.5\% \\
(5) & \emph{ICLR} 2023 & 10.0\% & 12.1\% & 1.0\% & 2.1\% \\
(6) & \emph{ICLR} 2023 & 12.5\% & 14.1\% & 1.1\% & 1.6\% \\
(7) & \emph{ICLR} 2023 & 15.0\% & 16.2\% & 1.1\% & 1.2\% \\
(8) & \emph{ICLR} 2023 & 17.5\% & 18.2\% & 1.1\% & 0.7\% \\
(9) & \emph{ICLR} 2023 & 20.0\% & 20.2\% & 1.1\% & 0.2\% \\
(10) & \emph{ICLR} 2023 & 22.5\% & 22.1\% & 1.1\% & 0.4\% \\
(11) & \emph{ICLR} 2023 & 25.0\% & 24.1\% & 1.2\% & 0.9\% \\
\cmidrule{1-6}
(12) & \emph{NeurIPS} 2022 & 0.0\% & 3.7\% & 0.3\% & 3.7\% \\
(13) & \emph{NeurIPS} 2022 & 2.5\% & 5.7\% & 1.0\% & 3.2\% \\
(14) & \emph{NeurIPS} 2022 & 5.0\% & 7.8\% & 1.0\% & 2.8\% \\
(15) & \emph{NeurIPS} 2022 & 7.5\% & 9.7\% & 1.1\% & 2.2\% \\
(16) & \emph{NeurIPS} 2022 & 10.0\% & 11.7\% & 1.1\% & 1.7\% \\
(17) & \emph{NeurIPS} 2022 & 12.5\% & 13.5\% & 1.1\% & 1.0\% \\
(18) & \emph{NeurIPS} 2022 & 15.0\% & 15.4\% & 1.1\% & 0.4\% \\
(19) & \emph{NeurIPS} 2022 & 17.5\% & 17.3\% & 1.1\% & 0.2\% \\
(20) & \emph{NeurIPS} 2022 & 20.0\% & 19.1\% & 1.2\% & 0.9\% \\
(21) & \emph{NeurIPS} 2022 & 22.5\% & 20.9\% & 1.2\% & 1.6\% \\
(22) & \emph{NeurIPS} 2022 & 25.0\% & 22.8\% & 1.1\% & 2.2\% \\
\cmidrule{1-6}
(23) & \emph{CoRL} 2022 & 0.0\% & 2.9\% & 1.0\% & 2.9\% \\
(24) & \emph{CoRL} 2022 & 2.5\% & 5.0\% & 0.9\% & 2.5\% \\
(25) & \emph{CoRL} 2022 & 5.0\% & 6.8\% & 0.9\% & 1.8\% \\
(26) & \emph{CoRL} 2022 & 7.5\% & 8.7\% & 1.0\% & 1.2\% \\
(27) & \emph{CoRL} 2022 & 10.0\% &10.6\% & 1.0\% & 0.6\% \\
(28) & \emph{CoRL} 2022 & 12.5\% & 12.6\% & 1.0\% & 0.1\% \\
(29) & \emph{CoRL} 2022 & 15.0\% & 14.5\% & 1.0\% & 0.5\% \\
(30) & \emph{CoRL} 2022 & 17.5\% & 16.3\% & 1.2\% & 1.2\% \\
(31) & \emph{CoRL} 2022 & 20.0\% & 18.2\% & 1.0\% & 1.8\% \\
(32) & \emph{CoRL} 2022 & 22.5\% & 20.0\% & 1.1\% & 2.5\% \\
(33) & \emph{CoRL} 2022 & 25.0\% & 22.0\% & 1.1\% & 3.0\% \\
\cmidrule[\heavyrulewidth]{1-6}
\end{tabular}
\label{table:GPT-3.5-validation}
\end{center}
\vspace{-5mm}
\end{table}

\begin{figure}[ht!] 
\centering
\includegraphics[width=1\textwidth]{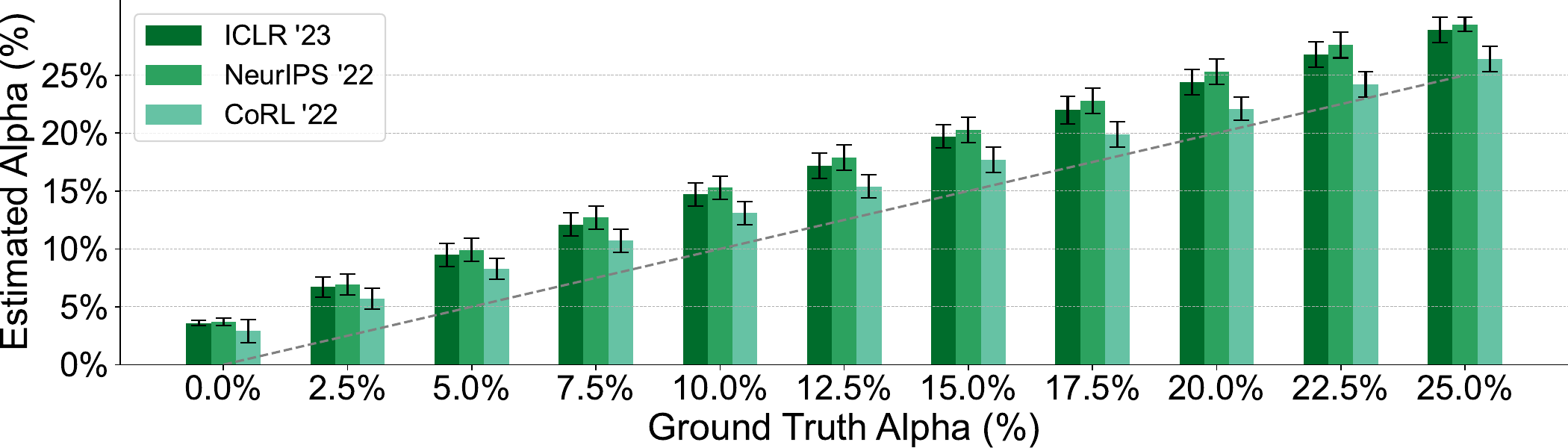}
\caption{
Results of the validation procedure from Section~\ref{sec: val}(model trained on reviews generated by GPT-3.5 and tested on reviews generated by GPT-4).
}
\label{fig: 3.5-4 val}
\end{figure}

\begin{table}[htb!]
\small
\begin{center}
\caption{
\textbf{Performance validation of GPT-4 AI reviews trained on reviews generated by GPT-3.5.}
}
\label{tab: Performance validation of GPT-4 by 3.5}
\begin{tabular}{lrcllc}
\cmidrule[\heavyrulewidth]{1-6}
\multirow{2}{*}{\bf No.} 
& \multirow{2}{*}{\bf \begin{tabular}[c]{@{}c@{}} Validation \\ Data Source 
\end{tabular} } 
& \multirow{2}{*}{\bf \begin{tabular}[c]{@{}c@{}} Ground \\ Truth $\alpha$
\end{tabular}}  
&\multicolumn{2}{l}{\bf Estimated} 
& \multirow{2}{*}{\bf \begin{tabular}[c]{@{}c@{}} Prediction \\ Error 
\end{tabular} } 
\\
\cmidrule{4-5}
 & & & $\alpha$ & $CI$ ($\pm$) & \\
\cmidrule{1-6}
(1) & \emph{ICLR} 2023 & 0.0\% & 3.6\% & 0.2\% & 3.6\% \\
(2) & \emph{ICLR} 2023 & 2.5\% & 6.7\% & 0.9\% & 4.2\% \\
(3) & \emph{ICLR} 2023 & 5.0\% & 9.5\% & 1.0\% & 4.5\% \\
(4) & \emph{ICLR} 2023 & 7.5\% & 12.1\% & 1.0\% & 4.6\% \\
(5) & \emph{ICLR} 2023 & 10.0\% & 14.7\% & 1.0\% & 4.7\% \\
(6) & \emph{ICLR} 2023 & 12.5\% & 17.2\% & 1.1\% & 4.7\% \\
(7) & \emph{ICLR} 2023 & 15.0\% & 19.7\% & 1.0\% & 4.7\% \\
(8) & \emph{ICLR} 2023 & 17.5\% & 22.0\% & 1.2\% & 4.5\% \\
(9) & \emph{ICLR} 2023 & 20.0\% & 24.4\% & 1.1\% & 4.4\% \\
(10) & \emph{ICLR} 2023 & 22.5\% & 26.8\% & 1.1\% & 4.3\% \\
(11) & \emph{ICLR} 2023 & 25.0\% & 28.9\% & 1.1\% & 3.9\% \\
\cmidrule{1-6}
(12) & \emph{NeurIPS} 2022 & 0.0\% & 3.7\% & 0.3\% & 3.7\% \\
(13) & \emph{NeurIPS} 2022 & 2.5\% & 6.9\% & 0.9\% & 4.4\% \\
(14) & \emph{NeurIPS} 2022 & 5.0\% & 9.9\% & 1.0\% & 4.9\% \\
(15) & \emph{NeurIPS} 2022 & 7.5\% & 12.7\% & 1.0\% & 5.2\% \\
(16) & \emph{NeurIPS} 2022 & 10.0\% & 15.3\% & 1.0\% & 5.3\% \\
(17) & \emph{NeurIPS} 2022 & 12.5\% & 17.9\% & 1.1\% & 5.4\% \\
(18) & \emph{NeurIPS} 2022 & 15.0\% & 20.3\% & 1.1\% & 5.3\% \\
(19) & \emph{NeurIPS} 2022 & 17.5\% & 22.8\% & 1.1\% & 5.3\% \\
(20) & \emph{NeurIPS} 2022 & 20.0\% & 25.3\% & 1.1\% & 5.3\% \\
(21) & \emph{NeurIPS} 2022 & 22.5\% & 27.6\% & 1.1\% & 5.1\% \\
(22) & \emph{NeurIPS} 2022 & 25.0\% & 29.4\% & 0.6\% & 4.4\% \\
\cmidrule{1-6}
(23) & \emph{CoRL} 2022 & 0.0\% & 2.9\% & 1.0\% & 2.9\% \\
(24) & \emph{CoRL} 2022 & 2.5\% & 5.7\% & 0.9\% & 3.2\% \\
(25) & \emph{CoRL} 2022 & 5.0\% & 8.3\% & 0.9\% & 3.3\% \\
(26) & \emph{CoRL} 2022 & 7.5\% & 10.7\% & 1.0\% & 3.2\% \\
(27) & \emph{CoRL} 2022 & 10.0\% &13.1\% & 1.0\% & 3.1\% \\
(28) & \emph{CoRL} 2022 & 12.5\% & 15.4\% & 1.0\% & 2.9\% \\
(29) & \emph{CoRL} 2022 & 15.0\% & 17.7\% & 1.1\% & 2.7\% \\
(30) & \emph{CoRL} 2022 & 17.5\% & 19.9\% & 1.1\% & 2.4\% \\
(31) & \emph{CoRL} 2022 & 20.0\% & 22.1\% & 1.0\% & 2.1\% \\
(32) & \emph{CoRL} 2022 & 22.5\% & 24.2\% & 1.1\% & 1.7\% \\
(33) & \emph{CoRL} 2022 & 25.0\% & 26.4\% & 1.1\% & 1.4\% \\
\cmidrule[\heavyrulewidth]{1-6}
\end{tabular}
\label{table:GPT-3.5-validation-on-GPT-4}
\end{center}
\vspace{-5mm}
\end{table}

\begin{figure}[htb!] 
    \centering
    \includegraphics[width=0.475\textwidth]{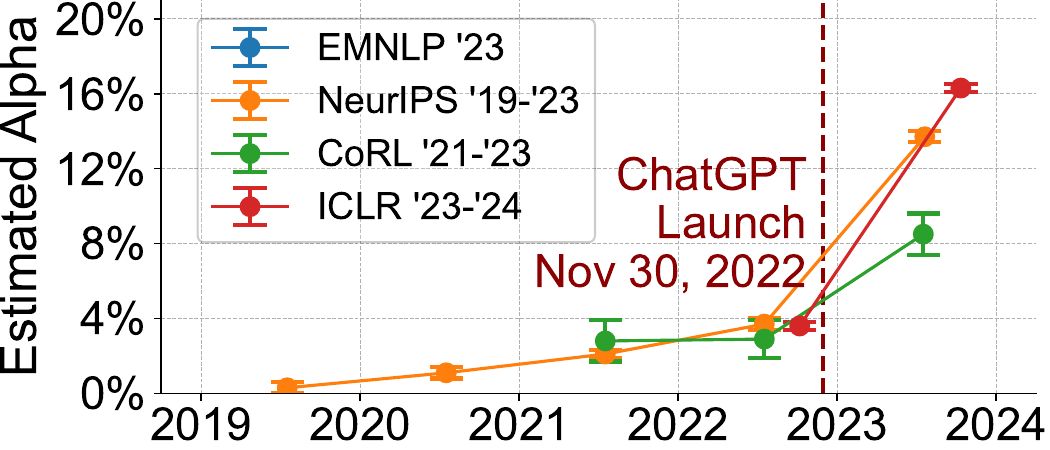}
\caption{
\textbf{Temporal changes in the estimated $\a$ for several ML conferences using the model trained on reviews generated by GPT-3.5.}}
\label{fig: GPT3 temporal}
\end{figure}

\begin{table}[htb!]
\small
\begin{center}
\caption{Temporal trends in the $\a$ estimate on official reviews using the model trained on reviews generated by GPT-3.5. The same qualitative trend is observed: $\a$ estimates pre-ChatGPT are close to 0, and there is a sharp increase after the release of ChatGPT.}
\label{tab: gpt-3.5 main}
\begin{tabular}{lrcll}
\cmidrule[\heavyrulewidth]{1-4}
\multirow{2}{*}{\bf No.} 
& \multirow{2}{*}{\bf \begin{tabular}[c]{@{}c@{}} Validation \\ Data Source 
\end{tabular} } 
&\multicolumn{2}{l}{\bf Estimated} 
\\
\cmidrule{3-4}
 & & $\alpha$ & $CI$ ($\pm$) \\
\cmidrule{1-4}
(1) & \emph{NeurIPS} 2019 & 0.3\%  & 0.3\% \\
(2) & \emph{NeurIPS} 2020 & 1.1\%  & 0.3\% \\
(3) & \emph{NeurIPS} 2021 & 2.1\%  & 0.2\% \\
(4) & \emph{NeurIPS} 2022 & 3.7\%  & 0.3\% \\
(5) & \emph{NeurIPS} 2023 & 13.7\% & 0.3\% \\
\cmidrule{1-4} 
(6) & \emph{ICLR} 2023 & 3.6\% & 0.2\% \\
(7) & \emph{ICLR} 2024 & 16.3\% & 0.2\% \\
\cmidrule{1-4} 
(8) & \emph{CoRL} 2021 & 2.8\% & 1.1\% \\
(9) & \emph{CoRL} 2022 & 2.9\% & 1.0\% \\
(10) & \emph{CoRL} 2023 & 8.5\% & 1.1\% \\
\cmidrule{1-4} 
(11) & \emph{EMNLP} 2023 & 24.0\% & 0.6\% \\
\cmidrule[\heavyrulewidth]{1-4}
\end{tabular}
\end{center}
\vspace{-5mm}
\end{table}


\newpage
\clearpage 

\section{LLM prompts used in the study}

\begin{figure}[htb!]
\begin{lstlisting}
Your task is to write a review given some text of a paper. Your output should be like the following format:
Summary:

Strengths And Weaknesses:

Summary Of The Review:
\end{lstlisting}
\caption{
Example system prompt for generating training data. Paper contents are provided as the user message. 
}
\label{fig:training-prompt}
\end{figure}

\begin{figure}[htb!]
\begin{lstlisting}
Your task now is to draft a high-quality review for CoRL on OpenReview for a submission titled <Title>:

```
<Paper_content>
```

======
Your task: 
Compose a high-quality peer review of a paper submitted to CoRL on OpenReview.

Start by "Review outline:".
And then: 
"1. Summary", Briefly summarize the paper and its contributions. This is not the place to critique the paper; the authors should generally agree with a well-written summary. DO NOT repeat the paper title. 

"2. Strengths", A substantive assessment of the strengths of the paper, touching on each of the following dimensions: originality, quality, clarity, and significance. We encourage reviewers to be broad in their definitions of originality and significance. For example, originality may arise from a new definition or problem formulation, creative combinations of existing ideas, application to a new domain, or removing limitations from prior results. You can incorporate Markdown and Latex into your review. See https://openreview.net/faq.

"3. Weaknesses", A substantive assessment of the weaknesses of the paper. Focus on constructive and actionable insights on how the work could improve towards its stated goals. Be specific, avoid generic remarks. For example, if you believe the contribution lacks novelty, provide references and an explanation as evidence; if you believe experiments are insufficient, explain why and exactly what is missing, etc.

"4. Suggestions", Please list up and carefully describe any suggestions for the authors. Think of the things where a response from the author can change your opinion, clarify a confusion or address a limitation. This is important for a productive rebuttal and discussion phase with the authors.

\end{lstlisting}
\caption{
Example prompt for generating validation data with prompt shift. 
Note that although this validation prompt is written in a significantly different style than the prompt for generating the training data, our algorithm still predicts the alpha accurately. 
}
\label{fig:validation-prompt-shift-prompt}
\end{figure}

\newpage 
\clearpage

\begin{figure}[htb!]
\begin{lstlisting}
The aim here is to reverse-engineer the reviewer's writing process into two distinct phases: drafting a skeleton (outline) of the review and then expanding this outline into a detailed, complete review. The process simulates how a reviewer might first organize their thoughts and key points in a structured, concise form before elaborating on each point to provide a comprehensive evaluation of the paper.


Now as a first step, given a complete peer review, reverse-engineer it into a concise skeleton.
\end{lstlisting}
\caption{
Example prompt for reverse-engineering a given official review into a skeleton (outline) to simulate how a human reviewer might first organize their thoughts and key points in a structured, concise form before elaborating on each point to provide a comprehensive evaluation of the paper.
}
\label{fig:skeleton-prompt-1}
\end{figure}

\begin{figure}[htb!]
\begin{lstlisting}
Expand the skeleton of the review into a official review as the following format:
Summary:

Strengths:

Weaknesses:

Questions:
\end{lstlisting}
\caption{
Example prompt for elaborating the skeleton (outline) into the full review. The format of a review varies depending on the conference.
The goal is to simulate how a human reviewer might first organize their thoughts and key points in a structured, concise form, and then elaborate on each point to provide a comprehensive evaluation of the paper.
}
\label{fig:skeleton-prompt-2}
\end{figure}

\begin{figure}[htb!]
\begin{lstlisting}
Your task is to proofread the provided sentence for grammatical accuracy. Ensure that the corrections introduce minimal distortion to the original content. 
\end{lstlisting}
\caption{
Example prompt for proofreading.
}
\label{fig:proofread-prompt}
\end{figure}

\newpage 
\clearpage
\section{Additional Information on LLM Parameter Settings}

We used the snapshot of GPT-4 from June 13th, 2023 (gpt-4-0613), for our experiments because this is the exact version of ChatGPT that was available during the peer review process of ICLR 2024, NeurIPS 2023, EMNLP 2023, and CoRL 2023.

Regarding the parameter settings, during our experiments, we set the decoding temperature to 1.0 and the maximum decoding length to 2048. We set the Top P hyperparameter to 1.0 and both frequency penalty and presence penalty to 0.0. Additionally, we did not configure any stop sequences during decoding.

\section{Additional Dataset Information}
\label{sec:Additional Dataset Information}

All the data are publicly available. For the machine learning conferences, we accessed peer review data through the official OpenReview API (\url{https://docs.openreview.net/reference/api-v2}), specifically the \url{/notes} endpoint. Each review contains an average of 25.94 sentences. For the Nature portfolio dataset, we developed a custom web scraper using python to access the article pages of 15 journals from the Nature portfolio, extracting peer reviews from papers accepted between 2019 and 2023. Each review in the Nature dataset comprises an average of 37.03 sentences.


The Nature portfolio dataset encompasses the following 15 Nature journals: 
Nature, 
Nature Communications, 
Nature Ecology \& Evolution, 
Nature Structural \& Molecular Biology, 
Nature Cell Biology, 
Nature Human Behaviour, 
Nature Immunology, 
Nature Microbiology, 
Nature Biomedical Engineering, 
Communications Earth \& Environment, 
Communications Biology, 
Communications Physics, 
Communications Chemistry,
Communications Materials,
and Communications Medicine. 
To create this dataset, we systematically accessed the web pages of the selected Nature portfolio journals, extracting peer reviews from papers accepted between 2019 and 2023. In total, our dataset comprises 25,382 peer reviews from 10,242 papers. We chose to focus on the Nature family journals for our baseline dataset due to their reputation for publishing high-quality, impactful research across multiple disciplines. 

Our framework breaks reviews down into a list of sentences, and the parameterization operates at the sentence level. We consider all sentences with 2 or more words and did not set a maximum limit for the number of words in a sentence. If reviewers leave a section blank, no sentences from that section are added to the corpus.

\begin{table}[ht!]
\centering
\caption{
Human Peer Reviews Data from Nature Family Journals.
}
\label{tab:nature_reviews}
\resizebox{0.96\textwidth}{!}{ 
\setlength{\tabcolsep}{3.5pt} 
\begin{tabular}{lcccc}
\toprule
\bf Journal & \bf Post ChatGPT & \bf Data Split & \bf \# of Papers & \bf \# of Official Reviews \\
\midrule
\rowcolor{green!10} 
Nature Portfolio 2022 (random split subset) & \bf \textcolor{darkgreen!70}{Before} & \bf \cellcolor{green!20} \textcolor{black!85}{Training} & 1,189 & 3,341 \\
\cmidrule{1-5} 
\rowcolor{green!10} 
Nature Portfolio 2019 & \bf \textcolor{darkgreen!70}{Before} & \bf \cellcolor{blue!10} \textcolor{blue!85}{Validation} & 2,141 & 4,394 \\
\rowcolor{green!10} 
Nature Portfolio 2020 & \bf \textcolor{darkgreen!70}{Before} & \bf \cellcolor{blue!10} \textcolor{blue!85}{Validation} & 2,083 & 4,736 \\
\rowcolor{green!10} 
Nature Portfolio 2021 & \bf \textcolor{darkgreen!70}{Before} & \bf \cellcolor{blue!10} \textcolor{blue!85}{Validation} & 2,129 & 5,264 \\
\rowcolor{green!10} 
Nature Portfolio 2022 & \bf \textcolor{darkgreen!70}{Before} & \bf \cellcolor{blue!10} \textcolor{blue!85}{Validation} & 511 & 1,447 \\
\midrule
\rowcolor{red!20} 
Nature Portfolio 2022-2023 & \bf \textcolor{red!70}{After} & \bf \textcolor{black!85}{Inference} & 2,189 & 6,200 \\
\bottomrule
\end{tabular}
\label{table:nature-data-table}
}
\end{table}

\paragraph{Ethics Considerations About LLM Analysis for Public Conferences}

The use of peer review data for research purposes raises important ethical considerations around reviewer consent, data licensing, and responsible use~\cite{dycke-etal-2023-nlpeer}. While early datasets have enabled valuable research, going forward, it is critical that the community establishes clear best practices for the ethical collection and use of peer review data.

OpenReview is a science communication initiative which aims to make the scientific process more transparent. Authors and peer reviewers agree to make their reviews public upon submission. In our work, we accessed this publicly available, anonymous peer review data through the public OpenReview API and confirm that we have complied with their terms of use. 

Efforts such as the data donation initiative at ACL Rolling Review (ARR), which requires explicit consent from authors and reviewers and provides clear data licenses~\cite{dycke-etal-2022-yes}, provide a promising model for the future. A key strength of our proposed framework is that it operates at the population level and only outputs aggregate statistics, without the need to perform inference on individual reviews. This helps protect the anonymity of reviewers and mitigates the risk of de-anonymization based on writing style, which is an important consideration when working with peer review data.

We have aimed to use the available data responsibly and ethically in our work. We also recognize the importance of developing robust community norms around the appropriate collection, licensing, sharing, and use of peer review datasets.

\newpage 
\clearpage
\section{Additional Results on LLaMA-2 Chat (70B), and Claude 2.1}

We have added additional validation experiments to test two additional models other than GPT-4: LLaMA-2 Chat (70B) and Claude-2.1. We used the same training and validation setup as our paper. We trained the estimator on ICLR 2018-2022 data, and performed the validation of different alphas on ICLR 2023 data. 
In the first experiment, we trained an estimator using data generated by LLaMA-2 Chat (70B). 
In the second experiment, we trained an estimator using data generated by Claude 2.1. 
As shown in the two results tables, our framework predicts the proportion of AI data (i.e., alpha) very well.

\begin{table}[htb!]
\small
\begin{center}
\caption{
\textbf{Validation Data Source Performance Comparison for LLaMA-2 Chat (70B).}
}
\begin{tabular}{lccccc}
\cmidrule[\heavyrulewidth]{1-5}
\multirow{2}{*}{\bf No.} 
& \multirow{2}{*}{\bf \begin{tabular}[c]{@{}c@{}} Validation \\ Data Source \end{tabular}} 
& \multirow{2}{*}{\bf \begin{tabular}[c]{@{}c@{}} Ground \\ Truth $\alpha$ \end{tabular}}  
& \multicolumn{2}{c}{\bf Estimated} \\
\cmidrule{4-5}
& & & $\alpha$ & $CI$ ($\pm$) \\
\cmidrule[\heavyrulewidth]{1-5}
(1) & \emph{ICLR} 2023 (LLaMA-2 Chat (70B)) & 0\% & 2.8\% & 0.5\% \\
(2) & \emph{ICLR} 2023 (LLaMA-2 Chat (70B)) & 2.5\% & 5.3\% & 0.5\% \\
(3) & \emph{ICLR} 2023 (LLaMA-2 Chat (70B)) & 5\% & 7.6\% & 0.5\% \\
(4) & \emph{ICLR} 2023 (LLaMA-2 Chat (70B)) & 7.5\% & 9.9\% & 0.5\% \\
(5) & \emph{ICLR} 2023 (LLaMA-2 Chat (70B)) & 10\% & 12.2\% & 0.6\% \\
(6) & \emph{ICLR} 2023 (LLaMA-2 Chat (70B)) & 12.5\% & 14.6\% & 0.6\% \\
(7) & \emph{ICLR} 2023 (LLaMA-2 Chat (70B)) & 15\% & 17\% & 0.6\% \\
(8) & \emph{ICLR} 2023 (LLaMA-2 Chat (70B)) & 17.5\% & 19.2\% & 0.6\% \\
(9) & \emph{ICLR} 2023 (LLaMA-2 Chat (70B)) & 20\% & 21.6\% & 0.7\% \\
(10) & \emph{ICLR} 2023 (LLaMA-2 Chat (70B)) & 22.5\% & 24\% & 0.7\% \\
(11) & \emph{ICLR} 2023 (LLaMA-2 Chat (70B)) & 25\% & 26.3\% & 0.7\% \\
\cmidrule[\heavyrulewidth]{1-5}
\end{tabular}
\end{center}
\end{table}

\begin{table}[htb!]
\small
\begin{center}
\caption{
\textbf{Validation Data Source Performance Comparison for Claude-2.1.}
}
\begin{tabular}{lccccc}
\cmidrule[\heavyrulewidth]{1-5}
\multirow{2}{*}{\bf No.} 
& \multirow{2}{*}{\bf \begin{tabular}[c]{@{}c@{}} Validation \\ Data Source \end{tabular}} 
& \multirow{2}{*}{\bf \begin{tabular}[c]{@{}c@{}} Ground \\ Truth $\alpha$ \end{tabular}}  
& \multicolumn{2}{c}{\bf Estimated} \\
\cmidrule{4-5}
& & & $\alpha$ & $CI$ ($\pm$) \\
\cmidrule[\heavyrulewidth]{1-5}
(1) & \emph{ICLR} 2023 (Claude-2.1) & 0\%    & 4.2\%  & 0.7\% \\
(2) & \emph{ICLR} 2023 (Claude-2.1) & 2.5\%  & 6.5\%  & 0.8\% \\
(3) & \emph{ICLR} 2023 (Claude-2.1) & 5\%    & 8.7\%  & 0.8\% \\
(4) & \emph{ICLR} 2023 (Claude-2.1) & 7.5\%  & 10.9\% & 0.8\% \\
(5) & \emph{ICLR} 2023 (Claude-2.1) & 10\%   & 13.2\% & 0.9\% \\
(6) & \emph{ICLR} 2023 (Claude-2.1) & 12.5\% & 15.4\% & 0.9\% \\
(7) & \emph{ICLR} 2023 (Claude-2.1) & 15\%   & 17.6\% & 0.9\% \\
(8) & \emph{ICLR} 2023 (Claude-2.1) & 17.5\% & 19.9\% & 0.9\% \\
(9) & \emph{ICLR} 2023 (Claude-2.1) & 20\%   & 22.1\% & 0.9\% \\
(10) & \emph{ICLR} 2023 (Claude-2.1)& 22.5\% & 24.4\% & 0.9\% \\
(11) & \emph{ICLR} 2023 (Claude-2.1)& 25\%   & 26.5\% & 0.9\% \\
\cmidrule[\heavyrulewidth]{1-5}
\end{tabular}
\end{center}
\end{table}

\newpage 
\clearpage
\section{Theoretical Analysis on the Sample Size}

{
\begin{theorem}
\label{thm:I1-corrected}
Assume the data $x_1,\dots,x_n$ are drawn i.i.d.\ from the mixture distribution $(1-\alpha^*)P + \alpha^* Q$ for some ground truth $\a^* \in (0,1)$
and that $P,Q$ are known exactly.
Further suppose there exists a constant $\kappa>0$ such that for all $x$, 
\[
\frac{(P(x)-Q(x))^2}{\max\{P(x)^2,Q(x)^2\}} \ge \kappa.
\]

Let $\hat{\a}$ be the MLE obtained by maximizing \eqref{eq: log likelihood}.
This MLE is not too far away from the ground truth $\alpha^*$ with high probability: for any $\delta\in(0,1)$, with probability at least $1-\delta$,
\begin{equation}
\label{eq:alpha-bound-interior}
|\hat{\a}_n-\alpha^*|
\;\le\;
\frac{1}{\kappa\,\alpha^*(1-\alpha^*)}\sqrt{\frac{2\log(2/\delta)}{n}}.
\end{equation}
\end{theorem}

\begin{proof}
Direct computation yields the following equations for the first and second derivatives of $\cL$:
\[
\cL'(\alpha)
=
\frac{1}{n}\sum_{i=1}^n
\frac{Q(x_i)-P(x_i)}{(1-\alpha)P(x_i)+\alpha Q(x_i)}, \hspace{.25in} \cL''(\alpha)
=
-\frac{1}{n}\sum_{i=1}^n
\frac{\big(Q(x_i)-P(x_i)\big)^2}{\big((1-\alpha)P(x_i)+\alpha Q(x_i)\big)^2}.
\]
Since $(1-\alpha)P(x_i)+\alpha Q(x_i)\le \max\{P(x_i),Q(x_i)\}$, we have
\[
\frac{\big(Q(x_i)-P(x_i)\big)^2}{\big((1-\alpha)P(x_i)+\alpha Q(x_i)\big)^2}
\;\ge\;
\frac{\big(Q(x_i)-P(x_i)\big)^2}{\max\{P(x_i)^2,Q(x_i)^2\}}
\;\ge\;\kappa
\]
by assumption. Therefore, for all $\alpha\in[0,1]$, $\cL''(\alpha) \leq -\kappa$, i.e., $\cL$ is $\kappa$-strongly concave on $[0,1]$. A standard consequence of $\kappa$-strong concavity is that for all $a,b\in[0,1]$,
\begin{equation}
\label{eq:strong-concavity-ineq}
\cL(a) \leq \cL(b) + \cL'(b)(a-b) - \frac{\kappa}{2}(a-b)^2.
\end{equation}
Apply~\eqref{eq:strong-concavity-ineq} with $a=\hat{\a}$ and $b=\alpha^*$:
\[
\cL(\hat{\a})
\le
\cL(\alpha^*) + \cL'(\alpha^*)(\hat{\a}-\alpha^*)
- \frac{\kappa}{2}(\hat{\a}-\alpha^*)^2.
\]
Since $\hat{\a}$ maximizes $\cL$, we have $\cL(\hat{\a})\ge \cL(\alpha^*)$, hence
\[
0
\le
\cL'(\alpha^*)(\hat{\a}-\alpha^*)
- \frac{\kappa}{2}(\hat{\a}-\alpha^*)^2.
\]
If $|\hat{\a}-\a^*|=0$ we are done; otherwise divide by $|\hat{\a}-\a^*|$ and use that the inequality forces
$\cL'(\alpha^*)|\hat{\a}-\a^*|\ge 0$:
\begin{equation}
\label{eq:distance-vs-score}
|\hat{\a}-\alpha^*| \leq \frac{2}{\kappa}|\cL'(\alpha^*)|.
\end{equation}

It remains to control the empirical score at the true mixture weight. Define
\[
S_i
:=
\frac{Q(x_i)-P(x_i)}{(1-\alpha^*)P(x_i)+\alpha^* Q(x_i)},
\qquad\text{so that}\qquad
\cL'(\alpha^*)=\frac{1}{n}\sum_{i=1}^n S_i.
\]
Under $x_i\sim (1-\a^*)P + \a^* Q$ we have
\[
\mathbb{E}[S_i]
=
\int \Big((1-\alpha^*)P(x)+\alpha^* Q(x)\Big)
\frac{Q(x)-P(x)}{(1-\alpha^*)P(x)+\alpha^* Q(x)}\,dx
=
\int (Q(x)-P(x))\,dx
=0.
\]

If $\alpha^*\in(0,1)$, it is easily seen that
\[
-\frac{1}{1-\alpha^*}
\le
\frac{Q(x)-P(x)}{(1-\alpha^*)P(x)+\alpha^* Q(x)}
\le
\frac{1}{\alpha^*},
\]
so $S_i\in[-\frac{1}{1-\alpha^*},\frac{1}{\alpha^*}]$. Applying Hoeffding's inequality therefore shows that with probability at least $1-\delta$,
\[
|\cL'(\alpha^*)|
\le
\frac{1}{\alpha^*(1-\alpha^*)}
\sqrt{\frac{\log(2/\delta)}{2n}}.
\]
Combining with~\eqref{eq:distance-vs-score} yields~\eqref{eq:alpha-bound-interior}.
\end{proof}
}

\end{document}